\documentclass[opre,nonblindrev]{informs3_hide}

\OneAndAHalfSpacedXI %

\usepackage{natbib}
 \bibpunct[, ]{(}{)}{,}{a}{}{,}%
 \def\bibfont{\small}%

\usepackage{appendix}
\usepackage{rotating}
\usepackage{fancyvrb}

 \makeatletter
\def\mathcolor#1#{\@mathcolor{#1}}
\def\@mathcolor#1#2#3{%
  \protect\leavevmode
  \begingroup
    \color#1{#2}#3%
  \endgroup
}
\makeatother

\usepackage[utf8]{inputenc}
\usepackage{amsmath,amsfonts,bm}
\usepackage{amssymb}
\usepackage{mathtools}
\usepackage{mathrsfs}
\usepackage{booktabs}
\usepackage{float}
\usepackage[english]{babel} %
\usepackage[protrusion=true,expansion=true]{microtype} %
\usepackage{graphicx}
\usepackage{subfigure}
\usepackage{array}
\usepackage{multirow}
\usepackage[inline]{enumitem}
\usepackage{hhline}
\usepackage[CJKbookmarks=true,
            bookmarksnumbered=true,
			bookmarksopen=true,
			colorlinks=true,
			citecolor=blue,
			linkcolor=blue,
			anchorcolor=red,
			urlcolor=blue]{hyperref}
\usepackage{booktabs} %
\usepackage{setspace}
\usepackage{smile}
\usepackage[dvipsnames]{xcolor}
\usepackage{algorithm}
\usepackage[noend]{algorithmic}

\usepackage{tabularx}
\usepackage{footnote}

\TheoremsNumberedThrough     %

\EquationsNumberedThrough    %

\ECRepeatTheorems

\EquationsNumberedThrough    %

\begin{document}

\RUNAUTHOR{Wang, Zhang, and Zhang}

\RUNTITLE{SPARKLE}

\TITLE{SPARKLE: A Nonparametric Approach for Online Decision-Making with High-Dimensional Covariates}

\ARTICLEAUTHORS{%
\AUTHOR{Wenjia Wang}
\AFF{Department of Industrial Systems Engineering and Management, National University of Singapore, Singapore 117576, \EMAIL{wenjiawang@nus.edu.sg}}
\AUTHOR{Qingwen Zhang}
\AFF{Division of Emerging Interdisciplinary Areas, The Hong Kong University of Science and Technology, Clear Water Bay, Hong Kong Special Administrative Region, China, \EMAIL{qzhangcg@connect.ust.hk}}
\AUTHOR{Xiaowei Zhang}
\AFF{Department of Industrial Engineering and Decision Analytics, The Hong Kong University of Science and Technology, Clear Water Bay, Hong Kong Special Administrative Region, China, \EMAIL{xiaoweiz@ust.hk}}
}

\ABSTRACT{Personalized services are central to today's digital economy, and their sequential decisions are often modeled as contextual bandits. Modern applications pose two main challenges: high-dimensional covariates and the need for nonparametric models to capture complex reward-covariate relationships. We propose SPARKLE, a novel contextual bandit algorithm based on a sparse additive reward model that addresses both challenges through (i) a doubly penalized estimator for nonparametric reward estimation and (ii) an epoch-based design with adaptive screening to balance exploration and exploitation. We prove a sublinear regret bound that grows only logarithmically in the covariate dimensionality; to our knowledge, this is the first such result for nonparametric contextual bandits with high-dimensional covariates. We also derive an information-theoretic lower bound, and the gap to the upper bound vanishes as the reward smoothness increases.
Extensive experiments on synthetic data and real data from video recommendation and personalized medicine show strong performance in high-dimensional settings.
}%

\maketitle

\section{Introduction}

Personalized services are now common in many parts of the digital economy, including product recommendation \citep{Li10contextual}, online advertising \citep{Wen17online}, mobile health \citep{Tewari2017}, where tailored recommendations are key to user engagement and satisfaction.
A standard way to formalize how these systems make decisions is the contextual bandit framework \citep{LattimoreSzepesvari2020}.

In this framework, a decision-maker repeatedly observes user-specific information (covariates), selects an action (arm), and then receives a feedback signal (reward).
The goal is to choose actions over time to maximize cumulative reward.
The challenge is that the reward function (i.e., the mapping from a covariate–arm pair to expected reward) is unknown and must be learned from ongoing interactions.

Applying contextual bandits in data-rich settings raises two central difficulties.
First, modern platforms observe many user attributes, so the covariates are high-dimensional.
Learning the reward function in such settings is hard due to the curse of dimensionality \citep{buhlmann2011statistics}.
Second, the true relationship between covariates, arm, and reward can be complex and highly nonlinear.
In such cases, linear or other parametric models are likely to be misspecified, which can lock the system into persistently suboptimal decisions and lead to cumulative regret that grows linearly over time \citep{hu2022smooth}.

Although high-dimensional nonparametric regression has attracted substantial interest in the machine learning literature
\citep{RavikumarLaffertyLiuWasserman09,raskutti2012minimax,tan2019doubly},
most existing studies focus on offline settings and typically assume independent observations.
In contextual bandit problems, by contrast, data are collected adaptively: at each time the decision-maker selects an arm based on past data and then observes a reward only for that arm.
This feedback structure couples the data with past decisions and introduces sophisticated intertemporal dependencies,  making standard offline analyses inapplicable.

The bandit setting also requires a careful exploration--exploitation strategy to learn about seemingly inferior arms without incurring excessive costs that significantly reduce cumulative reward.
Compared to parametric models, nonparametric reward models make the design and analysis of this strategy more challenging.
Parametric models impose global structure that facilitates extrapolation, allowing effective learning without dense coverage of the covariate space and thus reducing the need for exploration
\citep{bastani2021mostly}.
By contrast,
nonparametric models heavily rely on local information: accurate estimation of an arm’s reward at a given covariate value requires sufficient nearby observations.
However, bandit feedback reveals rewards only for selected arms and, through adaptive exploitation, tends to allocate observations to arms that already appear promising, leaving others with inadequate samples and higher estimation error.
As a result, nonparametric models are particularly vulnerable to the uneven, adaptively skewed coverage of the covariate space induced by bandit feedback  \citep{rigollet2010nonparametric,hu2022smooth}.
High dimensionality further complicates matters because reliably identifying the relevant covariates is itself challenging.

\subsection{Main Contributions}

We introduce a new contextual bandit algorithm tailored to high-dimensional covariates and nonparametric rewards. It achieves sublinear regret with only logarithmic dependence on the dimensionality.
We highlight our contributions as follows.

\subsubsection{Algorithm Design.}

We model each arm’s reward as a sum of univariate nonparametric functions, where only a small subset are nonzero and thus correspond to significant covariates.
The sparse additive structure provides sufficient modeling flexibility while remaining interpretable in high dimensions.
Assuming each component lies in a reproducing kernel Hilbert space (RKHS), we estimate the reward using a doubly penalized approach \citep{tan2019doubly} that controls function complexity via RKHS norms and induces variable selection.

Building on this reward model, we develop SPARKLE (SParse Additive Regularized Kernel LEarning), an epoch-based algorithm that combines progressive arm elimination with doubly penalized estimation in high dimensions.
The algorithm uses adaptive screening to match estimation effort to problem difficulty.
Early rounds use coarse estimates, built from fewer observations, to quickly remove inferior arms in parts of the covariate space where they are clearly separable. Later rounds apply more accurate estimation where arms are hard to distinguish.
This structure focuses effort where finer distinctions most improve decisions and thus lowers regret. Key design choices, such as epoch length, arm selection rules, and penalization levels, are determined by our new analysis of the doubly penalized estimator in the online setting.

\subsubsection{Theoretical Guarantees.}

We derive the first sublinear regret upper bound for high-dimensional nonparametric contextual bandits.
Notably, its dependence on the dimensionality $d$ is only logarithmic, indicating that the algorithm can handle high-dimensional covariates:
\[\tilde O\Bigl(T^{1-\frac{(2m-1)(1+\alpha)}{4m+2}}\log(d)\Bigr),\]
where $m$ is the RKHS smoothness of the reward functions, $\alpha$ is a margin parameter, and logarithmic factors in $T$ are suppressed.\footnote{The dependence on the sparsity level $s_0$ is also suppressed here; see Section~\ref{sec_regret} for details.}
We also establish an information-theoretic lower bound for RKHS contextual bandits.
The gap between the upper and lower bounds diminishes as the smoothness parameter $m$ increases.

A key technical challenge in our regret analysis is that adaptive arm selection alters the \emph{sample support} for each arm---that is, the set of covariate values that can be assigned to that arm with positive likelihood---even though the overall covariate support remains unchanged.
In effect, adaptive selection can leave some regions of the covariate space unavailable to certain arms.
This geometry complicates the analysis and makes prior results on the doubly penalized estimator, which assume fixed support in offline settings, inapplicable here.

To address this, we show that the sample supports induced by SPARKLE share a common geometric property we call \emph{$\mathscr{C}$-regularity}: for sets with this property, the projections onto each coordinate axis are finite unions of well-separated intervals.
This structure supports a rigorous error analysis via Sobolev space theory and yields uniform bounds in $L_\infty$, $L_2$, and RKHS norms under adaptive sampling.
This contrast much of the high-dimensional statistics literature \citep{buhlmann2011statistics}, which emphasizes $L_p$ norms in static settings.

\subsubsection{Numerics.}
We evaluate SPARKLE through comprehensive numerical experiments on both synthetic and real-world data.
On synthetic data, we confirm the predicted scaling of cumulative regret: logarithmic in dimensionality $d$, sublinear in horizon $T$, and geometric in sparsity $s_0$, with empirical exponents close to theory. Across baselines, SPARKLE achieves consistently lower regret, and the gap widens as $d$ increases.

This advantage extends to real-world online decision-making with high-dimensional covariates, including video recommendation and personalized medicine.
In short‑video recommendation, SPARKLE leverages continuous reward signals and rich user features to reduce regret and improve engagement.
In warfarin dosing, where linear models perform well offline \citep{international2009estimation}, SPARKLE remains competitive with a strong high-dimensional linear bandit algorithm baseline \citep{bastani2020online} and performs reliably even under a binary reward, outside its intended continuous‑reward setting.
These results show that SPARKLE handles complex, high-dimensional problems and maintains robust performance across reward structures and context types, underscoring its practical value.

\subsection{Related Literature}

The contextual bandit literature is extensive, with linear reward models serving as the main workhorse for algorithm design. Notable examples include \cite{Li10contextual}, \cite{abbasi2011improved}, \cite{goldenshluger2013linear}, \cite{AgrawalGoyal13}, \cite{dimakopoulou2019balanced}, and \cite{bastani2021mostly}. See also \cite{Agrawal19} and \citet{LattimoreSzepesvari2020} for overviews.

High-dimensional covariates complicate estimation and inference of unknown reward functions \citep{buhlmann2011statistics}. Algorithms designed for low or moderate dimensions often see performance degrade as the dimension grows. In particular, a regret bound that is sublinear in $T$ at low dimension may cease to be sublinear when the bound depends \emph{polynomially} on the dimensionality $d$. For example, the Greedy-First algorithm of \cite{bastani2021mostly} has cumulative regret $O(d^2\log^{3/2}(d)\log(T))$. In high-dimensional settings, $d$ can be comparable to or larger than the time horizon $T$, which prevents sublinear regret in $T$. Substantial progress has been made on high-dimensional linear contextual bandits by adapting sparse learning methods to online settings \citep{wang2018minimax,bastani2020online,hao2020high,QianIngLiu23,ChenWangFangWangLi24,RenZhou24}.
These algorithms estimate high-dimensional linear rewards using tools such as penalized regression with the minimax concave penalty \citep{Zhang10}, LASSO \citep{Tibshirani96}, interactive greedy procedures \citep{qian19an}, and best subset selection \citep{Miller02}.

The importance of nonparametric methods in online decision‑making has been increasingly recognized, as they better capture complex reward‑covariate relationships and reduce model misspecification \citep{chen2021nonparametric,
chen2022primal}.
Many studies relax linearity to general linearized model \citep{ren2020batched, blanchet2024delay} or assume rewards lie in infinite-dimensional function classes, typically Lipschitz or H\"older classes
\citep{yang2002randomized,slivkins2011contextual,valko2013finite,perchet2013multi,gur2022smoothness}.
However, nonparametric rewards pose challenges even in low-dimensional contextual bandit problems.
For H\"older rewards, \cite{hu2022smooth} establishes a minimax bound of
$\tilde{O}\bigl(\max\{T^{(\beta+d-\alpha \beta)/(2 \beta+d)}, 1\}\bigr)$ for cumulative regret, where $\beta$ is the H\"older smoothness parameter and $\alpha$ is a margin parameter.
Their algorithm integrates local polynomial regression with adaptive binning \citep{perchet2013multi}.
This approach is not well suited to high dimensions. Local polynomial regression depends on local neighborhoods, and in high dimensions it is hard to gather enough data in any small region because of the curse of dimensionality.

We instead assume reward functions lie in an RKHS with exponential spectral decay governed by a smoothness parameter $m$.
This leads to different theory and modeling behavior: RKHSs encode global structure through the kernel, while H\"older classes impose only local smoothness.
Practically, the global dependence makes RKHS intrinsically more adaptable to estimation with abundant covariates: kernels implicitly perform dimension reduction by weighting feature interactions according to the data geometry, whereas local methods suffer directly from the curse of dimensionality when coordinating information across sparse regions.

There are few studies on high-dimensional nonparametric contextual bandits; notable exceptions are \cite{reeve2018k} and \cite{li2023dimension}.
Both assume low effective dimensionality of the covariates and apply dimension reduction. \cite{reeve2018k} considers covariates supported on a low-dimensional manifold embedded in a high-dimensional space and develops a $k$-nearest-neighbor UCB algorithm.
\cite{li2023dimension} studies continuum-armed contextual bandits with $d$-dimensional covariates, combining localized LASSO with binning and voting to obtain regret that depends on  $d^*$ rather than $d$, where $d^*$ is the effective dimension.
However, both works assume $d$ is \emph{fixed} and does not grow with the time horizon $T$.
Our focus is the regime where $d$ can be larger than $T$ (so may increase with $T$), which is common in the high-dimensional contextual bandit literature.

\smallskip

Throughout the paper, we use the following notation.
For real numbers $a,b$, let $a\vee b=\max(a,b)$ and $a\wedge b=\min(a,b)$. Denote by $\lceil a\rceil$ the smallest integer greater than or equal to $a$ and by $\lfloor a\rfloor$ the largest integer less than or equal to $a$.
For sequences $(a_n)$ and $(b_n)$, we write $b_n\lesssim a_n$ to mean there exists a constant $C>0$ such that $b_n\le C a_n$ for some constant $C>0$ and all $n$, and $a_n\asymp b_n$ when both $a_n\lesssim b_n$ and $b_n\lesssim a_n$.
For a vector $\bv$, we treat it as a column vector by default and use $\bv^\intercal$ for its transpose.

The remainder of the paper is organized as follows. We formulate the sparse additive reward model in Section~\ref{sec:nonparacontextbandit} and conduct the (offline) doubly penalized estimation on a $\mathscr{C}$-regular support in Section~\ref{sec:offlinedoubly}.
Our contextual bandit algorithm, SPARKLE, is detailed in Section~\ref{sec:SPARKLE}, followed by the primary theoretical results in Section \ref{sec_regret}. We present simulation experiments in  Section~\ref{sec_numexp} and conclude  in Section~\ref{sec:con}. Technical proofs are included in the e-companion.

\section{Problem Formulation}\label{sec:nonparacontextbandit}

Consider a contextual bandit problem involving $K$ arms and a time horizon $T$.
At each time $t = 1, \ldots, T$, a decision-maker observes a
$d$-dimensional covariate vector $\bx_t$ which is generated independently from a fixed but unknown distribution with support $\Omega\subset \RR^d$, then selects an arm $\pi_t \in \{1,\ldots,K\}$ to pull, and receives a random reward $y_t \in \RR$.
Each arm $k$ is associated with an unknown reward function $f_k^*(\bx)$, and the reward $y_t$ corresponds to a noisy observation of the reward function for arm $k$:
\begin{equation}\label{eq:yi}
    y_t = f_k^*(\bx_t) + \varepsilon_{k,t} \quad \mbox{if }\pi_t = k,
\end{equation}
where $\varepsilon_{k,t}$ is an independent random variable with mean zero and finite variance, representing the noise.

The goal of the decision-maker is to design a policy $\pi$ which selects arm $\pi_t$ at time $t$ based on the previously collected data $\{\bx_1, y_{\pi_1,1},\ldots, \bx_{t-1}, y_{\pi_{t-1}, t-1}, \bx_t\}$ to maximize the expected cumulative reward over the time horizon.
We evaluate policy $\pi$  by comparing it to an \emph{oracle} policy $\pi^*$, which, with full knowledge of the reward functions, selects the arm $\pi^*_t = \argmax_{k} f^*_k(\bx_t)$ at time $t$.
Thus, the decision-maker seeks a policy $\pi$ to minimize the expected cumulative regret
\[ R_T(\pi) \coloneqq \mathbb{E} \biggl[\sum_{t=1}^T  \left(\max_k f^*_k (\bx_t)  -  f^*_{\pi_t}(\bx_t)\right)\biggr], \]
where the expectation is taken with respect to the joint distribution of the covariate sequence, the random rewards, and any random variables involved in defining policy $\pi$.

\subsection{Sparse Additive Reward Model}

Online decision-making in data-rich settings often involves many covariates and complex covariate–reward relationships. We therefore focus on nonparametric reward models that remain computationally efficient with high-dimensional covariates. Sparse additive models are well suited to this goal.

Suppose the reward function for each arm has an additive structure, where each component is a univariate function for a particular coordinate. With $\bx = (x_{(1)},\ldots,x_{(d)})$,
\begin{align}\label{eq_addreward}
    f_k^*(\bx) = \sum_{j=1}^{d} f^*_{k,(j)}(x_{(j)}),
\end{align}
where $f^*_{k,(j)}$ is a univariate function from a function class $\cF_j$, for all $k = 1,\ldots,K$ and $j=1,\ldots,d$.

To introduce sparsity, suppose that for each arm $k$, most components $f^*_{k,(j)}$ are zero.
Let $s_k \coloneqq \sum_{j=1}^d \II\{f^*_{k,(j)}\neq 0\}$ be the number of nonzero components of $f_k^*$, where $\II$ denotes the indicator function.
Sparsity means $s_k \ll d$.
Formally, we assume that for each $k=1,\ldots,K$, $f_k^* \in \cF$, where
\begin{align}\label{eq_functionclass}
  \cF \coloneqq  \left\{f =\sum_{j=1}^{d} f_{(j)}\left(x_{j}\right):f_{(j)} \in \cF_j \mbox{ for all } j=1, \ldots, d,\mbox{  and } \sum_{j=1}^d \II\{f_{(j)}\neq 0\} \leq s_0 \right\},
\end{align}
and $s_0$ is a pre-specified sparsity level.
Although $s_k$ is unknown, we may choose $s_0$ large enough to ensure $s_k \le s_0$ for all $k$.\footnote{Sparse additive models have received considerable attention in machine learning for balancing complexity and interpretability \citep{lin2006component,RavikumarLaffertyLiuWasserman09,meier2009high,raskutti2012minimax,tan2019doubly}.
The recent work of \cite{cai2022stochastic} studies bandit problems with sparse additive models, focusing on continuous-armed settings where the arm $\bx$ is a $d$-dimensional vector and the goal is to maximize a sparse additive reward $f(\bx)$.}

Suppose that for each $j$, the univariate function class $\cF_j = \mathcal{N}_{\Psi_{(j)}}$, an RKHS induced by a positive definite kernel $\Psi_{(j)}$.
We choose $\Psi_{(j)}$ to be a smooth kernel that induces an RKHS of functions differentiable up to a specified order;  see Section~\ref{sec:assumptions} for details.
An attractive feature of RKHSs is that estimating an RKHS function, an infinite-dimensional problem, reduces to a finite-dimensional one by the representer theorem \citep{KimeldorfWahba71}.
Section~\ref{subsec:RKHS} of the e-companion provides a brief introduction to RKHSs.

\subsection{Doubly Penalized Estimation}\label{subsec:doubly}

We adopt the double penalization method of \cite{koltchinskii2010sparsity} and \cite{raskutti2012minimax} for estimating sparse additive models in offline settings.\footnote{Other estimators for sparse additive models include those proposed by \cite{lin2006component} and \cite{RavikumarLaffertyLiuWasserman09}. The doubly penalized estimator in \eqref{eq_dp} generalizes these methods and is minimax-optimal; see \cite{raskutti2012minimax} and \citet[Chapter~4.4]{HastieTibshiraniWainwright15} for further discussion.}

Consider an additive kernel $\Psi(\bx,\tilde{\bx}) = \sum_{j=1}^d \Psi_{(j)}(x_{(j)}, \tilde{x}_{(j)})$
and a function $f^*\in \mathcal{N}_{\Psi}$.
Then, this function is additive with each component $j$ belong to $\mathcal{N}_{\Psi_{(j)}}$.
Let $\{(\bx_i, y_i):i=1,\ldots,n\}$ be a set of independent and identically distributed (i.i.d.) noisy observations of $f^*$.
For any additive function $f = \sum_{j=1}^d f_{(j)}$ with $f_{(j)}\in \mathcal{N}_{\Psi_{(j)}}$,
we define two different norms:
\begin{align}
    \|f\|_{\mathcal{N}_{\Psi}, 1} \coloneqq \sum_{j=1}^d \|f_{(j)}\|_{\mathcal{N}_{\Psi_{(j)}}}\quad \mbox{and} \quad
    \|f\|_{n, 1} \coloneqq \sum_{j=1}^d \|f_{(j)}\|_n.
\end{align}
Here, $\|\cdot\|_{\mathcal{N}_{\Psi_{(j)}}}$ denotes the RKHS norm on $\mathcal{N}_{\Psi_{(j)}}$, and
$\|\cdot\|_n$ denotes the \emph{empirical norm} with respect to the sample $\{(\bx_i,y_i) :i=1,\ldots,n\}$, defined as
$\|f_{(j)}\|_n \coloneqq \sqrt{n^{-1}\sum_{i=1}^n f_{(j)}^2(x_{i,(j)})}$,
where $x_{i,(j)}$ is the $j$-th element of $\bx_i =  (x_{i,(1)},\ldots, x_{i,(d)}) $.

The doubly penalized estimator of $f^*$ solves the regularized empirical risk minimization problem:
\begin{align}
\min_{f\in \mathcal{N}_{\Psi}} \frac{1}{n}\sum_{i=1}^n (y_i-f(\bx_i))^2 + \rho \|f\|_{\mathcal{N}_{\Psi}, 1} + \lambda  \|f\|_{n, 1}, \label{eq_dp}
\end{align}
where $\rho,\lambda>0$ are regularization parameters that control smoothness and sparsity, respectively, addressing the nonparametric and high-dimensional aspects of the problem.

By the representer theorem, the estimator can be expressed as
\begin{align}
    \hat{f}^{\mathsf{DP}} \coloneqq \sum_{i=1}^n \sum_{j=1}^d \hat{\beta}_{i,(j)} \Psi_{(j)}(\cdot, x_{i,(j)}), \label{eq:DP-estimator}
\end{align}
and the coefficient vector $\hat{\bbeta}_{(j)} = (\hat{\beta}_{1,(j)},\ldots,\hat{\beta}_{n,(j)})^\intercal$
are the optimal solution to
\begin{align}\label{eq_solvebeta}
    \min_{\substack{\bbeta_{(j)} \in \RR^n \\ j=1,\ldots,d}} \frac{1}{n} \biggl\| \by - \sum_{j=1}^d \bPsi_{(j)}\bbeta_{(j)}\biggr\|_2^2 + \rho \sum_{j=1}^d \sqrt{\bbeta_{(j)}^\intercal \bPsi_{(j)}\bbeta_{(j)}} + \lambda \sum_{j=1}^d \sqrt{\frac{1}{n}  \| \bPsi_{(j)} \bbeta_{(j)} \|_2^2},
\end{align}
where $\bPsi_{(j)}$ is the Gram matrix with entries $\Psi_{(j)}(x_{i,(j)}, x_{l,(j)})$ for all $i,l=1,\ldots,n$.

Although \eqref{eq_solvebeta} has no closed-form solution, it is a second-order cone program (SOCP), a standard convex optimization problem that can be efficiently solved by interior-point methods; see \cite{lobo1998applications}.
In the online setting, the sample $\{(\bx_i,y_i):i=1,\ldots,n\}$, and hence $\hat{f}^{\mathsf{DP}}$, depend implicitly on the adaptive data collection policy: the indices $i$ used in estimation varies across arms $k$ and changes over time. Consequently, the regularization parameters $(\rho,\lambda)$ should be adapted to these changing conditions.

\section{Rate Analysis of Doubly Penalized Estimator Under $\mathscr{C}$-Regular Support} \label{sec:offlinedoubly}

Although convergence rates for doubly penalized estimators are known in the machine learning literature, those results do not apply to our setting for two reasons. First, they usually assume that the support of the covariate vector $\bx$ is fixed. This makes sense in static settings where all data are observed at once. In our bandit setting, however, adaptive data collection makes the set of covariate values that an arm can receive with positive probability change over time.
This is less of a concern for parametric models but is critical for nonparametric reward estimation.
Second, prior analyses typically measure error in the $L_2$ norm, while regret analysis requires pointwise control, in particular $L_\infty$ bounds.

In Section~\ref{sec:assumptions} we state the assumptions used throughout the paper. In Section~\ref{sec:c-regularity} we define $\mathscr{C}$-regularity, a geometric property of the time-varying sample supports induced by our bandit algorithm.
In Section~\ref{sec:rate-with-c-regularity}, we present convergence rates for the doubly penalized estimator under $\mathscr{C}$-regular support, in the $L_2$, $L_\infty$, and RKHS norms.

\subsection{Basic Assumptions} \label{sec:assumptions}

\begin{assumption}[Nonvanishing Density]\label{assum_densitybound} The covariates $\{\bx_t:t=1,2,\ldots\}$ are i.i.d. draws from a fixed distribution $\mathsf{P}_X$. This distribution has a density $p(\bx)$ on $\Omega\subseteq[0,1]^d$ such that $p_{\min}\le p(\bx)\le p_{\max}$ for all $\bx\in\Omega$, for some constants $0<p_{\min}\le p_{\max}<\infty$.
\end{assumption}

\begin{assumption}[Sub-Gaussian Noise]\label{assum_subG}
The noise variables $\{\varepsilon_{k,t}:k=1,\ldots,K,\, t=1,2,\ldots T \}$ are i.i.d., mean-zero, and sub-Gaussian.
\end{assumption}

\begin{assumption}[Smooth Kernels]\label{assump:kernel} Let $\Omega_j$ be the projection of $\Omega\subset\mathbb{R}^d$ onto the $j$-th dimension.
For each $j=1,\ldots,d$, $\cF_j\subseteq \cN_\Phi(\Omega_j)$, where $\Phi:[0,1]\times[0,1]\mapsto\mathbb{R}$ is a positive definite kernel and $\cN_\Phi(\Omega_j)$ is its RKHS.
The kernel $\Phi$ is stationary, i.e., $\Phi(x,\tilde{x})=\phi(|x-\tilde{x}|)$ for some function $\phi$ and all $x,\tilde{x}\in[0,1]$.
Its Fourier transform $\mathscr{F}_\phi(\omega) \coloneqq  \int_{\RR} \phi(r) e^{-2\pi\mathrm{i} \omega r} \,\mathrm{d} r$ satisfies
\begin{equation}\label{eq:Fourier}
c_{1}\left(1+\omega^2\right)^{-m} \leq \mathscr{F}_\phi(\omega) \leq c_{2}\left(1+\omega^2\right)^{-m},
\end{equation}
for all $\omega\in \RR$ and some constants $m>3/2$, $c_1>0$, and $c_2>0$.
\end{assumption}

\begin{assumption}[Compatibility Condition]\label{assum_compatibility}
There exist constants $\xi>0$ and $\kappa>0$ such that for
any function $f \in \cF$ and
subset $S\subset \{1,...,d\}$ with $|S| \leq s_0 \leq C \sum_{j=1}^d \II\{f_{(j)}\neq 0\}$,
if
$
\sum_{j \in S^\complement}\|f_{(j)}\|_{L_2(\Omega)} \leq \xi \sum_{j \in S}\|f_{(j)}\|_{L_2(\Omega)},
$
then
$\kappa^{2}\bigl(\sum_{j \in S} \|f_{(j)}\|_{L_2(\Omega)}\bigr)^2 \leq\sum_{j \in S}\|f\|_{L_2(\Omega)}^2.
$
Here, $C>0$ is a constant independent of $d$,
$\cF$ and $s_0$ are defined in \eqref{eq_functionclass},
and $S^\complement$ denotes the complement of $S$.
\end{assumption}

Assumptions~\ref{assum_densitybound} and \ref{assum_subG} are standard in the contextual bandit literature \citep{perchet2013multi,bastani2020online,gur2022smoothness,hu2022smooth}.
Under Assumption~\ref{assump:kernel},
the RKHS $\cN_\Phi([0,1])$ equals the Sobolev space $\cW^m([0,1])$ as sets of functions,
and their norms are equivalent are equivalent; see Corollary~10.13 of \cite{wendland2004scattered}.
We assume $m>3/2$ because, by the Sobolev embedding theorem \citep{adams2003sobolev}, all functions in $\cW^m([0,1])$ are differentiable up to order $m-1/2$, thereby having continuous derivatives.

An important example satisfying \eqref{eq:Fourier} is the Mat\'ern class of kernels:
\[
\Phi(x,\tilde{x}) =  \frac{1}{\Gamma(\nu)2^{\nu-1}}\biggl(\frac{2\sqrt{\nu} |x-\tilde{x}|}{\ell}\biggr)^{\nu} \mathsf{B}_{\nu}\biggl(\frac{2\sqrt{\nu} |x-\tilde{x}|}{\ell}\biggr),
\]
where, $\nu >0$ is the smoothness parameter,
$\ell>0$ is the lengthscale, $\Gamma$ is the Gamma function, and $\mathsf{B}_{\nu}$ is the modified Bessel function of the second kind.
Mat\'ern kernels satisfy \eqref{eq:Fourier} with $m=\nu+1/2$ if $\nu>1$.

Assumption~\ref{assum_compatibility} is standard in the high-dimensional additive regression  literature \citep{meier2009high,tan2019doubly}.
It parallels the compatibility condition used in high-dimensional linear regression and linear contextual bandits  \citep{bickel2009simultaneous,bastani2020online}.
For any vector $\bv\in \RR^d$ with $\|\bv_{S^\complement}\|_1\leq \|\bv_S\|_1$, $\|\bv_S\|_1^2\leq C|S|\bv^\intercal \Mb \bv$ for some constant $C>0$, where $\bv_S \coloneqq (v_{(1)}\II\{1\in S\},\ldots, v_{(d)}\II\{d\in S\})$, $\|\cdot\|_1$ denotes the $L_1$-norm of a vector, and $\Mb$ is the covariance matrix of the covariate $\bx$.
Setting $f_{(j)}(x_{(j)}) = v_{(j)}x_{(j)}$ shows that Assumption~\ref{assum_compatibility} generalizes this linear compatibility condition and is in fact weaker.

\subsection{$\mathscr{C}$-Regularity}\label{sec:c-regularity}

We introduce a geometric property called $\mathscr{C}$-regularity, which underpins our analysis.
In offline settings this property is usually implicit for the covariate support.
In our online setting it matters because, although covariates come from a fixed support $\Omega$, a bandit algorithm may avoid parts of $\Omega$ for a given arm.
The resulting effective support for that arm is a random, time-varying subset of $\Omega$ with nontrivial geometry (see Section~\ref{sec:sample-supports}), and this substantially affects nonparametric estimation of that arm’s reward function.

\begin{definition}[$\mathscr{C}$-regularity]\label{def:c-regular}
    A set $\tilde{\Omega} \subset \RR^d$ is $\mathscr{C}$-regular if the following conditions hold:
    \begin{enumerate}[label=(\arabic*), noitemsep, nolistsep]
        \item For each $j=1,\ldots,d$, there exists an integer $p_j\geq 1$ and real numbers $a_{j,1}<b_{j,1}<\cdots<a_{j,p_{j}}<b_{j,p_{j}}$ such that
        $\bigl\{x_{(j)}: \bx=(x_{(1)},\ldots,x_{(d)})\in\tilde{\Omega}\bigr\} =  \bigcup_{l=1}^{p_{j}}[a_{j,l},b_{j,l}]$.
        \item $\min_{1\leq j\leq d,1\leq l\leq p_{j}}(b_{j,l} - a_{j,l})\geq \mathscr{C}$ and $\min_{1\leq j\leq d,1\leq l\leq p_{j}-1} (a_{j,l+1} - b_{j,l})\geq \mathscr{C}$.
    \end{enumerate}
\end{definition}

Equivalently, for each coordinate $j$, the projection of $\tilde{\Omega}$ is a finite union of closed intervals; every interval has length at least $\mathscr{C}$, and any two adjacent intervals are separated by at least $\mathscr{C}$.

The $\mathscr{C}$-regularity of the sample support is central to our error analysis. RKHS estimators depend on the full support of the target function, so violations of $\mathscr{C}$-regularity break standard approximation guarantees.
This property also allows us to use Sobolev extension theorems to match the RKHS and Sobolev spaces via Sobolev extension theorems, which in turn yields a unified convergence rate analysis for the doubly penalized estimator across multiple norms (Theorem~\ref{thm:superrorbound}).

\begin{figure}[ht]
    \FIGURE{
    \includegraphics[width=0.4\textwidth]{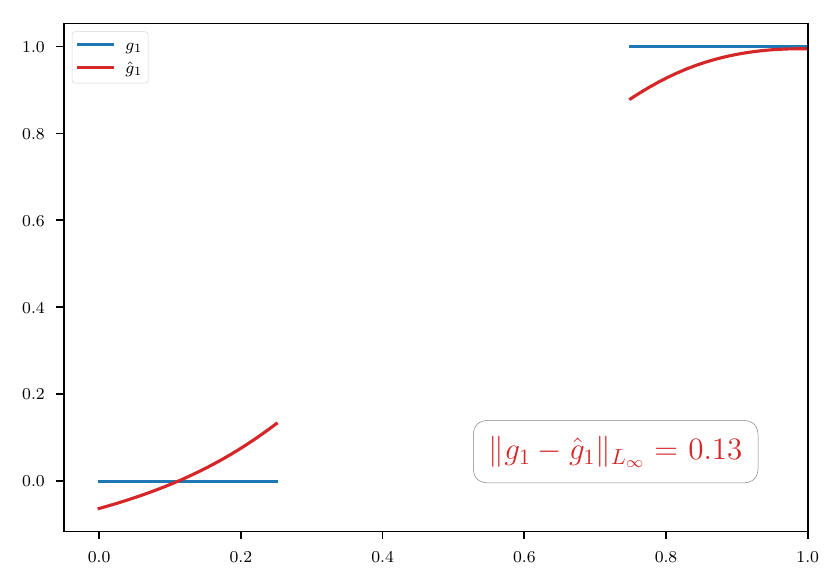}
    \includegraphics[width=0.4\textwidth]{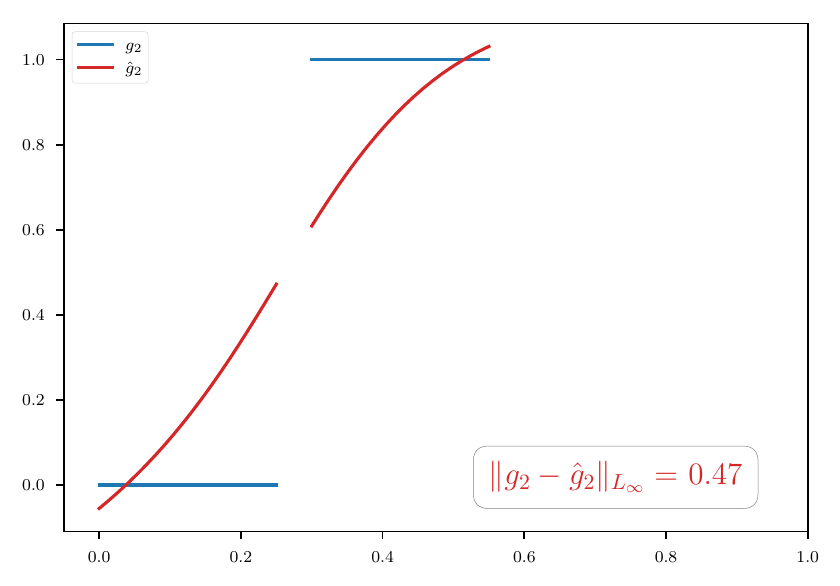}
    }
    {Small Support Gap Yields Large $L_\infty$ Error. \label{fig:g1g2}}
    {Left: $g_1(x)=0$ on $[0,\tfrac{1}{4}]$ and $g_1(x)=1$ on $[\tfrac{3}{4},1]$. Right: $g_2(x)=0$ on $[0,\tfrac{1}{4}]$ and $g_2(x)=1$ on $[\tfrac{3}{10},\tfrac{11}{20}]$. The estimators $\hat g_1$ and $\hat g_2$ are kernel ridge regression fits using 40 samples from each function, with $x\sim\mathrm{Unif}[0,1]$ and Gaussian noise $\sigma^2=0.05$.}
\end{figure}

To illustrate, consider the toy example in Figure~\ref{fig:g1g2} with functions $g_1$ and $g_2$. Each has support components of the same length, but the gap between components differs significantly. Although $g_1$ and $g_2$ have similar Sobolev norms,\footnote{We use the standard Sobolev norm
$\lVert g\rVert_{\cW^k(\tilde\Omega)}=\bigl(\sum_{|\alpha|\le k}\lVert D^\alpha g\rVert_{L_2(\tilde\Omega)}^2\bigr)^{1/2}$, where $D^\alpha$ denotes the weak derivative with multi-index $\alpha$.}
the estimator for $g_2$ has a much larger $L_\infty$ error because its gap is narrower. This shows the importance of keeping enough separation between support components.

\subsection{Convergence Rates}\label{sec:rate-with-c-regularity}

We establish convergence rates for the doubly penalized estimator under three norms, extending offline results on fixed domains to $\mathscr{C}$-regular supports, which we need for online learning.
The analysis proceeds in two steps.
First, we derive an $L_2$ error bound by adapting the arguments of \citet{tan2019doubly} to $\mathscr{C}$-regular supports.
Second, combining this $L_2$ bound with a new bound in the RKHS and Sobolev norms, we obtain the $L_\infty$ bound via sharp interpolation inequalities that relate different norms of Sobolev functions.
Our main technical novelty is tight control of the pointwise error under the geometry of $\mathscr{C}$-regular supports while incorporating sparsity-inducing penalties. This yields explicit dependence on the sparsity level $s$ and the factor $\log(d)$ in the convergence rates.

\begin{theorem}\label{thm:superrorbound}
Suppose Assumptions~\ref{assum_densitybound}--\ref{assum_compatibility} hold.
Let $\tilde{\Omega}\subseteq \Omega$ be $\mathscr{C}$-regular.
Consider i.i.d. samples $\{(\tilde{\bx}_i, \tilde{y}_i):i=1,\ldots,n\}$,
where $\tilde{\bx}_i\sim {\PP}_X(\cdot|\bx\in\tilde{\Omega})$ has a density bounded away from 0 and $\infty$, and
$\tilde{y}_i = f^*(\tilde{\bx}_i) + \tilde{\varepsilon}_i$ with $f^*\in\cF$ and sub-Gaussian noise $\tilde{\varepsilon}_i$.
Let $s\coloneqq \sum_{j=1}^d \II\{f^*_{(j)}\neq 0\}$, and let $\hat f$ be the doubly penalized estimator of $f^*$.
For any $\delta\in(0,1/2)$,
define
$
\rho_{n}=\lambda_n w_n$ and $\lambda_n=C_3 \bigl(\gamma_{n}+ \left(n^{-1}\log(d/\delta)\right)^{\frac{1}{2}}  \bigr),$ where $w_{n} \coloneqq{}  \bigl( C_4^{\frac{2m}{2m+1}} n^{-\frac{m}{2m+1}} \bigr) \vee \bigl(n^{-1}\log(d/\delta)\bigr)^{\frac{1}{2}}$, $
    \gamma_{n}  \coloneqq{}   \bigl( C_4^{ \frac{2m}{2m+1}} n^{-\frac{m}{2m+1}} \bigr) \wedge \bigl(C_4 n^{-\frac{1}{2}}\left(n^{-1}\log(d/\delta)\right)^{-\frac{1}{4m}} \bigr)$ for some constants $C_3,C_4>0$ independent of $(s,d,n)$.
If
\begin{equation}\label{eq:dim-slow-grow}
    \lim_{n\to\infty} w_{n}^{-\frac{2m-1}{4m^2}} \gamma_{n}+w_{n}^{-\frac{1}{2m}} (n^{-1}\log(d/\delta))^{\frac{1}{2}}= 0,
\end{equation}
then with probability at least $1-2\delta$, then
\begin{align}
    \|\hat{f}-f^*\|_{L_\infty(\tilde{\Omega})}  \leq{}&  Cs^{\frac{2m+1}{4m}} \Bigl(n^{-\frac{m}{2m+1}}+\left(n^{-1}\log(d/\delta)\right)^{\frac{1}{2}} \Bigr)^{1-\frac{1}{2m}}, \label{eq:L_inf-error}\\
    \|\hat{f}-f^*\|_{L_2(\tilde{\Omega})}
    \leq{}& C^\prime s^{\frac{1}{2}}\Bigl(n^{-\frac{m}{2m+1}}+\left(n^{-1}\log(d/\delta)\right)^{\frac{1}{2}} \Bigr), \label{eq:L_2-error}\\
    \|\hat{f}-f^*\|_{\mathcal{N}_\Psi(\tilde{\Omega})} \leq{}&  C^{\prime\prime}s, \label{eq:RKHS-error}
\end{align}
for some constants $C, C', C''>0$ independent of $(s,d,n)$.
\end{theorem}

As in high-dimensional statistical learning, Theorem~\ref{thm:superrorbound} allows the dimensionality $d$ to grow with the sample size $n$, provided the growth is not too fast (exponential growth is permitted), as required by the condition~\eqref{eq:dim-slow-grow}. This condition is automatically satisfied when $d$ is fixed. The $L_2$ bound \eqref{eq:L_2-error}  was established by \cite{tan2019doubly} for fixed domains; here we handle $\mathscr{C}$-regular supports.
The $L_\infty$ bound \eqref{eq:L_inf-error} and the RKHS bound \eqref{eq:RKHS-error} are new.

The $L_\infty$ bound is critical because it controls instantaneous regret in the bandit analysis. Its proof combines tools from functional analysis and high-dimensional statistics via Sobolev extension theorems \citep{adams2003sobolev}.
The $\mathscr{C}$-regularity of the sample support $\widehat{\Omega}$ (where the hat emphasizes its stochastic nature) ensures that any $f\in\mathcal{W}^m(\widehat{\Omega})$ extends to $\widetilde{f}\in\mathcal{W}^m(\mathbb{R})$ with a controlled extension constant.

The RKHS analysis underpins the $L_\infty$ result and also helps characterize the geometry of the supports generated by SPARKLE.
For Theorem~\ref{thm:superrorbound} to hold throughout the algorithm, the sample support for each arm must remain $\mathscr{C}$-regular across epochs, despite its adaptive nature.
The RKHS bound gives the control needed to maintain this regularity, allowing our error bounds to apply uniformly over the algorithm's run.

\section{Sparse Additive Regularized Kernel Learning}\label{sec:SPARKLE}

\subsection{Algorithm}\label{sec:algo}

We present SPARKLE in Algorithm~\ref{alg:sparkle}.
The algorithm is simple to implement and has three  components: (i) an epoch-based structure, (ii) a \emph{sequential screening} procedure to identify a candidate set of arms that are likely to be optimal for a given covariate, and (iii) uniform random selection of an arm from this set.

\begin{algorithm}[ht]
\caption{SPARKLE}\label{alg:sparkle}
\begin{algorithmic}[1]
\STATE \textbf{Input}: Epoch $\cT_q$
and error tolerance $\epsilon_q$, for all
$q=1,\ldots,Q$; see \eqref{eq:epoch-length} and \eqref{eq:error-tole}.
\STATE Initialize $\cK_0(\bx) =\{1,\ldots,K\} $ for all $\bx\in\Omega$.
\FOR{$q=1,...,Q$}
\FOR{$t\in \cT_q$}
\STATE Observe covariate vector $\bx_t$ and perform sequential screening:
\FOR{$h =1,\ldots, q-1$ } \label{algo_line:h-loop}
\STATE Compute the candidate set  \label{algo_line:h-loop-end}
\begin{equation}\label{eq:screening}
    \cK_h(\bx_t)=\left\{k\in \cK_{h-1}(\bx_t): \hat{f}_{k,h}(\bx_t)\geq\max_{l \in \cK_{h-1}(\bx_t)}\hat{f}_{l, h}(\bx_t)-\epsilon_h\right\}.
\end{equation}
\ENDFOR
\STATE Pull arm $\pi_t \in \cK_{q-1}(\bx_t)$ uniformly at random, and receive reward $y_t$.
\ENDFOR
\FOR{$k = 1,\ldots, K$}
\STATE Compute the index set $\mathcal{I}_{k, q} = \{t \in \cT_q: \pi_t = k\}$.
\STATE Update the regularization parameters $(\rho_{k,q},\lambda_{k,q})$ based on the sample size $|\mathcal{I}_{k,q}|$; see \eqref{eq:rho-and-lambda}.
\STATE Compute the doubly penalized estimator $\hat{f}_{k, q} = \hat{f}^{\mathsf{DP}}(\mathcal{I}_{k,q}, \rho_{k,q}, \lambda_{k,q})$.
\ENDFOR
\ENDFOR
\end{algorithmic}
\end{algorithm}

In SPARKLE, the time horizon $T$ is divided into a series of epochs\footnote{Throughout the paper, we assume that $T$ is known in advance. However, the algorithm can be adapted to an unknown time horizon  using the ``doubling trick'' \citep{LattimoreSzepesvari2020}.}
$\{\cT_q: q = 1, \ldots, Q\}$,  defined by the time nodes $0 = t_0 < t_1 < \cdots < t_Q = T $, that is, $\cT_q \coloneqq \{t: t_{q-1} < t \leq t_q \}$.
Within each epoch $\cT_q$, the doubly penalized estimator \eqref{eq:DP-estimator} for each arm $k$ is computed once, at the end of the epoch, using only data from that epoch, namely times $\mathcal{I}_{k, q} \coloneqq \{t \in \cT_q: \pi_t = k\}$.
The regularization parameters $(\rho,\lambda)$ are tuned based on $|\mathcal{I}_{k,q}|$, the number of pulls of arm $k$ in epoch $q$.
To highlight its dependence on both the sample and the regularization parameters,
the epoch-$q$ estimator for the reward function $f_k^*$ is written as
$\hat{f}_{k,q} \coloneqq \hat{f}^{\mathsf{DP}}(\mathcal{I}_{k,q}, \rho_{k,q}, \lambda_{k,q})$.

Consider a particular epoch $\cT_q$.
The arm-selection rule is fixed within this epoch and is determined entirely by estimators from \emph{all previous}  epochs. At each time $t \in \cT_q$,
the covariate observation $\bx_t$ is sequentially processed by these estimators,
starting with those from epoch $\cT_1$ and continuing through to epoch $\cT_{q-1}$.
This processing results in multiple rounds of arm screening, one for each epoch.

In round $h\in\{1,...,q-1\}$,
the estimators $\{\hat{f}_{k,h}:k=1,\ldots,K\}$ evaluate the surviving arms from the previous round, denoted by $\cK_{h-1}(\bx_t)$.
As shown by \eqref{eq:screening},
an arm $k$ is kept for the next round if its estimated reward is, up to the tolerance $\epsilon_h$, no smaller than that of every other surviving arm.
After the $q-1$ screening rounds,
an arm is selected uniformly at random from the final candidate set $\cK_{q-1}(\bx_t)$ to be pulled.\footnote{The loop over $h$ can stop early once $|\cK_h(\bx_t)|=1$, so there is no need to run all rounds up to $q-1$.}

SPARKLE has three design parameters: the epoch length $\tau_q \coloneqq |\cT_q|$,
the error tolerance $\epsilon_q$ for arm selection,
and the regularization parameters $(\rho_{k,q},\lambda_{k,q})$.
They are specified as follows.

\begin{itemize}
    \item Epochs. Let $C_1=C_6^{\frac{4m+2}{2m-1}}\vee C_6^{\frac{4m}{2m-1}}$, where $C_6$ is given in \eqref{eq:boundLinftyAq}. Set
    \begin{equation}\label{eq:epoch-length}
    \tau_q = \biggl\lceil C_1 \Bigl(s_0^{\frac{2m+1}{4m}}2^{q+4}\Bigr)^{\frac{4m+2}{2m-1}}\log(dT)\log(T) \biggr\rceil.
    \end{equation}
    With $C_2=(16s_0^{\frac{2m+1}{4m}})^{\frac{4m+2}{2m-1}}C_1$, the total epoch number $Q\leq \lceil{\frac{2m-1}{(4m+2)\log(2)}}(\log \frac{T}{C_2(\log(d)+\log(T))\log(T)})\rceil$.
    \item Error tolerances.
    \begin{align}\label{eq:error-tole}
        \epsilon_q=2^{-q}(\log(T))^{-\frac{2m-1}{4m}}.
    \end{align}
    \item Regularization parameters.
    Set $(\rho_{k,q},\lambda_{k,q})$ according to the convergence rates in Theorem~\ref{thm:superrorbound}, replacing $n$ and $\delta$ there with $\tau_{k,q} \coloneqq |\cI_{k,q}|$ and $T^{-1}$:
    \begin{align}\label{eq:rho-and-lambda}
         \rho_{k,q} =  \lambda_{k,q} w_{k,q} \quad\mbox{and}\quad \lambda_{k,q}= C_3 \Bigl(\gamma_{\tau_{k,q}} + \left(\tau_{k,q}^{-1}\log (d T)\right)^{\frac{1}{2}} \Bigr),
    \end{align}
    where $C_3$ and $C_4$ are  as in Theorem~\ref{thm:superrorbound}, and
    \begin{align*}
        w_{k,q} \coloneqq & {} \Bigl(C_4^{\frac{2m}{2m+1}} \tau_{k,q}^{-\frac{m}{2m+1}}\Bigr) \vee \Bigl(\tau_{k,q}^{-1}\log (d T)\Bigr)^{\frac{1}{2}}, \\
        \gamma_{k,q} \coloneqq & {}  \Bigl(C_4^{ \frac{2m}{2m+1}} \tau_{k,q}^{-\frac{m}{2m+1}} \Bigr) \wedge \Bigr(C_4 \tau_{k,q}^{-\frac{1}{2}}\Bigl(\tau_{k,q}^{-1}\log (d T)\Bigr)^{-\frac{1}{4m}}\Bigr).
    \end{align*}
\end{itemize}

\begin{remark}
Epoch-based designs are common in contextual bandits \citep{foster2018practical,Simchi-LeviXu22} and give SPARKLE both computational and theoretical benefits. On the computational side, updating estimators only at the end of each epoch reduces the number of SOCP solves in \eqref{eq_solvebeta} to $O(\log T)$. On the theoretical side, keeping the arm-selection rule fixed within an epoch removes temporal dependencies that would otherwise arise from continuous policy updates, which simplifies the analysis.

\end{remark}

\subsection{Dynamics of Exploration--Exploitation Trade-off}\label{sec:extrade}

SPARKLE adapts exploration and exploitation through the evolution of candidate arm sets.
When multiple arms remain competitive, exploration continues; when the candidate set becomes a singleton, the policy exploits automatically. For each new context $\bx_t$ in epoch $q$, the screening iterations \eqref{eq:screening} are applied. During the process, the active candidate sets are nested:
$
\{1,\ldots,K\}=\cK_0(\bx)\supseteq \cK_1(\bx)\supseteq \cdots \supseteq \cK_{T-1}(\bx)
$.
Thus the process moves from initial pure exploration ($\cK_0(\bx)=\{1,\ldots,K\}$) to pure exploitation once $|\cK_h(\bx)|=1$, after which screening stops by an absorption property\footnote{This property implies that when computing $\cK_{q-1}(\bx_t)$ (lines \ref{algo_line:h-loop}–\ref{algo_line:h-loop-end} in Algorithm~\ref{alg:sparkle}), the loop over $h$ can terminate as soon as $|\cK_h(\bx_t)|=1$, avoiding further iterations up to $q-1$.}:
\begin{align}\label{eq:nested-2}
    |\cK_h(\bx)|=1\quad\mbox{implies}\quad \cK_h(\bx) = \cK_{h+1}(\bx) = \cdots=\cK_{T-1}(\bx).
\end{align}

This exploration--exploitation flow can be formalized via evolving \emph{exploration region} and \emph{exploitation region}.
For each arm $k$ and epoch $\cT_q$, define
\begin{align}
&\text{(exploration region)}\qquad     \cM_{k,q} \coloneqq \left\{\bx \in\Omega :k\in\cK_{q-1}(\bx), \, |\cK_{q-1}(\bx)|>1  \right\}, \label{eq:exploration-region} \\
&\text{(exploitation region)}\qquad     \cH_{k,q} \coloneqq \{\bx \in\Omega :\cK_{q-1}(\bx)=\{k\}\}. \label{eq:exploitation-region}
\end{align}

The exploration region \eqref{eq:exploration-region} comprises contexts where arm $k$ is statistically indistinguishable from competing arms under the current estimation uncertainty.
By contrast, the exploitation region \eqref{eq:exploitation-region} consists of contexts where the estimated reward of arm $k$ exceeds that of every other arm by at least the prescribed tolerance.
This tolerance, derived from estimator accuracy, sets a safety margin guaranteeing that arm $k$ is highly likely to be optimal.
The dynamics of the exploration--exploitation trade-off is governed by three relationships, stated in Proposition~\ref{prop:HMsupport}.

\begin{proposition}
\label{prop:HMsupport}
    For all $k=1,\ldots,K$ and $q\geq 1$, we have:
    $\text{(1) }\cM_{k,q+1} \subseteq \cM_{k,q},$ $\text{(2) }\cH_{k,q+1} \supseteq \cH_{k,q}$, and $\text{(3) }\cH_{k,q+1} \setminus \cH_{k,q} \subseteq \cM_{k,q}. $
\end{proposition}

\begin{figure}[ht]
\FIGURE{
\includegraphics[width=0.9\textwidth]{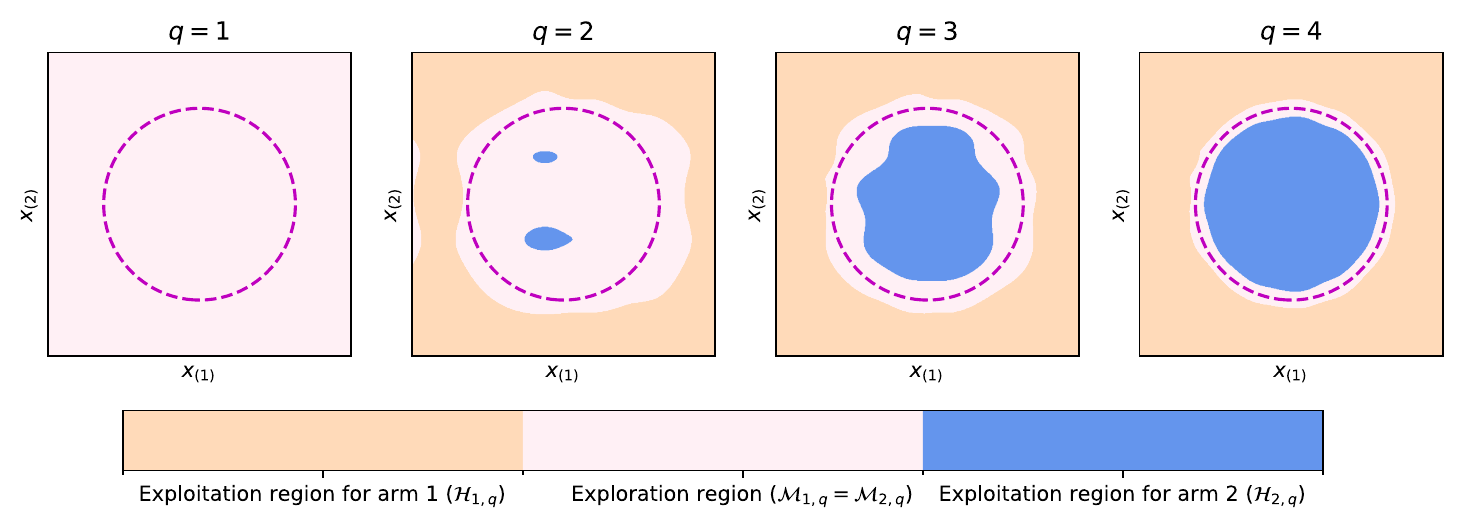}
}
{Evolution of Exploration and Exploitation Regions. \label{fig:evolution} }
{This example involves two arms and two-dimensional covariates. The reward functions are $f_1^*(\bx) = (x_{(1)}^2 + x_{(2)}^2)/10 - 1$ and $f_2^*(\bx) = 0$, where $\bx= (x_{(1)}, x_{(2)})$.
The decision boundary that determines the optimal arm is indicated by a dashed line.
When $K=2$, the exploration regions for both arms are identical by the definition \eqref{eq:exploration-region}, that is, $\cM_{1, q}= \cM_{2, q}$ for all $q\geq 1$. }
\end{figure}

Proposition~\ref{prop:HMsupport} shows that, as the algorithm progresses, the exploration region shrinks while the exploitation region expands, with growth in the latter coming only from the former. Thus exploration regions transition into exploitation regions but never the reverse, yielding a one-way progression: once uncertainty is resolved, it remains resolved, substantially reducing the exploration burden. Although such irreversibility could, in principle, propagate early errors, our uniform accuracy across epochs (Proposition~\ref{prop:Aqhold}) mitigates this risk. Figure~\ref{fig:evolution} illustrates the shift using an examples with two arms.

The reliability of SPARKLE’s exploration--exploitation transition comes from its core idea of contextual difficulty routing: a sequential arm screening procedure that matches estimation precision to how hard it is to tell arms apart in a region. Historical estimators $\hat f_{k,h}$ act as increasingly fine sieves.
In early rounds, coarse sieves quickly resolve easy regions with large reward gaps, enabling safe, immediate exploitation with few samples.
Later, finer sieves focus on hard regions near decision boundaries, keeping exploration only where comparisons remain uncertain.
As a result, exploration effort concentrates where discrimination is most difficult, which uses data efficiently.

This process is reinforced by feedback. When several arms stay active near a boundary, uniform randomization among them directs new samples to those high-uncertainty contexts. The added data improves later estimators, which further sharpens screening where it matters.

If one relied only on the latest estimators, one would need high precision everywhere, which would force broad exploration and lead to high regret.
SPARKLE avoids this by concentrating exploration on critical regions and shrinking the scope of exploration elsewhere.
To keep estimation error low within these localized supports, we need regularity of the sample supports. Sparse additive models can fail if the critical regions are highly fragmented and the overall support violates $\mathscr{C}$-regularity. Section~\ref{sec:sample-supports} introduces the needed geometric properties, and Proposition~\ref{prop:cnstlength} provides the guarantees.

\subsection{Sample Supports}\label{sec:sample-supports}

As shown in Figure \ref{fig:evolution}, the \emph{sample support} of a given arm (i.e., the support of the covariate distribution conditional on that arm being selected) is typically only a subset of the full support $\Omega$.
For epoch $q$, define the covariate support for arm $k$ as
$\cS_{k,q}=\{\bx\in\Omega:\PP(\bx_t=\bx|\pi_t=k,t\in\mathcal{T}_q)>0\}$.
By the selection rule,  $\cS_{k,q}=\cM_{k,q}\cup\cH_{k,q}$.
Corollary \ref{prop:sample-support} collects the main properties of these sample supports induced by SPARKLE.

\begin{corollary} \label{prop:sample-support}
For all $k=1,\ldots,K$ and $q\geq 1$, under Assumption \ref{assum_densitybound},
\begin{align*}
 \text{(1) }\cS_{k,q+1} \subseteq  \cS_{k,q};\quad
 \text{(2) } \bx\in \cS_{k,q} \text{ if and only if }k\in \cK_{q-1}(\bx);\quad
 \text{(3) } \bigcup_{k=1}^K \cS_{k,q} = \Omega.
\end{align*}
\end{corollary}

Statement (1) of Corollary~\ref{prop:sample-support} indicates that the sample support for each arm shrinks across epochs, which is a direct result of Proposition \ref{prop:HMsupport}.
Due to the shrinking property,
the sample support for each arm is generally only part of $\Omega$, except in the first epoch (i.e., $\cS_{k,1} = \Omega$ for all $k$).

A natural concern is extrapolating an estimator beyond its support, which can cause large errors.
Statement (2) of Corollary~\ref{prop:sample-support} rules this out.
At time $t\in\cT_q$, we evaluate $\hat f_{k,h}(\bx_t)$ only for $k\in\cK_{h-1}(\bx_t)$ and $h=1,\ldots,q-1$. Since $k\in\cK_{h-1}(\bx_t)$ if and only if $\bx_t\in\cS_{k,h}$, each such evaluation is not extrapolative, as $\bx_t$ falls within the sample support of the estimator.

The support $\cS_{k,q}$ is not fixed in advance.
It forms stochastically and evolves over time through the algorithm’s interaction with three randomness sources: the random covariate $\bm{x}_t \sim \mathsf{P}_X$, the reward noise $\varepsilon_{k,t}$ and randomized arm selection (uniform sampling from $\mathcal{K}_{q-1}(\bm{x}_t)$).
This randomness propagates recursively because $\mathcal{S}_{k,q}$ is jointly defined by all historical estimators $\{\widehat{f}_{k,h}\}_{h=1}^{q-1}$ and error tolerances $\{\epsilon_h\}_{h=1}^{q-1}$.

Formally, $\cS_{k,q}$ is a nested superlevel set determined by the estimators and error thresholds:
\begin{align}\label{eq:Skqnestedlevelset}
    \mathcal{S}_{k,q} = \bigcap_{h=1}^{q-1} \left\{ \bm{x} \in S_{k,h} : \widehat{f}_{k,h}(\bm{x}) \geq \max_{l \in \mathcal{K}_{h-1}(\bm{x})} \widehat{f}_{l,h}(\bm{x}) - \epsilon_h \right\}.
\end{align}

The nested superlevel set representation in \eqref{eq:Skqnestedlevelset} implies that the sample support  $\mathcal{S}_{k,q}$ contracts across epochs toward the nearly optimal region $\tilde{\mathcal{R}}_k:=\{\bx\in\Omega|\max_{1\leq i\leq K}f^*_i(\bx)-f^*_k(\bx)\leq T^{-\frac{2m-1}{4m}}\}$ (see Figure \ref{fig:evolution}).
Proposition~\ref{prop:Aqhold} in Section \ref{sec_regret} gives uniform sup-norm convergence of  $\widehat{f}_{k,q}$ to $f^*_k$ for all arms and epochs, with error by $1/8 \epsilon_q$.
The error tolerance $\epsilon_q$ in \eqref{eq:error-tole} decays geometrically but dominates the critical scale $O(T^{-\frac{2m-1}{4m}})$.
Hence, $\mathcal{S}_{k,q}$ shrinks over $q$  yet continues to cover $\tilde{\mathcal{R}}_k$.

For this contraction to preserve the $\mathscr{C}$-regularity of the sample supports,
the true nearly optimal region $\tilde{\cR}_k$ must itself satisfy the regularity conditions in Assumption \ref{assum_optprob} as follows.

\begin{assumption}\label{assum_optprob}
Define the nearly optimal region $\tilde{\mathcal{R}}_k=\{\bx\in\Omega|\max_{1\leq i\leq K}f^*_i(\bx)-f^*_k(\bx)\leq T^{-\frac{2m-1}{4m}}\}$.
\begin{enumerate}[label=(\roman*), noitemsep, nolistsep]
\item There exist pairwise disjoint, connected sets $\{\tilde{\mathcal{R}}_{k,l}\}_{l=1}^{n_k}$ that satisfy $\min_{1\leq k\leq K, 1\leq l \leq n_k}\PP
(X\in\tilde{\mathcal{R}}_{k,l})\geq \tilde p$ for some constant $\tilde p>0$, such that
$\tilde{\mathcal{R}}_k = \cup_{l=1}^{n_k} \tilde{\mathcal{R}}_{k,l}$.
\item The projection of $\tilde{\mathcal{R}}_k$ onto the $j$-th dimension has gaps lower bounded by some $\mathfrak{r}_0>0$ for all $j$.
\end{enumerate}

\end{assumption}

Assumption \ref{assum_optprob} encodes the key elements of $\mathscr{C}$-regularity.
The lower bound $\widetilde{p}$ ensures that each connected component $\tilde{\mathcal{R}}_{k,l}$ has enough probability mass,
which in turn implies a minimum length in its coordinate-wise projections.
The gap parameter $\mathfrak{r}_0$
ensures a minimum separation between distinct projected components.

While Assumption \ref{assum_optprob} ensures the global structure of the near-optimal regions $\widetilde{\mathcal{R}}_k$ is well-behaved, it does not fully regulate how the algorithm's sample supports $\mathcal{S}_{k,q}$ evolve adaptively. As $\mathcal{S}_{k,q}$ contracts toward $\widetilde{\mathcal{R}}_k$, local irregularities near decision boundaries could break $\mathscr{C}$-regularity if the reward functions behave pathologically. Specifically, intermediate sample supports must avoid excessive fragmentation that would violate the minimum interval length $\mathscr{C}$, and avoid tiny holes that would violate the minimum gap $\mathscr{C}$ between components. If $\mathscr{C}$-regularity fails, the $L_\infty$ estimation error can become unbounded, breaking the regret analysis and allowing the feedback loop between estimation and data collection to amplify errors.

\begin{figure}[ht]
\FIGURE{
\includegraphics[width=0.7\textwidth]{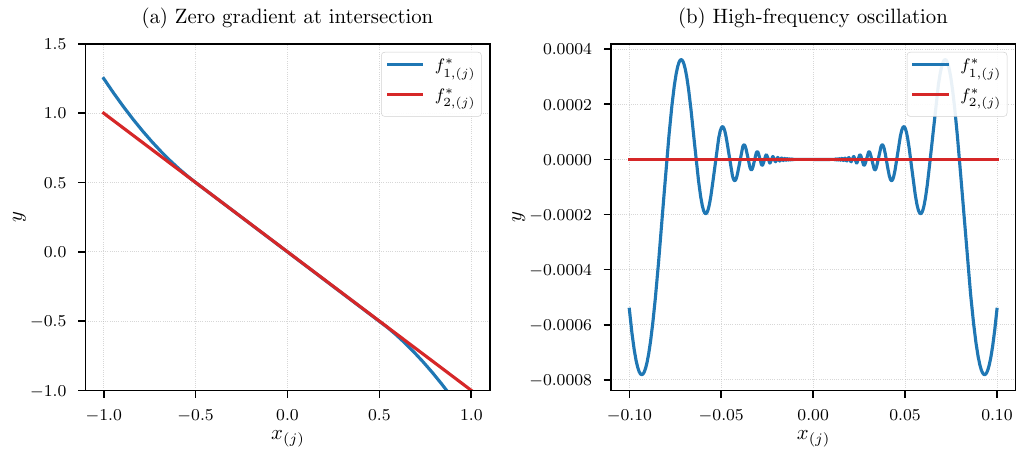}
}
{Pathological Behaviors Precluded by Assumption~\ref{assum:regularity}.
\label{fig:illustrationofassum8} }
{Subfigures (a) and (b) show functions whose behavior near the decision boundaries violates Assumption~\ref{assum:regularity}.}
\end{figure}

To highlight the issues, consider a two-armed setting. We mainly rule out two pathological scenarios. First, abrupt loss of response: arm 1 outperforms arm 2 in some contexts, yet in adjacent neighborhoods their rewards are not only identical but also respond identically to context changes, suggesting a sudden loss of discriminative power on all high-dimensional covariates. Second, erratic oscillation: the identity of the optimal arm flips between the two arms over arbitrarily small neighborhoods.
Figure \ref{fig:illustrationofassum8} visualizes a one‑dimensional component function of these undesirable local behaviors (but noted that our concerned pathological case arises only when all high dimensions simultaneously suffer from such irregularity).

Both are implausible in practice.
Exact equality can occur at isolated points, but it is rare for all high-dimensional context variables to respond identically to both arms at a given point; sharp alternation conflicts with the smooth preference patterns seen in real systems.
To rule out these degeneracies during  learning, we introduce an additional local regularity assumption.

\begin{assumption}\label{assum:regularity}
   Assume $\Omega$ is $\mathscr{C}$-regular. Define
   $\cQ_{\bx} = \argmax_{1\leq k \leq K}f_k^*(\bx)$, the set of arms that attain the maximum reward at covariate $\bx$.
   There exist positive constants $\mathfrak{r}, \tilde c$ and $c^{\prime}$ such that for all $k=1,\ldots,K$:
    \begin{itemize}
        \item[(i)] Define $\cX_0^{(k)} = \{\bx \in \Omega: k\in \cQ_{\bx}, |\cQ_{\bx}|>1\}$ and $B(\cX_0^{(k)})=\bigcup_{\bx\in\cX_0^{(k)}} B(\bx,\mathfrak{r})$, where $B(\bx,\mathfrak{r})$ denotes the ball in $\Omega$ centered at $\bx$ with radius $\mathfrak{r}$. For any $\bx_1\in B(\cX_0^{(k)})\setminus\cX_0^{(k)}$, let $\bx_0$ be the projection of $\bx_1$ on $\cX_0^{(k)}$. Then, for any $i,j \in \cQ_{\bx_0}$,  at least one of the following is true:
        \begin{itemize}
                        \item[$\bullet$] $|f_i^*(\bx_0+\mathfrak{t}(\bx_1-\bx_0))-f_j^*(\bx_0+\mathfrak{t}(\bx_1-\bx_0))|\leq T^{-\frac{2m-1}{4m}}$ for all $0\leq\mathfrak{t}\leq 1$;

            \item[$\bullet$] $\left|\nabla_{\bv}\big(f^*_i(\bx_0+\mathfrak{t}(\bx_1-\bx_0))-  f^*_j(\bx_0+\mathfrak{t}(\bx_1-\bx_0))\big)\right|>\tilde{c}$ ~for all $0\leq\mathfrak{t}\leq 1$, where $\bv=\frac{(\bx_1-\bx_0)}{\|\bx_1-\bx_0\|_2}$.
        \end{itemize}

        \item[(ii)] For all $l\notin \cQ_{\bx_0}$ with $\bx_0 \in \cX_0^{(k)}$ and $\bx_1\in B(\bx_0,\mathfrak{r})$, $\max_{1\leq k\leq K}f_k^*(\bx_1)-f_l^*(\bx_1)> c^\prime$.

    \end{itemize}
\end{assumption}

Statement (i) characterizes the behavior of the competing arms with their transition from a state of ties ($|\cQ_{\bx_0}|>1$) to separation within an $\mathfrak{r}$-neighborhood of the decision boundary. For any point $\bx_1$ near (but not on) the boundary $\mathcal{X}_0^{(k)}$, consider the path to its projection $\bx_0$: the difference between any two competing arms must either remain uniformly small ($\leq T^{-\frac{2m-1}{4m}}$), preserving support connectivity, or exhibit a separation rate bounded from below ($|\nabla{\bv}(f_i^* - f_j^*)| > \tilde{c}$), indicating distinct directional responses. Statement (ii) enforces local competitor set invariance near decision boundaries. In $\mathfrak{r}$-neighborhoods of decision boundaries, only arms optimal at ${\bm{x}_0}$ can compete for optimality.

Assumption \ref{assum:regularity} regulates the geometry of the sample supports $\mathcal{S}_{k,q}$ by imposing local conditions on the reward functions near their decision boundaries.
The assumption avoids global restrictions and instead uses a flexible ``small difference or noticeable gradient'' requirement that rules out pathological cases.
These conditions are verifiable from the problem primitives and do not involve on sample-dependent randomness.
This local geometric characterization is key to our analysis of the stochastic behavior of $\mathcal{S}_{k,q}$.

Although Assumption \ref{assum:regularity} may appear complex, it can be guaranteed by simpler sufficient conditions.
Lemma \ref{lemma:regularity} provides such a condition for the two-armed case.

\begin{lemma}\label{lemma:regularity}
Suppose $K=2$ and Assumption \ref{assump:kernel} holds. If either $|f^*_1(\bx)-f^*_2(\bx)|\leq T^{-\frac{2m-1}{4m}}$ for all $\bx\in\Omega$, or  $\|\nabla f^*_1(\bx)- \nabla f^*_2(\bx)\|_2>2\tilde{c}$ for all $\bx\in \cX_0^{(1)}$
and some constant $\tilde{c} > 0$,
then Assumption \ref{assum:regularity} holds.
\end{lemma}

The gradient condition in Lemma \ref{lemma:regularity} is mild, as it only excludes the
set of functions exhibiting perfect tangency at their intersection.
Such tangency is atypical in high dimensions, as it would require not only $f_1^*(\bx_0)=f_2^*(\bx_0)$ but also $\nabla f_1^*(\bx_0)=\nabla f_2^*(\bx_0)$ across all $d$ dimensions.
In practice, this gradient condition is easy to satisfy.
For example, if even one single covariate affects the two reward functions differently at $\bx_0$,
which is a common situation when the rewards depend on different sparse covariate sets, then the gradients differ and the condition holds.

We stress that Assumptions~\ref{assum_optprob} and~\ref{assum:regularity} are not just for analytical convenience; they are needed to make the problem well posed.
Relaxing them changes the problem class and makes learning substantially harder.
We make this precise in Theorem~\ref{thm:lbwithoutregularity}
in Appendix~\ref{sec:price}.
It shows that any non-anticipating policy has regret that is asymptotically much larger than the upper bound we prove under the assumptions. These conditions exclude pathological cases that are implausible in practice and are key to efficient learning.

At a high level, each arm's sample support under SPARKLE adapts to the nonparametric reward estimate, allowing exploitation in well-separated regions and sustained exploration near decision boundaries.
Analyzing this requires new technical tools, since we must track the geometry of time-varying supports in high dimensions and control error propagation across epochs.
Our approach uses uniform control of estimation errors that captures both value deviation and smoothness deviation between the estimator and the true reward, via RKHS methods. This links high-dimensional nonparametric statistics in RKHS with contextual bandits to analyze nonstationary support geometry and its interaction with sequential decision-making.

\section{Regret Analysis}\label{sec_regret}

In Section~\ref{subsubsec_rde},
we present a regret decomposition based on whether the covariate $\bx_t$ lies in the exploration or exploitation regions defined in Section~\ref{sec:extrade}.
We state additional technical assumptions for the regret analysis
in Section~\ref{subsec:more-assump} and prove uniform convergence of the doubly penalized reward estimators across all epochs in Section~\ref{subsubsec_est}.
We then derive an upper bound on SPARKLE's expected cumulative regret in Section~\ref{subsubsec_final} and complement it with information-theoretic lower bounds for alternative algorithms in Section~\ref{sec:regret_lb}.

\subsection{Regret Decomposition}\label{subsubsec_rde}

Let $R_T$ denote the expected cumulative regret of SPARKLE.
Our analysis begins with a decomposition of $R_T$ that matches the epoch structure and the sequential screening in SPARKLE.
At any time $t$ in epoch $q$ (i.e., $t \in \mathcal{T}_q$), the algorithm selects an arm $\pi_t$ through a series of screening rounds.
The selection rule implies that the covariate $\bx_t$  lies in either an exploration region or an exploitation region relative to the selected arm $\pi_t$.
We therefore decompose $R_T$ by region and by epoch:
\begin{align}\label{eq:Rt}
    R_T&=\sum_{q=1}^{Q}\sum_{t\in \cT_q}\EE(\max_{k}f_k^*\left(\bx_t\right)-f_{\pi_t}^*\left(\bx_t\right))\nonumber\\
    &=\underbrace{\sum_{t\in \cT_1}\EE(\max_{k}f_k^*\left(\bx_t\right)-f_{\pi_t}^*\left(\bx_t\right))}_{I_1}+\underbrace{\sum_{q=2}^{Q}\sum_{t\in \cT_q}\EE\left((\max_{k}f_k^*\left(\bx_t\right)-f_{\pi_t}^*\left(\bx_t\right))\II(\bx_t\in\cH_{\pi_t, q})\right)}_{I_2}\nonumber\\
    &+\underbrace{\sum_{q=2}^{Q}\sum_{t\in \cT_q}\EE\left((\max_{k}f_k^*\left(\bx_t\right)-f_{\pi_t}^*\left(\bx_t\right))\II(\bx_t\in\cM_{\pi_t,q})\right)}_{I_3}.
\end{align}

In the decomposition above, $I_1$ is the regret in the initial epoch $\cT_1$, which is a phase of pure exploration over the entire covariate space $\Omega$ with no historical data.
$I_2$ is the regret when the algorithm exploits, i.e., when $\bx_t$ lies in the exploitation region $\cH_{\pi_t,q}$ of the selected arm $\pi_t$. By contrast, $I_3$ is the regret when the algorithm explores, i.e., when $\bx_t$ lies in the exploration region $\cM_{\pi_t,q}$ of the selected arm.
The three parts are exhaustive and mutually exclusive, and the regions are disjoint by construction.

Since $f_k^*\in \cF$ and $\cF_j\subset \cN_{\Phi}([0,1])$ for all $k = 1,\ldots,K$ and $j=1,...,d$, where $\cF$ and $\cF_j$ are defined in~\eqref{eq_functionclass}, the reproducing property of RKHSs implies that for any $\bx\in \Omega$,
\begin{align*}
    |f_k^*(\bx)| = & \left|\sum_{j=1}^{d} f^*_{k, (j)}\left(x_{j}\right)\right|
    = \left|\sum_{j=1}^{d} \langle f^*_{k, (j)}, \Phi(x_{j} - \cdot)\rangle_{\cN_{\Phi}([0,1])} \right|\nonumber\\
    \leq & \sum_{j=1}^{d} \|f^*_{k, (j)}\|_{\cN_{\Phi}([0,1])} \|\Phi(x_{j} - x_{j} )\|_{\cN_{\Phi}([0,1])} \leq s_0 \Phi(0) \max_{1\leq k\leq K, 1\leq j\leq d} \|f^*_{k, (j)}\|_{\cN_{\Phi}([0,1])},
\end{align*}
where the first equality uses \eqref{eq_addreward}, the first inequality is due to the Cauchy--Schwarz inequality, and the last inequality holds because  the sparsity of $f_k^*$ is upper bounded by $s_0$. Therefore, $\|f_k^*\|_{L_\infty(\Omega)} \leq  C_{12}s_0$, for a constant $C_{12}>0$ independent of $k$ and $s_0$.

By the definition of $\tau_1$ (the length of epoch 1), the first term $I_1$ in \eqref{eq:Rt}  satisfies
\begin{align}\label{eq_regret_I1'}
    I_1 \leq 2\tau_1\max _{1\leq k\leq K}\|f_k^*\|_{L_\infty(\Omega)}\leq 2C_1C_{12}s_0 \left(32s_0^{\frac{2m+1}{4m}}\right)^{\frac{4m+2}{2m-1}}(\log(d)+\log(T))\log(T).
\end{align}

In the subsequent sections, we bound the more involved terms $I_2$ and $I_3$, which account for most of the regret.
The analysis hinges on uniform accuracy of the estimators $\hat f_{k,q}$ across epochs and on geometric properties of the sample supports.

\subsection{Additional Assumptions}\label{subsec:more-assump}

We introduce two assumptions from the nonparametric bandit literature to facilitate the regret analysis.

\begin{assumption}\label{assum_optimalregionprob}
    Each arm $k = 1,\ldots,K$ has a corresponding optimal region $\mathcal{R}_k=\{\bx\in\Omega|f^*_k(\bx)= \max_{1\leq i\leq K}f^*_i(\bx)\}$ with $\min_{1 \leq k \leq K}\PP(X\in\mathcal{R}_k)\geq p^* > 0$.
\end{assumption}

Assumption~\ref{assum_optimalregionprob} imposes a positive lower bound on the probability mass of the region where each arm is optimal.
This ensures that each arm receives enough data in each epoch, avoiding data scarcity.
Adequate samples are needed to estimate all reward functions accurately across epochs, which in turn is crucial for the screening procedure; see Proposition~\ref{prop:sfd}.

\begin{assumption}[Margin Condition]\label{assum_margin}
    There exist constants $C_0>0$, $0\leq \alpha\leq 1$, and $0\leq\delta_0\leq T^{-\frac{2m-1}{4m}}$ such that for any $\delta \geq \delta_0$ and all $i$ and $j$ in $\cK$ with $i\neq j$,  $\PP(0<|f_i^*(X)-f_j^*(X)|\leq \delta)\leq C_0\delta^\alpha$.
\end{assumption}

This variant of  Tsybakov's margin condition \citep{Tsybakov04} controls the covariate density near decision boundaries.
The threshold $\delta_0 \le T^{-\frac{2m-1}{4m}}$ reflects the critical scale for nonparametric estimation; smaller $\delta_0$ gives tighter probability control, and $\delta_0=0$ recovers the standard condition.
The parameter $\alpha\in[0,1]$ measures problem difficulty: larger $\alpha$ means faster decay in the probability of small reward gaps, so optimal arms are easier to separate. We restrict to $\alpha\le 1$ because $\alpha>1$ would imply the reward functions coincide almost everywhere, yielding trivial zero regret as shown by Lemma~\ref{lemma:alphasmallerthan1}.

\begin{lemma}\label{lemma:alphasmallerthan1}
    Suppose Assumptions \ref{assum_densitybound}, \ref{assump:kernel}, and \ref{assum_optimalregionprob} hold. In addition, suppose Assumption \ref{assum_margin} holds with $\alpha>1$. Then, $f_k^*(\bx)=f_j^*(\bx)$ for any $k,j=1,\ldots,K$.
\end{lemma}

Intuitively, when $m>3/2$, the smoothness of the reward functions prevents the pairwise differences from changing too sharply near decision boundaries, which favors smaller $\alpha$. See Section~2.7 of \cite{hu2022smooth} for a related discussion.

\subsection{Reward Function Estimation}\label{subsubsec_est}

The efficacy of SPARKLE hinges on the accuracy of the doubly penalized estimators $\hat f_{k,q}$ constructed at the end of each epoch $q$.
We now present results that ensure this accuracy throughout the algorithm.
These results connect the offline estimation theory from Theorem~\ref{thm:superrorbound} with the online data collection, ensuring that the sample supports $S_{k,q}$ have favorable geometry and contain enough data for reliable estimation.

We introduce a ``good event'' $\mathcal{A}_q$ to formalize two requirements at the end of epoch $q$: accurate reward estimators and well-behaved sample supports.

\begin{definition}[Event $\mathcal{A}_{q}$]\label{def_event_A}
For each epoch $q$,  $\mathcal{A}_{q}$ is the event that
(i) $\cS_{k,q}$ is $\mathscr{C}$-regular for all $k=1,\ldots,K$; and (ii)
the following three bounds hold for
some constants $C_6,C_7,C_8>0$ independent of $s$, $d$ and $T$:
\begin{align}
  \max_{1\leq k\leq K}\|\hat{f}_{k,q}-f_k^*\|_{L_\infty(\cS_{k,q})}\leq{}& C_6s_0^{\frac{2m+1}{4m}}\left(\tau_{q}^{-\frac{m}{2m+1}}+\sqrt{\frac{\log(d)+\log(T)}{ \tau_{q}}}\right)^{1-\frac{1}{2m}}\leq \frac{1}{8}\epsilon_q,\label{eq:boundLinftyAq}\\
     \max_{1\leq k\leq K}\|\hat{f}_{k,q} - f^*_k\|_{L_2(\cS_{k,q})}
    \leq{}& C_7 s_0^{\frac{1}{2}}\left(\tau_{q}^{-\frac{m}{2m+1}}+\sqrt{\frac{\log(d)+\log(T)}{ \tau_{q}}}\right),\nonumber\\
    \max_{1\leq k\leq K}\|\hat{f}_{k,q} -f^*_k\|_{\mathcal{N}_\Phi(\cS_{k,q})} \leq{}& C_8 s_0\nonumber.
\end{align}
\end{definition}

The second inequality in \eqref{eq:boundLinftyAq} holds by construction.
For completeness, let $\mathcal{A}_0$ be the sure event.
The sequential screening of
SPARKLE uses all past estimators, so performance in epoch $q$ depends on earlier epochs.
In particular, accuracy in epoch $q$ relies on occurrence of $\mathcal{A}_1,\ldots,\mathcal{A}_{q-1}$.
Conditioned on this history of good events, the data collected in epoch $q$ retains the statistical properties needed to ensure that $\mathcal{A}_q$ holds with high probability.

\begin{proposition}[Sufficient Data]\label{prop:sfd}
Suppose Assumption \ref{assum_optimalregionprob} holds and the event $\bigcap_{1 \leq h \leq q-1} \mathcal{A}_h$ occurs. Then,
\begin{align}
    \PP\left(\tau_{k,q}\geq \frac{p^*}{2K}\tau_q\bigg|\{\cD_{h}\}_{h=1}^{q-1}\right)\geq 1-e^{-\frac{\tau_q(p^*)^2}{2K^2}},
\end{align}
for all $k=1,\ldots,K$,
where $\tau_{k,q} = |\cI_{k,q}|$ is the number of covariate observations assigned to arm $k$ during epoch $q$,   $\tau_q$ is the length of epoch $q$,
and $\cD_{q}\coloneqq \{(\bx_i,y_i) :i\in\mathcal{T}_{q}\}$ represents the data from epoch $q$.
\end{proposition}

Proposition~\ref{prop:sfd} implies that, conditional on past estimators performing well, each arm receives a sample size that is at least a constant fraction of the epoch length with high probability. In other words, every arm accumulates a sufficient number of samples.

\begin{proposition}[Well-Behaved Distribution]\label{prop:dst&dist}
Suppose Assumptions \ref{assum_densitybound} and \ref{assum_optimalregionprob} hold.
Let $\cD_{k,q}\coloneqq \{(\bx_i,y_i) :i\in\mathcal{I}_{k,q}\}$  denote the data received by arm $k$ during epoch $q$.
Then,
$\cD_{k,q}$ consists of conditionally i.i.d. samples given the history $\{\cD_{h}\}_{h=1}^{q-1}$.
Moreover, let $\mu_k^{(q)}(\bx)$ denote the marginal density of any of $\{\bx_t:t\in\cT_{q}\}$, conditional on $\{\cD_{h}\}_{h=1}^{q-1}$, $\bigcap_{1 \leq h \leq q-1} \mathcal{A}_h$, and $\pi_t=k$.
Then,
\begin{align*}
\begin{cases}
     \displaystyle \frac{1}{K} p_{\min } \leq \mu_k^{(q)}(\bx) \leq {\frac{K}{p^*} p_{\max }}, & \mbox{ if } \bx\in \cS_{k,q}, \\[1em]
    \mu_k^{(q)}(\bx) = 0, & \mbox{ if }\bx\in \Omega\setminus\cS_{k,q}.
\end{cases}
\end{align*}
\end{proposition}

This proposition characterizes the epoch‑$q$ data through two key properties: conditional independence and distributional regularity.
The conditional i.i.d. structure arises because the arm selection rule is fixed within the epoch, and uniform randomization over the candidate set $\mathcal{K}_{q-1}(\bx_t)$ ensures that the conditional distribution of covariates assigned to each arm remains well behaved.
These properties are essential for applying our offline estimation theory within the online learning procedure.

\begin{proposition}\label{prop:cnstlength}
     Suppose Assumptions \ref{assump:kernel}, \ref{assum:regularity}, and  \ref{assum_optimalregionprob} hold. Under the event $\bigcap_{1 \leq h \leq q-1} \mathcal{A}_h$,  the sample support $\cS_{k,q}$ is $\mathscr{C}$-regular for all $k=1,\ldots,K$ whenever $T>C$ for some constant $C>0$.
\end{proposition}

As discussed in Section~\ref{sec:sample-supports}, the covariate support contracts and fragments across epochs, which would otherwise complicate the analysis of the reward estimators. Proposition~\ref{prop:cnstlength} addresses this by leveraging the accuracy of past estimators and the arm selection rule to show that all sample supports remain $\mathscr{C}$-regular.
This persistent regularity provides the geometric foundation needed to construct accurate estimators on these adaptively generated domains.

Taken together, Propositions~\ref{prop:sfd} and \ref{prop:dst&dist} and Proposition~\ref{prop:cnstlength} provide a rigorous foundation for reward function estimation at each epoch: the doubly penalized estimators are trained on sufficient samples drawn from a well-behaved distribution supported on a $\mathscr{C}$-regular set.
This ensures that, with high probability, the estimators attain the accuracy guarantees formalized in Proposition~\ref{prop:Aqhold}.

\begin{proposition}\label{prop:Aqhold}
        Suppose Assumptions \ref{assum_densitybound}--\ref{assum_optimalregionprob} hold, and $T>e^{1\vee (1/C_9)}\vee C_5$, where $C_9=\frac{p^{*2}}{2K^2}C_2$.
    Then, for all $1\leq q\leq Q-1$,
    \begin{align}\label{eq_thmaq_1}
        \PP\left(\mathcal{A}_{q+1}|\bigcap_{1 \leq h \leq q} \mathcal{A}_h\right)\geq 1-\frac{3K}{T} \quad \mbox{ and }\quad \PP\left(\bigcap_{1 \leq h \leq q} \mathcal{A}_h\right)\geq 1-q\frac{3K}{T}.
    \end{align}
\end{proposition}

Proposition~\ref{prop:Aqhold} is central to our regret analysis. It ensures that the estimators are uniformly accurate across all epochs with high probability. For the base case $q=1$, the initial support $\mathcal{S}_{k,1}=\Omega$ is $\mathscr{C}$-regular, and the i.i.d. samples (by Assumption~\ref{assum_densitybound}) allow a direct application of Theorem~\ref{thm:superrorbound}. The proof proceeds by induction on $q$, showing that accurate estimation and well-behaved data carry over from one epoch to the next.

\subsection{Regret Upper Bound}\label{subsubsec_final}

With the uniform estimation accuracy from Proposition~\ref{prop:Aqhold}, we now bound the regret terms $I_2$ (exploitation) and $I_3$ (exploration) from the decomposition in Section~\ref{subsubsec_rde}.

The algorithm switches to exploitation (selects a single arm) for a covariate $\bx_t$ only when past estimators indicate that one arm is clearly better than all others in the candidate set.
Under the good event that these estimators are accurate (i.e., $\bigcap_{l=1}^h \mathcal{A}_l$), Lemma~\ref{prop:armioptimal} confirms that the chosen arm is the true optimum, so the instantaneous regret is zero.

\begin{lemma}\label{prop:armioptimal}
Fix an arbitrary epoch $q$.
If the event $\bigcap_{1 \leq h \leq q-2} \mathcal{A}_h$ occurs, then $\argmax_{1\leq j \leq K}f^*_j(\bx_t)\in \cK_{q-2}(\bx_t)$.
Moreover, if the event $\bigcap_{1 \leq h \leq q-1} \mathcal{A}_h$ occurs, $k\in\cK_{q-2}(\bx_t)$, and $\hat{f}_{k,q-1}(\bx_t)-\hat{f}_{j, q-1}(\bx_t)>\epsilon_{q-1}$ for all $j\in \cK_{q-2}(\bx_t)\setminus \{k\}$, then arm $k$ is optimal.
\end{lemma}

Exploitation regret can occur only if at least one earlier estimator fails.
The total exploitation regret $I_2$ is therefore controlled by the sum of these failure probabilities over time, which Proposition~\ref{prop:Aqhold} bounds. A direct calculation shows this sum grows only logarithmically in $T$:
\begin{align}\label{eq_regret_I2}
    I_2 \lesssim \sum_{q=2}^{Q}\sum_{t\in \cT_q}\PP\left(\bigcup_{1 \leq h \leq q-1} \mathcal{A}_h^\complement\right)= O(s_0 \log T),
\end{align}
which is negligible relative to the main regret source.

On the other hand, exploration regret arises when the algorithm randomizes among multiple candidate arms.
We bound $I_3$ using two facts: (i) the per-round regret is small because candidate arms have estimated rewards within $\epsilon_{q-1}$, and under the event $\bigcap_{h=1}^{q-1}\mathcal{A}h$ their true rewards are also within a constant multiple of $\epsilon{q-1}$; (ii) the exploration frequency is controlled by the margin condition (Assumption~\ref{assum_margin}), which bounds the probability of covariates with optimality gap below $\delta$ by $\delta^\alpha$.
Hence, $I_3$ depends on the exploration frequency, the per-round regret, and the epoch parameters. Substituting the geometric schedules for $\tau_q$ and $\epsilon_q$ gives:
\begin{align}\label{eq_regret_I3}
   I_3 \lesssim \sum_{q=2}^{Q}\tau_q\left(\frac{5}{4}\epsilon_{q-1}\right)^{1+\alpha}= \widetilde{O}\left( s_0^{1 + \frac{6m+1}{2m(2m-1)}} T^{1 - \frac{(2m-1)(1+\alpha)}{4m+2}} \log(d) \right).
\end{align}

Combining the bounds for the initial exploration $I_1 = \widetilde{O}(1)$ in \eqref{eq_regret_I1'}, the negligible exploitation regret $I_2$ in \eqref{eq_regret_I2}, and the dominating exploration regret $I_3$ in \eqref{eq_regret_I3}, we arrive at the main theorem of this paper.

\begin{theorem}\label{thm:regretupperbound}
   Suppose Assumptions \ref{assum_densitybound}--\ref{assum_margin} hold. Then, the expected cumulative regret of SPARKLE satisfies
    \begin{align*}
        R_T=\tilde O\left(s_0^{1 + \frac{6m+1}{2m(2m-1)}}T^{1-\frac{(2m-1)(1+\alpha)}{4m+2}}\log(d)\right).
    \end{align*}
\end{theorem}

This is the first sublinear regret bound\footnote{The bound is sublinear in $T$ because $m>3/2$ and $\alpha\in[0,1]$ so the exponent of $T$ is between 0 and 1.} for nonparametric contextual bandits that  depends only logarithmically on $d$, improving on prior results that suffer from the curse of dimensionality \citep{gur2022smoothness,hu2022smooth}.
The bound clarifies how regret depends on key problem characteristics: the smoothness $m$ governs estimation efficiency, and the margin parameter $\alpha$ controls the probability mass of near‑optimal regions. Note that $s_0$ is a constant governing the sparsity level of all arms that can be omitted in the asymptotic analysis. Its exponent decays to 1 as the smoothness parameter $m$ increases, matching the optimal linear dependency on $s_0$ attained in the limit $m \to \infty$ \citep{bastani2020online}.
In the same limit, our rate matches the best-known results for high-dimensional linear bandits under general margin conditions \citep{bastani2020online,QianIngLiu23}.
This is natural because the RKHS with $m\to\infty$ includes linear functions while allowing richer smooth classes.
The result thus unifies parametric and nonparametric regimes and shows that SPARKLE adapts across settings while keeping a logarithmic dependence on $d$.

\subsection{Regret Lower Bounds}\label{sec:regret_lb}

To complement the upper bound, we prove information-theoretic lower bounds for RKHS contextual bandits.
This setting poses two challenges compared with H\"older class framework that is standard in the nonparametric bandit literature \citep{audibert2007fast,rigollet2010nonparametric,hu2022smooth}.
First, the global nature of the RKHS norm precludes the infinitely many ``bump function'' constructions that work for H\"older classes (as long as each remains sufficiently smooth).
Second, standard hard problem instances for H\"older classes involve fragmented supports, which violate our regularity assumption on the sample supports.

\begin{theorem}\label{thm_lb} Fix $\alpha>0$ and $m>0$ with $\alpha m\le 1$. For any non-anticipating policy $\pi$, there exists a contextual bandit instance with horizon $T$ that satisfies Assumptions \ref{assum_densitybound}--\ref{assum_margin} such that, for some constant $C>0$ depending only on the instance class (not on $\pi$),
\begin{align*}
     R_T(\pi) \geq C T^{1-\frac{m(1+\alpha)}{2m+1}}.
\end{align*}
\end{theorem}

This lower bound is close to our upper bound in terms of the horizon $T$, with a gap of order $\tilde{O}(T^{\frac{1+\alpha}{4m+2}})$.
This gap is inherent to the RKHS norm, which measures function smoothness globally rather than locally.
Nevertheless, this gap is small and vanishes as the smoothness parameter $m \to \infty$.
We conjecture it can be closed with a sharper approximation argument and leave this for future work.

To handle high dimensionality, we use an additive structure, which yields a favorable $\log(d)$ dependence in the regret.
Without additivity, the problem suffers from the curse of dimensionality.
Theorem~\ref{thm_lb_withd} formalizes this by providing a dimension-dependent lower bound for general RKHS functions.
In particular, dropping additivity changes the critical value  in Assumption \ref{assump:kernel}, \ref{assum_optprob}, \ref{assum:regularity}, \ref{assum_margin} from $T^{-\frac{2m-1}{4m}}$ to $T^{-\frac{2m-d}{4m}}$ in Assumption \ref{new_assump:kernel}, \ref{new_assum_optprob}, \ref{new_assum:regularity}, \ref{new_assum_margin},
because the former set applies to the one-dimensional RKHS governing each univariate component under additivity, whereas the latter applies to the full $d$-dimensional RKHS without additivity. This latter set is weaker; see Appendix~\ref{sec:modified-assumptions}.

\begin{theorem}\label{thm_lb_withd}
Fix $\alpha>0$ and $m>0$ with $\alpha m\le d$. For any non-anticipating policy $\pi$, there exists a contextual bandit instance with horizon $T$ that satisfies Assumptions \ref{assum_densitybound}, \ref{assum_subG}, \ref{new_assump:kernel}, \ref{assum_compatibility}, \ref{new_assum_optprob},
\ref{new_assum:regularity},\ref{assum_optimalregionprob}, \ref{new_assum_margin} such that, for some constant $C>0$  depending only on the instance class (not on $\pi$),
\begin{align*}
     R_T(\pi) \geq C T^{1-\frac{m(1+\alpha)}{2m+d}}.
\end{align*}
\end{theorem}

The gap between the lower bounds in Theorems \ref{thm_lb} and \ref{thm_lb_withd} quantifies the benefit of sparse additive modeling for mitigating the curse of dimensionality in contextual bandits.
Moreover, our hard instances also lie in the H\"older class $\cC^{r,m-r}(\Omega)$, yielding a byproduct proof of the lower bound for nonparametric contextual bandits with H\"older reward functions and without additivity,
where $r = \lfloor m \rfloor$ is the integer part of $m$ and $ m - r$ is the fractional part.
This complements and aligns with \citet{hu2022smooth}, despite differences in function class specifications.
Taken together, these lower bounds give a more complete theoretical picture: they specify fundamental performance limits and highlight how additivity mitigates the curse of dimensionality.

\section{Numerical Experiments}\label{sec_numexp}

In this section, we compare SPARKLE's performance with several representative contextual bandit algorithms across multiple settings. We present two sets of results: evaluations on sparse synthetic data (Section~\ref{sec:syn-data}) and two real‑world applications (video recommendation and personalized medicine) using real data (Section~\ref{sec:real-data}).

\subsection{Synthetic Data}\label{sec:syn-data}

We use sparse additive reward functions with dimensionality $d$ and sparsity level $s$ in a two-armed setting ($K=2$). For arm $k$,
\[
f^*_k(\bx)=\sum_{j=1}^s 2\sin(x_{(j+k)}).
\]
where $\bx = (x_{(1)},...,x_{(d)})$.
At each time $t$, the covariate
$\bx_t=(x_{t,(1)},\ldots,x_{t,(d)})$ has independent entries $x_{t,(j)}$ drawn uniformly from $(-5,5)$ for $j=1,\ldots,d$.
The observation noise $\varepsilon_{k,t}$ has the normal distribution with mean zero and variance $0.05$ for all $k$ and $t$.
Additional details of the experimental design are provided in Section~\ref{ec-sec:exp-details} of the e‑companion.

\begin{figure}[ht]
\FIGURE{
\begin{tabular}{ccc}
    \subfigure[$s=2,T=500$]{\includegraphics[width=0.3\textwidth]{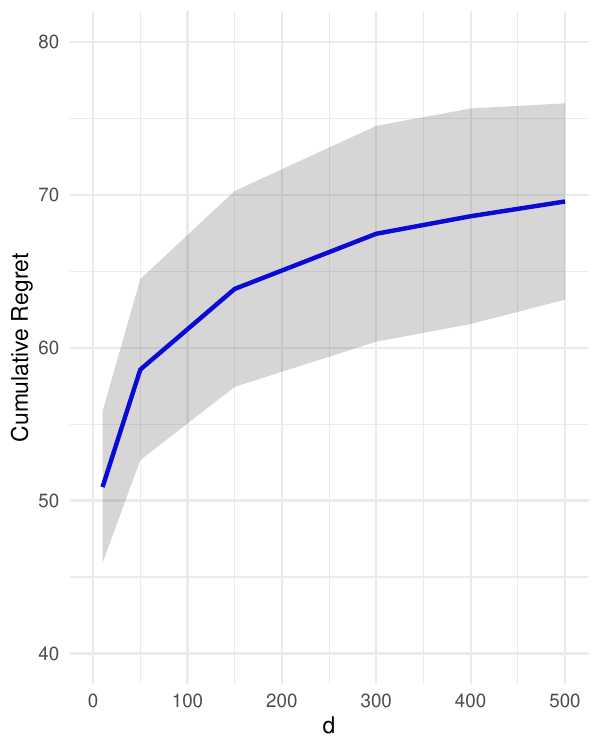}}
\subfigure[$s=2,d=20$]{\includegraphics[width=0.3\textwidth]{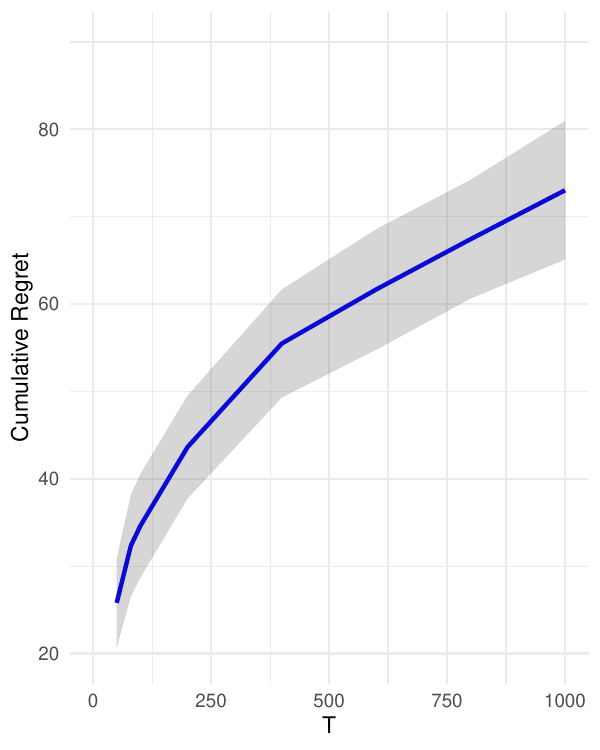}}
\subfigure[$d=20,T=1500$]{\includegraphics[width=0.3\textwidth]{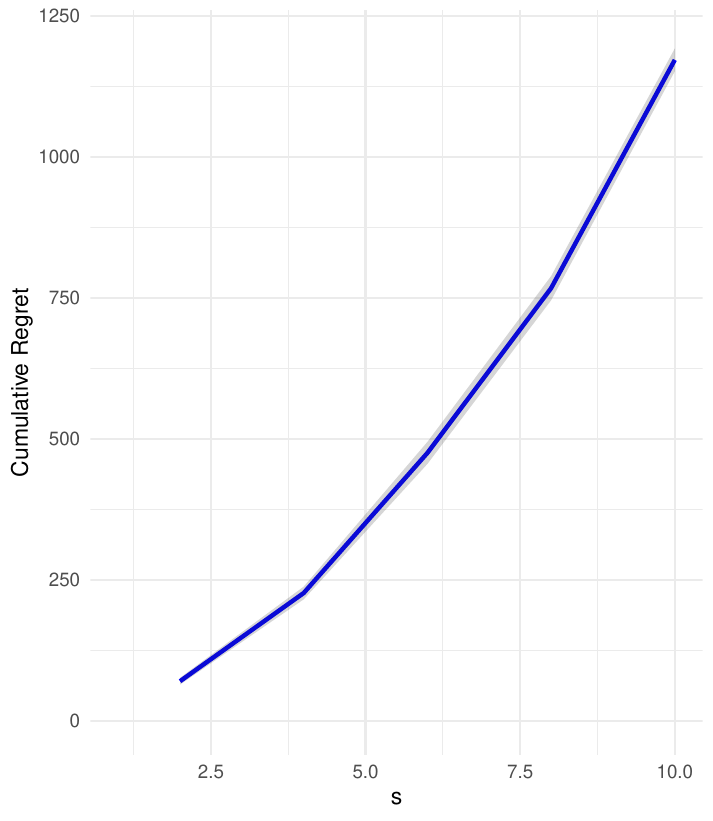}}
\end{tabular}
}
{Effect of $(d, T, s)$ on SPARKLE's Cumulative Regret. \label{fig:dependency}}
{Each setting is repeated 10 times. Shaded bands show 95\% confidence intervals.}
\end{figure}

We first study how SPARKLE’s cumulative regret depends on three parameters: the dimensionality $d$, the horizon $T$, and the sparsity level $s$.
We vary one parameter at a time while fixing the other two, and compute the cumulative regret.
From the regret upper bound in Theorem~\ref{thm:regretupperbound}, we expect logarithmic dependence on $d$, sublinear dependence on $T$, and geometric dependence on $s$.
As shown in Figure~\ref{fig:dependency}, the trend in $d$ is clearly logarithmic, supporting the method's efficiency in high-dimensional settings.
The trends for $T$ and $s$ also agree with the theory.

\begin{figure}[ht]
\FIGURE{
\begin{tabular}{cc}
    \subfigure[$K=2,s=2,d=20$]{\includegraphics[width=0.4\textwidth]{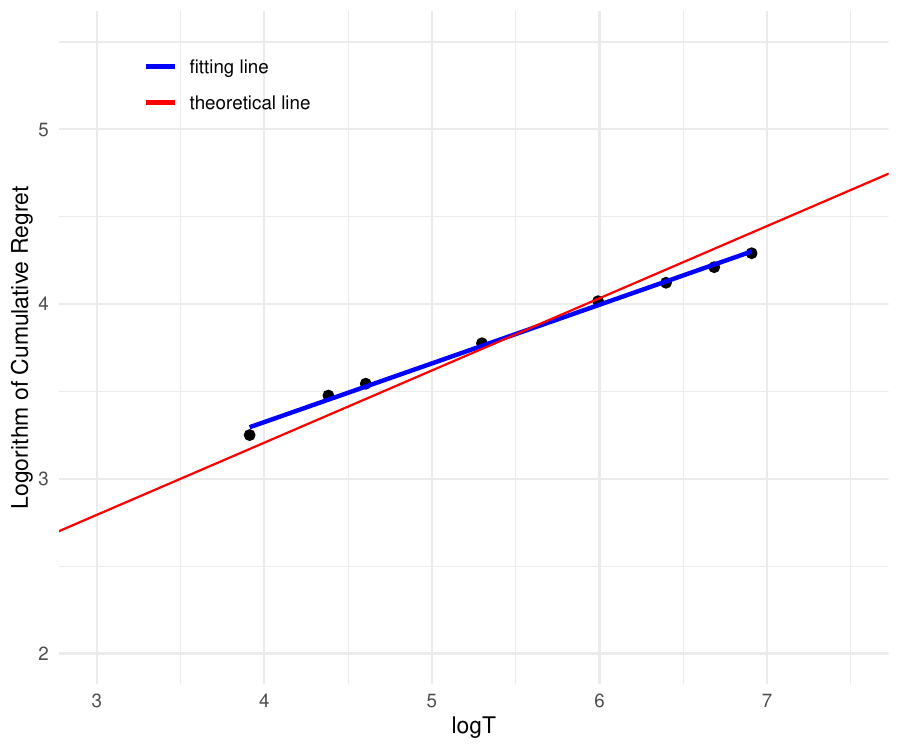}}
\subfigure[$K=2, d=20, T=1500$]{\includegraphics[width=0.4\textwidth]{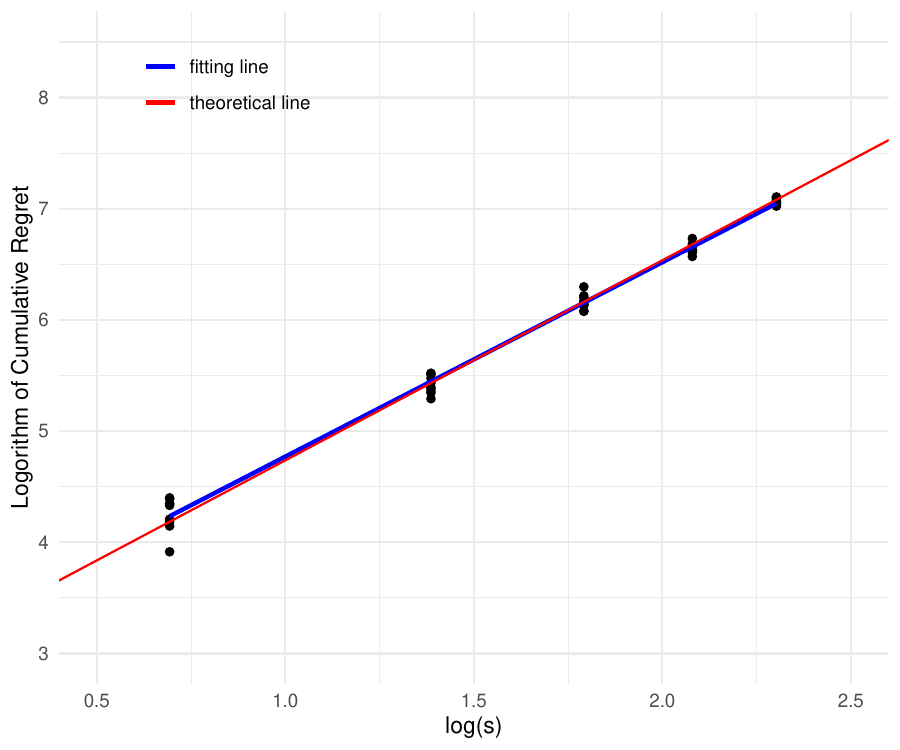}}
\end{tabular}
}
{Dependence of SPARKLE's Cumulative Regret on $(T, s)$ on the Log–Log Scale. \label{fig:fitting_line} }
{Each setting is repeated 10 times.  The fitted lines come from linear regressions of $\log(R_T)$ on $\log(T)$ and on $\log(s)$.
}
\end{figure}

To assess tightness of the regret upper bound, we run additional experiments and report the results in Figure~\ref{fig:fitting_line}.
The empirical exponents in $T$ and $s$ are upper bounded by the theoretical exponents, in line with our analysis, and are close to them in value, suggesting the upper bound are tight in practice.

Furthermore, we compare SPARKLE with several contextual bandit baselines to show its effectiveness on nonlinear, sparse, high-dimensional problems.
Our benchmarks include:
\begin{itemize}[noitemsep, nolistsep]
    \item Nonlinear LASSO Bandit  \citep{bastani2020online}. This algorithm uses monomial basis functions to approximate nonlinear rewards, since linear models are inadequate for capturing the complex reward structure considered here.
    \item
    $k$NN-UCB \citep{reeve2018k}. This algorithm estimates rewards with $k$-nearest neighbors and selects arms using the upper confidence bound  method, and it is suitable for high-dimensional data under low intrinsic dimension.

    \item Naive KRR. This variant replaces SPARKLE's doubly penalized estimator with kernel ridge regression (See Section~\ref{subsec:RKHS} of the e-companion) using a non-additive kernel, so it applies only an RKHS-norm penalty and does not induce sparsity.

    \item Additive KRR. This variant replaces the doubly penalized estimator with KRR using an additive kernel, which enforces additivity but still does not induce sparsity.

\end{itemize}
The last two variants serve as ablations to assess the value of the doubly penalized estimator in high-dimensional settings.
We exclude the smooth bandit algorithm \citep{hu2022smooth} because it is computationally infeasible in high dimensions.

\begin{figure}[ht]
\FIGURE{
\begin{tabular}{ccc}
    \subfigure[$K=2,s=5,d=30$]{\includegraphics[width=0.33\textwidth]{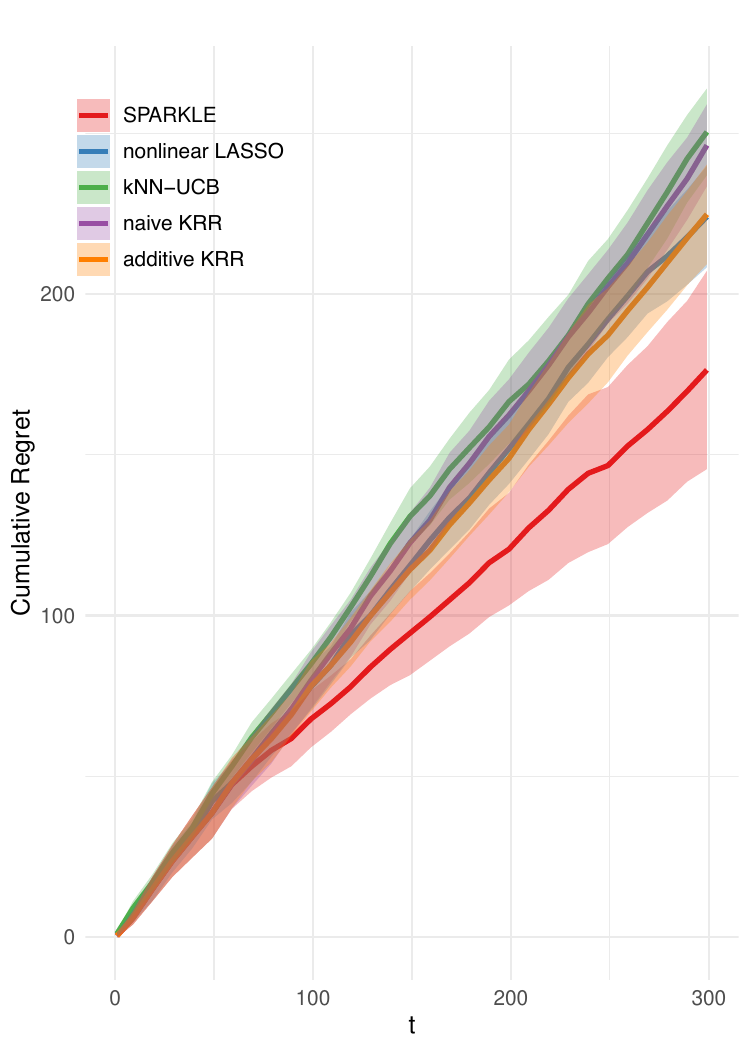}}
\subfigure[$K=5,s=2,d=100$]{\includegraphics[width=0.33\textwidth]{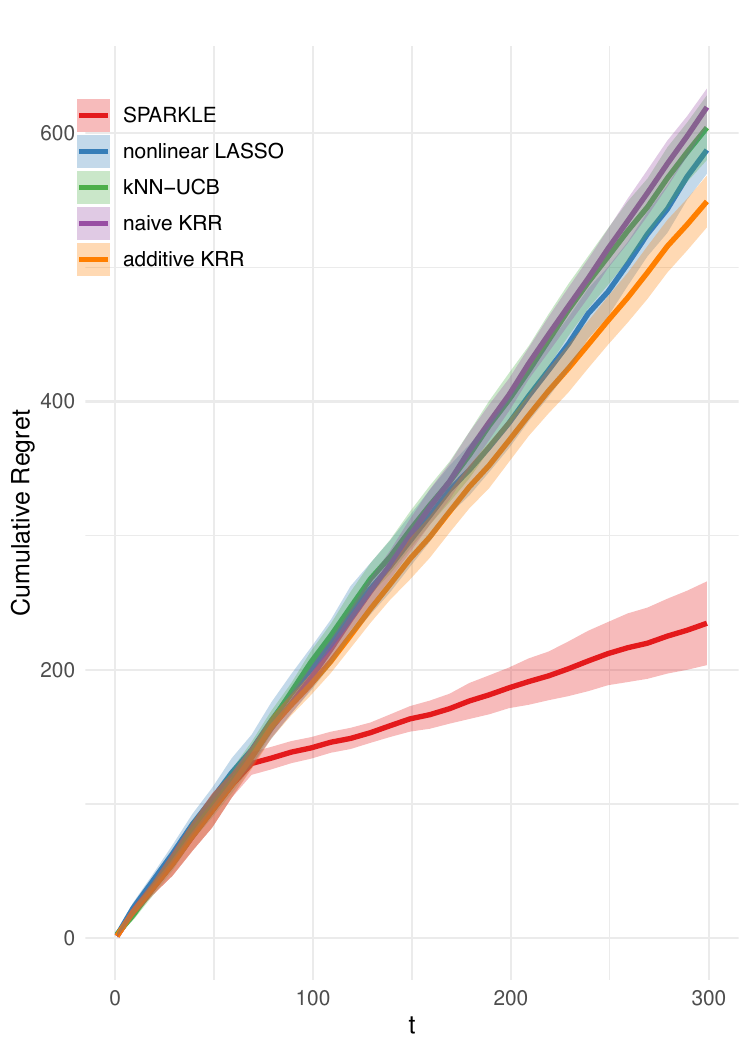}}
\subfigure[$K=2,s=2,d=500$]{\includegraphics[width=0.33\textwidth]{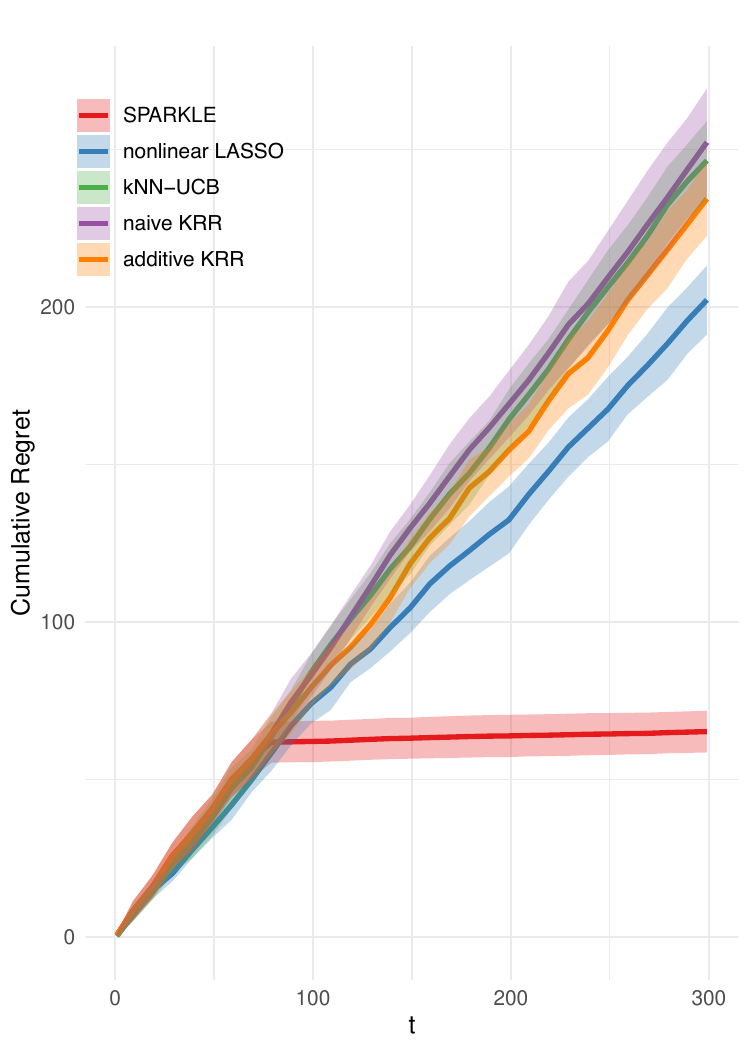}}
\end{tabular}
}
{Algorithm Comparison: Synthetic Data. \label{fig:comparison} }
{Each setting is repeated 10 times. Shaded bands show 95\% confidence intervals.}
\end{figure}

The results are shown in Figure~\ref{fig:comparison}.
SPARKLE consistently outperforms the other algorithms by a substantial margin, and the gap widens as $d$ increases.
Notably, although the nonlinear LASSO Bandit is a natural choice for handling complex rewards in high dimensions, its performance degrades when the original covariate space is already high-dimensional and the approximation requires projecting to a space whose dimensionality is at least $n$ times larger, where $n$ is the highest degree of the monomials in the basis.
These findings support the effectiveness of SPARKLE.

\subsection{Real-World Applications}\label{sec:real-data}

\subsubsection{Video Recommendation.}

Modern short-video platforms deliver personalized content in real time to maximize user engagement.
The main challenge is to model the complex relationships between high-dimensional user attributes and viewing behavior while optimizing an engagement metric.
A contextual bandit formulation fits this interactive setting, with videos as arms and user attributes as covariate vectors.
We conduct experiments on the KuaiRec dataset \citep{gao2022kuairec}, a rich dataset from Kuaishou, a leading video-sharing platform in China.  Unlike traditional recommendation datasets that are extremely sparse, KuaiRec has near-complete observational coverage (99.6\% density in the core subset), offering almost complete feedback across the user‑video space  (that is, almost all users in this dataset watched almost all videos and their responses were recorded).

We consider a two-armed problem where two specific videos (IDs: 2333 and 9582) are the arms.
The reward is the watch ratio, defined as the actual play duration divided by the total video duration, which provides a continuous engagement signal that captures nuanced viewing behavior.
To construct informative covariate vectors, we combine multiple sources of user information. We first extract attributes from demographic profiles and social network features. We then augment these with features derived from behavioral patterns, including aggregated watch statistics, engagement metrics, and content preferences from the interaction history. To avoid information leakage, we exclude the two target videos when computing all global features. For policy evaluation, we leverage the near-complete observations in the dataset to obtain counterfactual feedback directly.
The processed dataset contains 1,411 users with 65-dimensional covariate vectors and their responses to each of the two videos.
This high-dimensional feature space captures complexity in user preferences and behavior, allowing effective policy learning for personalized recommendations.

\begin{figure}[ht]
\FIGURE{
\begin{tabular}{cc}
    {\includegraphics[width=0.49\textwidth]{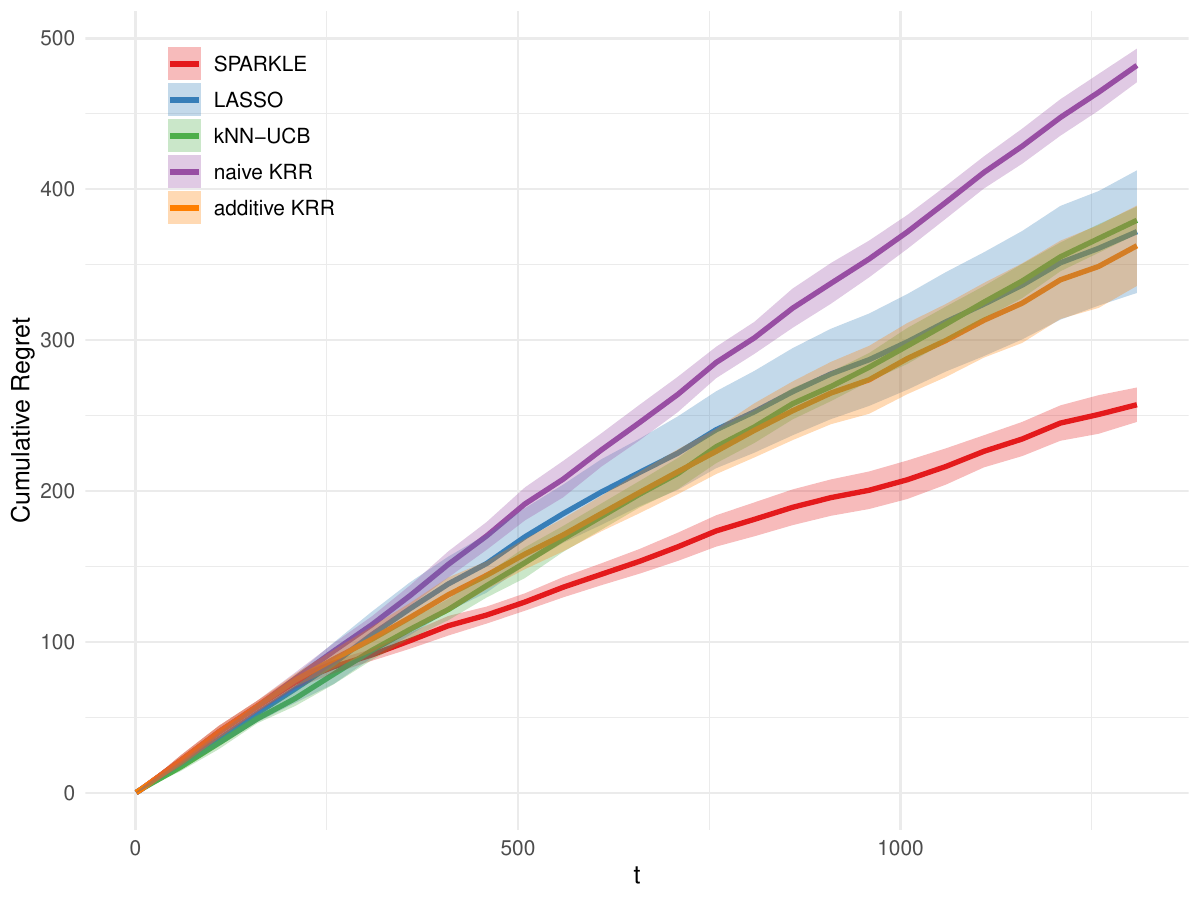}}
{\includegraphics[width=0.49\textwidth]{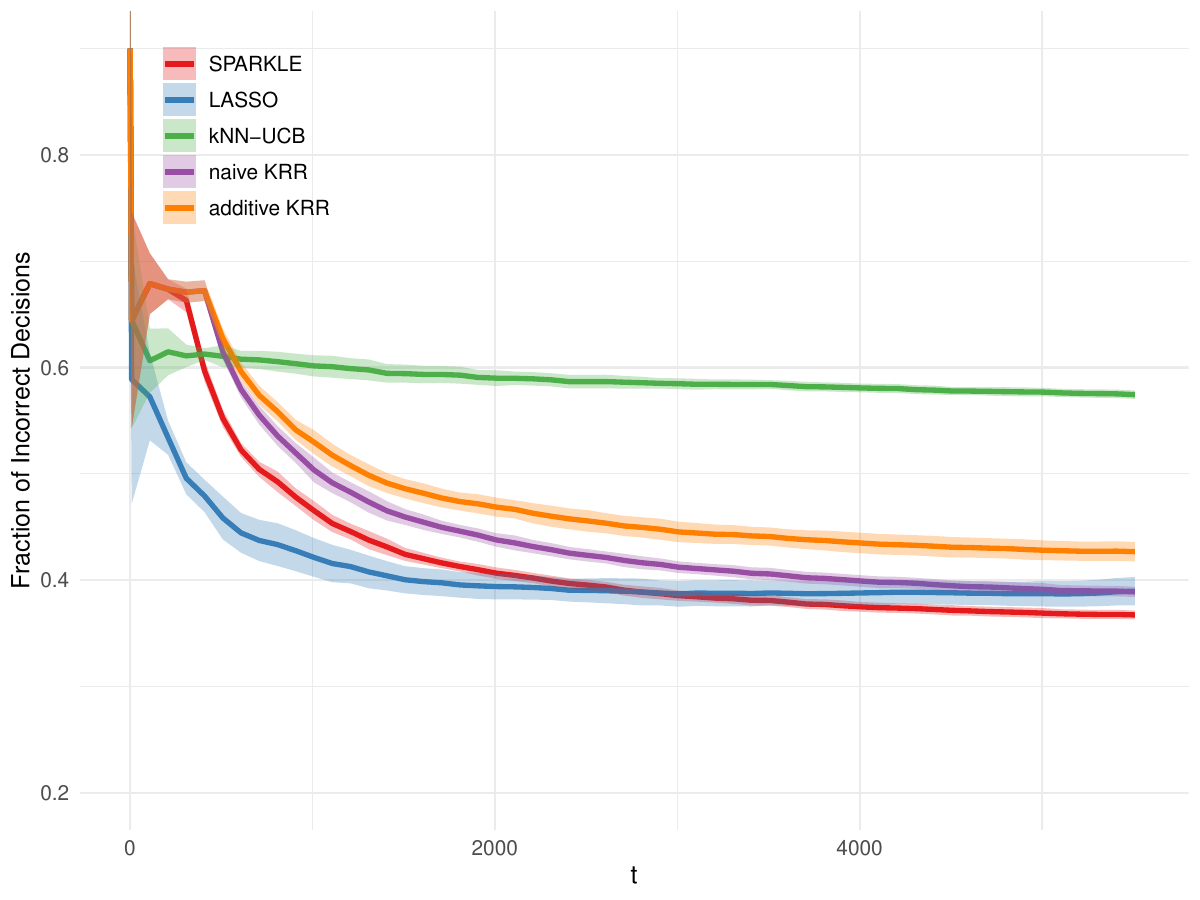}}
\end{tabular}
}
{Algorithm Comparison: Video Recommendation (Left) and Warfarin Dosing (Right) \label{fig:realdata}}
{Left: expected cumulative regret in the video recommendation task. Right: average fraction of incorrect dosing decisions in the warfarin dosing task. Results are averaged over 10 independent random permutations. Shaded bands show 95\% confidence intervals.}
\end{figure}

We compare SPARKLE with the same benchmark algorithms used in Section~\ref{sec:syn-data}  over 10 independent random permutations, except that the LASSO Bandit \citep{bastani2020online} is implemented in its standard linear form.
The left panel of Figure~\ref{fig:realdata} shows that SPARKLE consistently outperforms the other methods across all permutations, leveraging continuous rewards and rich rich contextual information. This performance in a recommendation setting indicates that SPARKLE handles high-dimensional contexts and nuanced reward structures common in modern decision-making problems.

\subsubsection{Warfarin Dosing.}

To demonstrate SPARKLE's flexibility across diverse reward structures, we evaluate it on personalized warfarin dosing, a classic problem in personalized medicine with a narrow therapeutic window and high inter‑individual variability.
A contextual bandit that learns dose decisions from sequential patient data can enable safer, more personalized care.
We use the public International Warfarin Pharmacogenetics Consortium (IWPC) dataset, which aggregates data from multiple institutions and has been used in prior high-dimensional bandit studies \citep{bastani2020online}. The dataset includes demographics, medical history, concomitant medications, and genetic variants, forming a rich set of covariates that influence warfarin metabolism and efficacy.

Following standard preprocessing for high-dimensional medical data, we perform data cleaning, one‑hot encoding of categorical variables, imputation for missing values, and add missingness indicators. After preprocessing, the dataset contains 5,528 patients described by 118 covariates.
We simulate an contextual bandit environment in which, at each round, an algorithm observes a patient's covariate vector and recommends one of three dosage buckets defined by the IWPC: (1) low dose, under 3 mg/day (about 33\% of cases), (2) medium dose, 3-7 mg/day (54\% of cases), and (3) high dose, over 7 mg/day (13\% of cases).
The reward is $0$ if the recommended dose falls within the patient’s true therapeutic range (based on the recorded stable dose) and $-1$ otherwise.
Counterfactual outcomes come directly from the observed data.

For evaluation, we adopt the LASSO Bandit framework \citep{bastani2020online} and compare the average fraction of incorrect dosing decisions as a function of the number of patients treated (lower is better).
Although prior work \citep{international2009estimation} finds that ordinary least squares linear regression achieves strong cross-validation accuracy for predicting the correct warfarin dose from patient-level variables, SPARKLE’s sparse additive RKHS structure is flexible enough to perform well even under this simple reward model.

As shown in the right panel of Figure~\ref{fig:realdata}, SPARKLE outperforms the other algorithms, with substantial advantages over those nonparametric methods that struggle in high dimensions.
Early on, the LASSO Bandit  \citep{bastani2020online} is competitive due to the simplicity of its linear reward model.
With more data, SPARKLE learns more detailed patterns from the high-dimensional covariates and achieves higher dosing accuracy.
This illustrates a common strength of high-dimensional nonparametric methods: they are stable with limited data and improve as samples grow.

\section{Conclusion}\label{sec:con}

In this paper, we present SPARKLE, an algorithm for contextual bandits with general nonlinear rewards and high-dimensional covariates.
We model the reward with a sparse additive structure in which each component is an RKHS function and only a few components are nonzero, allowing flexible and interpretable contributions from different feature subsets.
Our algorithm extends high-dimensional linear contextual bandit methods (such as the LASSO Bandit) to a nonparametric setting that accounts for reward smoothness, relaxing and broadening the linear-reward assumption that underlies much of the prior literature.

On the theory side, we establish, to our knowledge, the first sublinear regret bound for nonparametric contextual bandits that grows logarithmically in the covariate dimensionality,
$\tilde O\big(T^{1-\frac{(2m-1)(1+\alpha)}{4m+2}}\log d\big)$.
The bound highlights how smoothness ($m$) of the RKHS components affects the regret.
We also provide an information-theoretic lower bound, and the gap between the two bounds vanishes as $m \to \infty$.
Together, these results show that SPARKLE retains favorable dimensional scaling while adapting to function smoothness, bridging linear bandits and flexible nonparametric models.

On the empirical side, we run extensive experiments across domains. SPARKLE outperforms sparse linear baselines and nonparametric methods that ignore high dimensionality, with the advantage growing as $d$ increases. In video recommendation (continuous rewards) and warfarin dosing (binary rewards), it delivers consistent gains and improves with more data.
These results suggest SPARKLE is effective across reward types and context regimes and is practical for real-world decision problems.

We conclude by noting limitations and future directions. First, it remains open whether whether the doubly penalized estimator in SPARKLE can achieve a faster $L_\infty$ convergence rate, matching the best known rates for low-dimensional linear estimators.
Such a result would strengthen the regret analysis of SPARKLE.
Second, our framework currently relies on univariate additive components; extending it to capture higher-order interactions and overlapping group structures could increase expressiveness while preserving identifiability and computational tractability.
Finally, exploring adaptive hyperparameter selection and scalable implementations for very large $d$ and $T$ would further improve its applicability.

\begin{appendices}

\section{Consequence of Relaxing Assumptions~\ref{assum_optprob} and \ref{assum:regularity}}\label{sec:price}

Without Assumptions~\ref{assum_optprob}--\ref{assum:regularity}, which impose $\mathscr{C}$-regularity on the sample supports,
the learning problem becomes prohibitively difficult.
Theorem~\ref{thm:lbwithoutregularity} shows that, without these assumptions, the expected cumulative regret of any non-anticipating policy grows at a strictly higher order than SPARKLE’s regret under the assumptions as $T\to\infty$, not merely by a constant factor, implying unbounded performance degradation for any algorithm.

\begin{theorem}\label{thm:lbwithoutregularity}
Fix $\alpha>0$, and $m>0$ with $\alpha m > \frac{2m + 1}{2m - 1}$.
Let $\bar{R}_T = \tilde{O}\bigl(T^{1-\frac{(2m-1)(1+\alpha)}{4m+2}}\bigr)$ be the SPARKLE's regret upper bound from Theorem~\ref{thm:regretupperbound}, omitting the $\log(d)$ factor. For any non-anticipating policy $\pi$, there exists a contextual bandit instance with horizon $T$ that satisfies Assumption \ref{assum_densitybound}, \ref{assum_subG}, \ref{assump:kernel}, \ref{assum_compatibility}, \ref{assum_optimalregionprob}, and \ref{assum_margin}, but violates Assumption \ref{assum_optprob} and \ref{assum:regularity}, such that
\begin{align*}
     \frac{R_T(\pi)}{\bar{R}_T}
\rightarrow\infty  \text{\quad as\quad}T\rightarrow \infty.
\end{align*}
\end{theorem}

\section{Modified Assumptions in Theorem \ref{thm_lb_withd}}\label{sec:modified-assumptions}
\renewcommand{\theassumption}{\arabic{assumption}$'$} %
\setcounter{assumption}{2} %

\begin{assumption}[Smooth Kernels]\label{new_assump:kernel}
For each $k=1,\ldots,K$, the reward function $f_k^* \in \cN_{\Phi^{\prime}}(\Omega)$, where $\Phi^{\prime}:\Omega\times \Omega\mapsto \RR$ is a positive definite kernel
and $\cN_{\Phi^{\prime}}(\Omega)$ is its RKHS.
The kernel $\Phi'$ is stationary, i.e., $\Phi'(\bx, \widetilde{\bx}) = \phi'(\|\bx - \widetilde{\bx}\|)$ for some function $\phi'$ and all $\bx, \widetilde{\bx}\in\Omega$.
Its Fourier transform $\mathscr{F}_{\Phi^{\prime}}(\omega) \coloneqq  \int_{\RR^d} \Phi^{\prime}(r) e^{-2\pi\mathrm{i} \omega^\intercal r} \,\mathrm{d} r$ satisfies
\begin{equation*}
c_{1}\left(1+\|\omega\|^2_2\right)^{-m} \leq \mathscr{F}_{\Phi^{\prime}}(\omega) \leq c_{2}\left(1+\|\omega\|^2_2\right)^{-m},
\end{equation*}
for all $\omega\in \RR^d$ and some constants $m>3/2$, $c_1>0$, and $c_2>0$.
\end{assumption}
\setcounter{assumption}{4} %

\begin{assumption}\label{new_assum_optprob}
Define the nearly optimal region $\tilde{\mathcal{R}}_k=\{\bx\in\Omega|\max_{1\leq i\leq K}f^*_i(\bx)-f^*_k(\bx)\leq T^{-\frac{2m-d}{4m}}\}$.
\begin{enumerate}[label=(\roman*), noitemsep, nolistsep]
\item
There exist pairwise disjoint, connected sets $\{\tilde{\mathcal{R}}_{k,l}\}_{l=1}^{n_k}$ that satisfy $\min_{1\leq k\leq K, 1\leq l \leq n_k}\PP
(X\in\tilde{\mathcal{R}}_{k,l})\geq \tilde p$ for some constant $\tilde p>0$, such that
$\tilde{\mathcal{R}}_k = \cup_{l=1}^{n_k} \tilde{\mathcal{R}}_{k,l}$.
\item The projection of $\tilde{\mathcal{R}}_k$ onto the $j$-th dimension has gaps lower bounded by some $\mathfrak{r}_0>0$ for all $j$.
\end{enumerate}

\end{assumption}

\begin{assumption}\label{new_assum:regularity}
  Assume $\Omega$ is $\mathscr{C}$-regular. Define
   $\cQ_{\bx} = \argmax_{1\leq k\leq K}f_k^*(\bx)$, the set of arms that attain the maximum reward at covariate $\bx$.
  There exist positive constants $\mathfrak{r}, \tilde c$ and $c^{\prime}$ such that for all $k=1,\ldots,K$:
    \begin{itemize}
        \item[(i)] Define $\cX_0^{(k)} = \{\bx \in \Omega: k\in \cQ_{\bx}, |\cQ_{\bx}|>1\}$ and $B(\cX_0^{(k)})=\bigcup_{\bx\in\cX_0^{(k)}} B(\bx,\mathfrak{r})$, where $B(\bx,\mathfrak{r})$ denotes the ball in $\Omega$ centered at $\bx$ with radius $\mathfrak{r}$. For any $\bx_1\in B(\cX_0^{(k)})\setminus\cX_0^{(k)}$, let $\bx_0$ be the projection of $\bx_1$ on $\cX_0^{(k)}$. Then, for any $i,j \in \cQ_{\bx_0}$,  at least one of the following is true:
        \begin{itemize}
                      \item[$\bullet$] $|f_i^*(\bx_0+\mathfrak{t}(\bx_1-\bx_0))-f_j^*(\bx_0+\mathfrak{t}(\bx_1-\bx_0))|\leq T^{-\frac{2m-d}{4m}}$ for all $0\leq\mathfrak{t}\leq 1$;

            \item[$\bullet$] $\left|\nabla_{\bv}\big(f^*_i(\bx_0+\mathfrak{t}(\bx_1-\bx_0))-  f^*_j(\bx_0+\mathfrak{t}(\bx_1-\bx_0))\big)\right|>\tilde{c}$ ~for all $0\leq\mathfrak{t}\leq 1$, where $\bv=(\bx_1-\bx_0)/\|\bx_1-\bx_0\|_2$.
        \end{itemize}

        \item[(ii)] For all $l\notin \cQ_{\bx_0}$ with $\bx_0 \in \cX_0^{(k)}$ and $\bx_1\in B(\bx_0,\mathfrak{r})$, $\max_{k\leq j\leq K}f_k^*(\bx_1)-f_l^*(\bx_1)> c^\prime$.

    \end{itemize}\end{assumption}
\setcounter{assumption}{7}
\begin{assumption}[Margin Condition]\label{new_assum_margin}
    There exist positive constants $C_0$, $\alpha$, and $\delta_0\leq T^{-\frac{2m-d}{4m}}$ such that for any $\delta \geq \delta_0$ and all $i$ and $j$ in $\cK$ with $i\neq j$,  $\PP(0<|f_i^*(X)-f_j^*(X)|\leq \delta)\leq C_0\delta^\alpha$.
\end{assumption}

\end{appendices}

\bibliographystyle{informs2014} %
\bibliography{ref} %

\ECSwitch

\ECHead{Supplemental Material}
\EquationsNumberedBySection
In this e-companion, we present supporting materials for the main article, including additional assumptions, experimental details as well as the technical proofs for the primary propositions and theorems.

\section{Overview of Reproducing Kernel Hilbert Spaces}\label{subsec:RKHS}

An RKHS is a type of Hilbert space that is equipped with a reproducing kernel.
Given a set $\Omega\subseteq \RR^d$, let $\Psi:\Omega \times \Omega \mapsto \RR$ be a positive definite kernel:
that is, it is a symmetric function, and
$\sum_{i=1}^n \sum_{l=1}^n \beta_i \beta_l \Psi(\bx_i, \bx_l) \geq 0$,
for all $n\geq 1$, $\bx_i\in\Omega$, and $\beta_i\in \RR$ with $i=1,\ldots,n$.
By the Moore--Aronszajn theorem, $\Psi$ induces a unique RKHS $\mathcal{N}_{\Psi}$ equipped with an inner product $\langle \cdot, \cdot \rangle_{\mathcal{N}_{\Psi}}$.
Specifically,
\begin{align*}
\mathcal{N}_{\Psi} \coloneqq \biggl\{
g=\sum_{i=1}^\infty \beta_i \Psi(\cdot, \bx_i): \  &{} \bx_i \in\Omega, \ \beta_i \in \RR  \mbox{ for all $i\geq 1$ such that } \|g\|_{\mathcal{N}_{\Psi}} < \infty
\biggr\},
\end{align*}
where $\|\cdot\|_{\mathcal{N}_{\Psi}} \coloneqq \langle \cdot, \cdot \rangle_{\mathcal{N}_{\Psi}}^{\frac{1}{2}}$ denotes the RKHS norm, and for any $g,\tilde{g}\in \mathcal{N}_{\Psi}$  with
$g = \sum_{i=1}^{\infty} \beta_{i} \Psi(\cdot, \bx_{i})$
and $\tilde{g} = \sum_{l=1}^{\infty} \tilde{\beta}_l \Psi(\cdot, \tilde{\bx}_{l})$, the inner product of these two functions is defined as
\[\langle g, \tilde{g} \rangle_{\mathcal{N}_{\Psi}} = \sum_{i=1}^{\infty}\sum_{l=1}^{\infty} \beta_i \tilde{\beta}_l \Psi(\bx_{i}, \tilde{\bx}_{l}).\]
The term ``reproducing'' stems from the \emph{reproducing property} of RKHS functions, which states that
$g(\bx) = \langle g, \Psi(\cdot, \bx) \rangle_{\mathcal{N}_{\Psi}}$ for all $g\in\mathcal{N}_{\Psi}$ and $\bx\in\Omega$. This structure is particularly powerful in functional analysis and machine learning, as it allows function evaluation to be expressed via inner products.

RKHSs offer a flexible framework for function modeling, with flexibility driven by the choice of kernel. By selecting different kernels, one can span spaces from affine functions to complex nonlinear mappings.  For example, the bilinear kernel $\Psi(\bx, \tilde{\bx}) = \bx^\intercal \tilde{\bx}$ yields $\mathcal{N}_{\Psi}$ as the space of linear functions $g(\bx) = \bbeta^\intercal \bx$, where $\|g\|_{\mathcal{N}_{\Psi}} = \|\bbeta\|_2$ (the Euclidean norm of $\bbeta$).
For smooth functions,   the Mat\'ern family provides kernels with an explicit smoothness parameter that controls differentiability; see  \cite{ScholkopfSmola02} for more examples.
More generally, additive kernels
\begin{align}\label{eq:additive-ker}
    \Psi(\bx,\tilde{\bx}) = \sum_{j=1}^d \Psi_{(j)}(x_{(j)}, \tilde{x}_{(j)})
\end{align}
generate RKHS containing additive functions, which decompose into a sum of univariate components.

RKHSs are well-suited for nonparametric estimation of unknown functions, as facilitated by the representer theorem \citep[Section 4.2]{ScholkopfSmola02}.  This theorem allows the simplification of what would generally be an infinite-dimensional optimization problem---estimating a function  from its noisy observations via regularized empirical risk minimization over a function class---into a finite-dimensional problem.

Consider an arbitrary loss function $L:\left(\Omega\times \mathbb{R}^2\right)^n \mapsto \mathbb{R} \cup\{\infty\}$, and a strictly increasing function $g:[0, \infty) \mapsto \mathbb{R}$.
Let   $\{(\bx_i, y_i):i=1,\ldots,n\}$ be i.i.d. observations of a function $f^* \in \mathcal{N}_{\Psi}$.
Define an estimator of $f^*$ as the optimal solution to
\begin{align}\label{eq:fhatkrr}
\min_{f \in \mathcal{N}_{\Psi}} L\left(\left(x_1, y_1, f\left(x_1\right)\right), \ldots,\left(x_n, y_n, f\left(x_n\right)\right)\right) + \eta g(\|f\|_{\mathcal{N}_{\Psi}}),
\end{align}
where $\eta>0$ is a regularization parameter.

By the representer theorem, the estimator takes the form
$\hat{f}(\cdot)=\sum_{i=1}^n \hat{\beta}_i \Psi(\cdot,\bx_i)$, where
$\hat{\bbeta}= (\hat{\beta}_1,\ldots, \hat{\beta}_n)^\intercal$ solves the finite-dimensional optimization problem
\begin{align}
\min_{\bbeta \in \RR^n}  L(\left(x_1, y_1, (\bPsi \bbeta)_1\right), \ldots,\left(x_n, y_n, (\bPsi \bbeta)_n\right)) + \eta g(\sqrt{\bbeta^\intercal \bPsi \bbeta}), \label{eq:KRR-quad}
\end{align}
where $\by = (y_1,\ldots,y_n)^\intercal$ and $\bPsi$ is the Gram matrix with entries $\Psi(\bx_i, \bx_l)$ for all $i,l=1,\ldots,n$.

For example, with the $L_2$ loss and $g(z)=z^2$ in \eqref{eq:fhatkrr},
the estimator reduces to kernel ridge regression (KRR), a standard method for estimating RKHS functions. The optimization in \eqref{eq:KRR-quad} becomes
\begin{align*}
\min_{\bbeta \in \RR^n} \frac{1}{n}  \| \by - \bPsi \bbeta \|^2_2 + \eta \bbeta^\intercal \bPsi \bbeta,
\end{align*}
which has a closed-form solution.

However, the RKHS norm $\|\cdot\|_{\mathcal{N}_{\Psi}}$ does not induce sparse solutions in the KRR method, even when $\Psi$ is an additive kernel \eqref{eq:additive-ker}, for which the norm is expressed as $\|f\|_{\mathcal{N}_{\Psi}}^2 = \sum_{j=1}^d \|f_{(j)}\|_{\mathcal{N}_{\Psi_{(j)}}}^2$.
This structure is analogous to the $L_2$ norm in linear models, which also fails to induce sparsity in ridge regression.

This limitation makes KRR unsuitable for high-dimensional settings, where it typically suffers from the curse of dimensionality.
For nonparametric problems,
the sample complexity that KRR requires to achieve a desired estimation accuracy grows exponentially with the dimensionality $d$; see, e.g., \cite{TuoWangWu20}.
Therefore, a different form of penalization is necessary to enforce sparsity in the function estimator.
The method of double penalization  has recently emerged as an effective solution to address this issue.

\section{Additional Details of Numerical Experiments}\label{ec-sec:exp-details}
The algorithm parameters are set as follows: $\tau_q=\lceil7\times10^{-5}(s^{\frac{2m+1}{4m}}2^{q+4})^{\frac{4m+2}{2m-1}}(\log(d)+\log(T))\log(T)\rceil$, $m=2.5$, $C_3=0.01, C_4=1$. Note that here we assume the sparsity level $s$ is known and let $s_0=s$ in order to investigate the actual dependency on $s$.

In Figure \ref{fig:fitting_line}, we need to compute the exponents of $s$ and $T$ in the theoretical upper bound, which requires the value of $\alpha$ in margin condition. Although $\alpha$ is unknown, we can estimate it as $\alpha\approx 0.76$ based on Assumption \ref{assum_margin} as follows. Under the settings that $K=2,s=2,d=20$, for a given $\delta$, to estimate $\PP(0<|f_i^*(X)-f_j^*(X)|\leq \delta)$, a straightforward method is adopted: using the frequency of event $\{0<|f_i^*(X)-f_j^*(X)|\leq \delta\}$ as an estimate.
We first generate $X$ for 10,000 times, where $X^{(j)}\sim U(-5,5),j=1,...,d$. Then, we record the frequency of event $\{0<|f_1^*(X)-f_2^*(X)|\leq \delta\}$ happens over the 10,000 trials. We repeat the process for $\delta$ evenly distributed on $(0,1)$, and collect the data samples $\{(\delta_i,p_i)\}_{i=1}^{100}$, where $p_i:=\hat{\PP}_{X\sim \mathsf{P}_X}(0<|f_i^*(X)-f_j^*(X)|\leq \delta_i)$. By running a linear regression model $\log p_i=\beta_0+\beta_1 \log \delta_i$ based on samples $\{(\delta_i,p_i)\}_{i=1}^{100}$, we use the OLS estimate $\hat{\beta}_1$ as an estimate of $\alpha$, which gives $\hat{\alpha}\approx 0.76$. Then, using $m=2.5$, the theoretical exponents of $T$ and $s$ are approximately 0.41 and 1.8, respectively. To estimate empirical exponents, we fit linear regression models $\log(R_T) = \alpha_1 + \beta_1 \log(T)+\epsilon$ and $\log(R_T)=\alpha_2+\beta_2\log(s)+\epsilon$. The estimated coefficients, $\hat{\beta}_1=0.35$ and $\hat{\beta}_2=1.75$, are statistically significant with $p$-values $<10^{-4}$.

To make a fair comparison, we set the algorithm parameters as follows. In practice, the underlying sparsity is unknown. Thus, in SPARKLE, we choose $s_0=5\geq s$ as a relative large number and use $\tau_q=\lceil1.5\times10^{-5}(s_0^{\frac{2m+1}{4m}}2^{q+4})^{\frac{4m+2}{2m-1}}(\log(d)+\log(T))\log(T)\rceil$, $m=2.5$, $C_3=0.01$,  and $C_4=1$. For nonlinear LASSO Bandit algorithm, we set $q=1$ and $h=5$, as suggested in \cite{bastani2020online}, and set the degree of monomials as $n=10$ and regularization parameters as $\lambda_1=\lambda_{2,0}=0.01$. For the $k$NN-UCB bandit algorithm, we set $\theta=2$, $\varphi \equiv 1$ following the suggestions in Appendix B of \cite{reeve2018k}. For the naive KRR bandit algorithm and additive KRR bandit algorithm, we use the optimal parameters under $L_\infty$-norm error, i.e., $\lambda_{0}^\prime(\tau_{i,q})^{-1}$ and $\lambda_{0}^{\prime\prime}(\tau_{i,q})^{-1}$ with $\lambda_{0}^\prime=\lambda_{0}^{\prime\prime}=0.01$. We consider three settings of $(K,s,d)$ with horizon $T=300$: (1)$K=2,s=5,d=30$; (2)$K=5,s=2,d=100$; (3)$K=2,s=2,d=500$.

\section{Technical Proofs}
\subsection{Proof of Theorem \ref{thm:superrorbound}}\label{proof:prop1}
We first present several lemmas used in this proof. Lemma \ref{lem:errorbound} is a direct result of Proposition 4 in \cite{tan2019doubly}, and Lemma \ref{lemma:interp-ineq} states an interpolation inequality for additive models.

\begin{lemma}\label{lem:errorbound}
Suppose the conditions in Theorem \ref{thm:superrorbound} hold.
We have that for sufficiently large $n$,
\begin{align*}
    & \|\hat f - f^*\|_{L_2(\tilde{\Omega})}^2 + C_{10}\sum_{j=1}^{d}\left(\rho_n\left\|\hat f_{(j)} - f^*_{(j)}\right\|_{\mathcal{N}_\Phi(\tilde{\Omega})}+\lambda_n\left\|\hat f_{(j)} - f^*_{(j)}\right\|_{L_2(\tilde{\Omega})}\right)\nonumber\\
    \leq & C_{11} s\left\{n^{-\frac{m}{2m+1}}+\sqrt{\log (d / \delta) / n}\right\}^2,
\end{align*}
with probability at least $1-2\delta$.

\end{lemma}
\begin{lemma}[Interpolation inequality for additive model]\label{lemma:interp-ineq}
Let $\tilde \Omega\subset [0,1]^d$ be a $\mathscr{C}$-regular region, and $\tilde \Omega_j$ be its projection on the $j$-th dimension. If $$\sum_{j \in S^\complement}\left\|f_{(j)}\right\|_{L_2(\tilde \Omega_j)} \leq \xi \sum_{j \in S}\left\|f_{(j)}\right\|_{L_2(\tilde \Omega_j)}$$ holds, where $\xi$ is defined in Assumption \ref{assum_compatibility}, then we have for some positive constant $K_1$,
\begin{align*}
    \|f\|_{L_\infty(\tilde{\Omega})} \leq K_1\|f\|_{L_2(\tilde{\Omega})}^{1-\frac{1}{2m}}\|f\|_{H_1^m(\tilde \Omega)}^{\frac{1}{2m}},
\end{align*}
where $\|f\|_{H_1^m(\tilde \Omega)} = \sum_{j=1}^d \|f_{(j)}\|_{H_m(\tilde \Omega_j)}$ and $H_m(\tilde \Omega_j)$ is the Sobolev space with smoothness $m$ defined on $\tilde \Omega_j$.

\end{lemma}

\noindent\textit{Proof of Lemma~\ref{lemma:interp-ineq}.}
The proof is given by the following inequalities:
\begin{align*}
    & \|f\|_{L_\infty(\tilde\Omega)}\nonumber\\
    \leq &  \sum_{j=1}^d \|f_{(j)}\|_{L_\infty(\tilde\Omega_j)}\nonumber\\
    \leq &  C\sum_{j=1}^d \|f_{(j)}\|_{L_2(\tilde\Omega_j)}^{1-\frac{1}{2m}}\|f_{(j)}\|_{H^m(\tilde\Omega_j)}^{\frac{1}{2m}}\nonumber\\
    = & C\sum_{j\in S}\|f_{(j)}\|_{L_2(\tilde\Omega_j)}^{1-\frac{1}{2m}}\|f_{(j)}\|_{H^m(\tilde\Omega_j)}^{\frac{1}{2m}} + C\sum_{j\in S^\complement}\|f_{(j)}\|_{L_2(\tilde\Omega_j)}^{1-\frac{1}{2m}}\|f_{(j)}\|_{H^m(\tilde\Omega_j)}^{\frac{1}{2m}}\nonumber\\
    \leq & C\left(\sum_{j\in S}\|f_{(j)}\|_{L_2(\tilde\Omega_j)}\right)^{\frac{2m-1}{2m}}\left(\sum_{j\in S}\|f_{(j)}\|_{H^m(\tilde\Omega_j)}\right)^{\frac{1}{2m}}+C\left(\sum_{j\in S^\complement}\|f_{(j)}\|_{L_2(\tilde\Omega_j)}\right)^{\frac{2m-1}{2m}}\left(\sum_{j\in S^\complement}\|f_{(j)}\|_{H^m(\tilde\Omega_j)}\right)^{\frac{1}{2m}}\nonumber\\
    \leq & C(1+\xi^{1-\frac{1}{2m}})\left(\sum_{j\in S}\|f_{(j)}\|_{L_2(\tilde\Omega_j)}\right)^{1-\frac{1}{2m}}\|f\|_{H_1^m(\tilde \Omega)}^{\frac{1}{2m}}\nonumber\\
    \leq & C(1+\xi^{1-\frac{1}{2m}})\|f\|_{L_2(\tilde\Omega)}^{1-\frac{1}{2m}}\|f\|_{H_1^m(\tilde \Omega)}^{\frac{1}{2m}}.
\end{align*}
The first inequality is by the triangle inequality. The second inequality is by the interpolation inequality on one dimensional case with a common constant $C$, which can be done since $\tilde\Omega$ is $\mathscr{C}$-regular. Specifically, let $\tilde \Omega_j = \cup_{l=1}^{n_j}\tilde \Omega_{jl}$, where each $\tilde \Omega_{jl}$ is a connected and bounded domain, such that $\min_{1\leq k,l\leq n_j}{\rm dist}(\tilde \Omega_{jk},\tilde \Omega_{jl})\geq \mathscr{C} > 0$ for some constant $\mathscr{C}$. Denote the restriction of $f_{(j)}$ on $\tilde \Omega_{jl}$ as $f_{(jl)}$. Then, we can apply the usual Sobolev extension theorem to each $f_{(jl)}$, and denote the extended function as $f_{(jl)}^{\mathsf{e}}$. Hence, we have
\begin{align*}
    \|f_{(jl)}^{\mathsf{e}}\|_{L_2(\RR)} \leq c_1\|f_{(jl)}\|_{L_2(\tilde \Omega_j)},
    \|f_{(jl)}^{\mathsf{e}}\|_{H^m(\RR)} \leq c_2\|f_{(jl)}\|_{H^m(\tilde \Omega_j)},
\end{align*}
for all $l =1,...,n_j$. Choose smooth cutoff functions $\varphi_l$ such that $\varphi_l\in \cC^{\infty}(\hat \Omega_{jl})$ with $\varphi_l(x) = 1$ for all $x\in \tilde \Omega_{jl}$ and $\varphi_l(x) = 0$ for all $x\in \RR\setminus \hat \Omega_{jl}$, where $\hat \Omega_{jl}$ contains all points that have distance at most $\mathscr{C}/3$ away from $\tilde \Omega_{jl}$. Clearly, all $\hat \Omega_{jl}$'s are separated with distance at least $\mathscr{C}/3$, which allows us to define the function
\begin{align*}
    f_{(j)}^{\mathsf{e}}(x) = \sum_{l=1}^{n_j} \varphi_l(x)f_{(jl)}^{\mathsf{e}}(x) = \left\{\begin{array}{ll}
        \varphi_l(x)f_{(jl)}^{\mathsf{e}}(x) & \mbox{ if } x\in \hat \Omega_{jl}, \\
        0 & {\rm otherwise}.
    \end{array}\right.
\end{align*}
Since $f_{(jl)}^{\mathsf{e}}$ is an extension of $f_{(jl)}$ on $\RR$, and the norm of $\varphi_l$ can be bounded by a constant, it can be seen that $f_{(j)}^{\mathsf{e}}$ is an extension of $f_{(j)}$ with
\begin{align*}
    \|f_{(j)}^{\mathsf{e}}\|_{L_2(\RR)} \leq c_3\|f_{(j)}\|_{L_2(\tilde \Omega_j)},\\
    \|f_{(j)}^{\mathsf{e}}\|_{H^m(\RR)} \leq c_4\|f_{(j)}\|_{H^m(\tilde \Omega_j)}
\end{align*}
with some constants $c_3,c_4>0$ independent of $f_{(j)}$, since $\tilde \Omega$ is $\mathscr{C}$-regular.
Hence, applying the interpolation inequality to $f_{(j)}^{\mathsf{e}}$ yields
\begin{align*}
    \|f_{(j)}\|_{L_\infty(\tilde\Omega_j)} \leq \|f_{(j)}^{\mathsf{e}}\|_{L_\infty(\RR)}\leq C\|f_{(j)}^{\mathsf{e}}\|_{L_2(\RR)}^{1-\frac{1}{2m}}\|f_{(j)}^{\mathsf{e}}\|_{H^m(\RR)}^{\frac{1}{2m}}\leq Cc_3^{1 -\frac{1}{2m}}c_4^{\frac{1}{2m}}\|f_{(j)}\|_{L_2(\tilde\Omega_j)}^{1-\frac{1}{2m}}\|f_{(j)}\|_{H^m(\tilde\Omega_j)}^{\frac{1}{2m}},
\end{align*}
where the third inequality is by the H\"older inequality, the fourth inequality is by the condition of Lemma \ref{lemma:interp-ineq}, specifically $$\sum_{j \in S^\complement}\left\|f_{(j)}\right\|_{L_2(\tilde \Omega_j)} \leq \xi \sum_{j \in S}\left\|f_{(j)}\right\|_{L_2(\tilde \Omega_j)},$$
and the last inequality is by the theoretical compatibility condition, i.e., Assumption \ref{assum_compatibility}.
\hfill \Halmos
\endproof

\noindent\textit{Proof of Theorem \ref{thm:superrorbound}.}
Lemma \ref{lem:errorbound} implies that
with probability at least $1-2\delta$,
\begin{align}\label{eq:boundL2}
        \|\hat f - f^*\|_{L_2(\tilde{\Omega})}^2
    \lesssim s\left\{n^{-\frac{m}{2m+1}}+\sqrt{\log (d / \delta) / n}\right\}^2
\end{align}
and
\begin{align}\label{eq:boundH1m}
    \sum_{j=1}^{d}\rho_n\left\|\hat f_{(j)} -f^*_{(j)}\right\|_{\mathcal{N}_\Phi(\tilde{\Omega})} \lesssim s\left\{n^{-\frac{m}{2m+1}}+\sqrt{\log (d / \delta) / n}\right\}^2.
\end{align}

By our choice of $\lambda_n$ and $w_n$, it can be seen that $$w_n\geq \frac{1}{2}\left(C_4^{\frac{2m}{2m+1}} n^{-\frac{m}{2m+1}} + \left(\frac{\log (d / \delta)}{n}\right)^{1 / 2}\right),$$ which implies $$\rho_{n}=\lambda_n w_n\gtrsim \left(\gamma_{n}+\sqrt{\frac{\log (d / \delta)}{n}}\right)\left (n^{-\frac{m}{2m+1}}+\sqrt{\frac{\log (d / \delta)}{n}}\right).$$
Hence, \eqref{eq:boundH1m} gives us
\begin{align}\label{eq:boundH1mnorho}
\sum_{j=1}^{d}\left\|\hat f_{(j)} -f^*_{(j)}\right\|_{\mathcal{N}_\Phi(\tilde{\Omega})} &\lesssim \rho_n^{-1}s\left\{n^{-\frac{m}{2m+1}}+\sqrt{\log (d / \delta) / n}\right\}^2\nonumber\\
& \lesssim \frac{s(n^{-\frac{m}{2m+1}}+\sqrt{\log (d / \delta) / n)}}{\gamma_{n}+\sqrt{\log(d/\delta)/n}}.
\end{align}
If
$$C_4^{ \frac{2m}{2m+1}} n^{-\frac{m}{2m+1}}\le C_4 n^{-\frac{1}{2}}\left(\frac{\log (d / \delta)}{n}\right)^{-\frac{1}{4m}},$$
we have $\gamma_{n}=C_4^{ \frac{2m}{2m+1}} n^{-\frac{m}{2m+1}}$, which implies $$ \frac{n^{-\frac{m}{2m+1}}+\sqrt{\log (d / \delta) / n}}{\gamma_{n}+\sqrt{\log(d/\delta)/n}}\lesssim 1.$$
On the other hand, if $$C_4^{ \frac{2m}{2m+1}} n^{-\frac{m}{2m+1}} \geq C_4 n^{-\frac{1}{2}}\left(\frac{\log (d / \delta)}{n}\right)^{-\frac{1}{4m}} ,$$ we have $n^{-\frac{m}{2m+1}}\lesssim \sqrt{\log (d / \delta) / n}$ and $\gamma_{n}=C_4 n^{-\frac{1}{2}}\left(\frac{\log (d / \delta)}{n}\right)^{-\frac{1}{4m}}$, which imply
\begin{align*}
    \frac{n^{-\frac{m}{2m+1}}+\sqrt{\log (d / \delta) / n}}{\gamma_{n}+\sqrt{\log(d/\delta)/n}} \leq \frac{n^{-\frac{m}{2m+1}}+\sqrt{\log (d / \delta) / n}}{\sqrt{\log(d/\delta)/n}} \lesssim 1.
\end{align*}
Combining these two cases with \eqref{eq:boundH1mnorho}, we can conclude that
\begin{align*}
    \sum_{j=1}^{d}\left\|\hat f_{(j)} -f^*_{(j)}\right\|_{\mathcal{N}_\Phi(\tilde{\Omega})} &\lesssim s.
\end{align*}

In the proof of Proposition 4 in \cite{tan2019doubly}, it has been shown that for $\hat f - f^*$, the condition of Lemma \ref{lemma:interp-ineq} is satisfied. %
Hence we can apply Lemma \ref{lemma:interp-ineq} to $\hat f - f^*$, together with \eqref{eq:boundL2}, and obtain that with probability at least $1-2\delta$,
\begin{align*}
    \|\hat f - f^*\|_{L_\infty(\tilde{\Omega})} &\leq K_1\|\hat f - f^*\|_{L_2(\tilde{\Omega})}^{1-\frac{1}{2m}}\|\hat f - f^*\|_{H_1^m}^{\frac{1}{2m}}\nonumber\\
    &\lesssim s^{\frac{2m+1}{4m}}(n^{-\frac{m}{2m+1}}+\sqrt{\log (d / \delta) / n})^{1-\frac{1}{2m}}.
\end{align*}
This finishes the proof. \hfill\Halmos

\subsection{Proof of Proposition~\ref{prop:HMsupport}}

Recall that $$\cH_{k,q}:= \{\bx \in\Omega :\cK_{q-1}(\bx)=\{k\}\}=\bigcup_{h=1}^{q-1}\{\bx\in\Omega:|\cK_{h-1}(\bx)|>1,\cK_{h}(\bx)=\{k\}\}$$ and $\cM_{k,q} \coloneqq \left\{\bx \in\Omega :k\in\cK_{q-1}(\bx), \, |\cK_{q-1}(\bx)|>1  \right\}.$

\noindent\textit{Proof of Statement (1)}:

Obviously, for any $\bx\in\cM_{k,q+1}$,  $|\cK_{q}(\bx)|>1$ and $k\in\cK_{q}(\bx)$ implies that $|\cK_{q-1}(\bx)|>1$ and $k\in\cK_{q-1}(\bx)$, which is equivalent to $\bx\in \cM_{k,q}$. Thus, we can conclude that $\cM_{k,q+1}\subseteq \cM_{k,q}$.

\noindent\textit{Proof of Statement (2)}:

By the definition of $\cH_{k,q}$, it is obvious that $\cH_{k,q+1}:= \bigcup_{h=1}^{q}\{\bx\in\Omega:|\cK_{h-1}(\bx)|>1,\cK_{h}(\bx)=\{k\}\}=\cH_{k,q}\cup \{\bx\in\Omega:|\cK_{q-1}(\bx)|>1,\cK_{q}(\bx)=\{k\}\}$, which implies that $\cH_{k,q+1} \supseteq \cH_{k,q}$.

\noindent\textit{Proof of Statement (3)}:

First, we prove that for any $h\neq l$, $\{\bx\in\Omega:|\cK_{h-1}(\bx)|>1,\cK_{h}(\bx)=\{k\}\}$ and $\{\bx\in\Omega:|\cK_{l-1}(\bx)|>1,\cK_{l}(\bx)=\{k\}\}$ are disjoint. Assume that there exists $\bx\in \Omega$ such that both $|\cK_{h-1}(\bx)|>1,\cK_{h}(\bx)=\{k\}$ and $|\cK_{l-1}(\bx)|>1,\cK_{l}(\bx)=\{k\}$ hold. Then, $|\cK_{h-1}(\bx)|>1$ and $|\cK_{l}(\bx)|=1$ implies that $l>h-1$, since the number of the candidate arms is non-increasing. Similarly we have $|\cK_{l-1}(\bx)|>1$ and $|\cK_{h}(\bx)|=1$ implies that $h>l-1$. Therefore, we must have $l=h$, which contradicts the assumption and finishes the proof.

According to the statement that for $h\neq l$, $\{\bx\in\Omega:|\cK_{h-1}(\bx)|>1,\cK_{h}(\bx)=\{k\}\}$ and $\{\bx\in\Omega:|\cK_{l-1}(\bx)|>1,\cK_{l}(\bx)=\{k\}\}$ are disjoint, we have $$\cH_{k,q+1}\setminus\cH_{k,q}=\{\bx\in\Omega:|\cK_{q-1}(\bx)|>1,\cK_{q}(\bx)=\{k\}\},$$ which is a subset of $\{\bx\in\Omega:k\in\cK_{q-1}(\bx),|\cK_{q-1}(\bx)|>1\} = \cM_{k,q}$, since $\cK_{q}(\bx)=\{k\}$ implies $k\in\cK_{q-1}(\bx)$. \hfill\Halmos

\subsection{Proof of Corollary \ref{prop:sample-support}}

\noindent\textit{Proof of Statement (1)}:
Since $\cS_{k,q}=\cM_{k,q}\cup\cH_{k,q}$, we have $\cS_{k,q+1}=\cM_{k,q+1}\cup\cH_{k,q+1}=\cM_{k,q+1}\cup\cH_{k,q+1}\setminus \cH_{k,q}\cup \cH_{k,q}$. By Proposition \ref{prop:HMsupport}, we have $\cM_{k,q+1}\cup\cH_{k,q+1}\setminus \cH_{k,q}\subseteq \cM_{k,q}$, then $\cS_{k,q+1}=\cM_{k,q+1}\cup\cH_{k,q+1}\subseteq \cM_{k,q}\cup\cH_{k,q}=\cS_{k,q}$.

\noindent\textit{Proof of Statement (2)}:
By the selection rule, $\PP(\bx_t=\bx|\pi_t=k,t\in\mathcal{T}_q)>0$ is equivalent to $k\in\cK_{q-1}(\bx)$. Then, apparently we have $\bx\in \cS_{k,q}$ if and only if $k\in \cK_{q-1}(\bx)$.

\noindent\textit{Proof of Statement (3)}:
For any $\bx\in\Omega$, $\PP(\bx_t=\bx)>0$ by Assumption \ref{assum_densitybound}. If there exists $\bx\in\Omega$ s.t. $\bx\notin \bigcup_{k=1}^K \cS_{k,q}$, then $\PP(\bx_t=\bx|\pi_t=k,t\in\mathcal{T}_q)=0$ for all $k=1,\ldots,K$. By the Bayesian rule we have $\PP(\bx_t=\bx)=0$, which is contradictory. Thus we can verify that $\bigcup_{k=1}^K \cS_{k,q} = \Omega$.
\hfill\Halmos

\subsection{Proof of Lemma~\ref{lemma:alphasmallerthan1}}

This proof is a direct result of Proposition 3.4 of \cite{audibert2007fast}. First, since we assumed $m>3/2$ in Assumption \ref{assump:kernel}, by the embedding relationship of Sobolev space and H\"older space $\mathcal{W}^{m}([0,1]) \subset \cH^{\beta=m-\frac{1}{2}}([0,1])$, we have $\beta>1$, where $\beta$ is the smoothness parameter for the H\"older space. Under the the strong density assumption and  margin assumption (stated in Assumptions \ref{assum_densitybound} and \ref{assum_margin}, respectively), and by following the same procedures as in the proof of Proposition 3.4 in \cite{audibert2007fast}, if there exists $t>\delta$ with $\delta\leq T^{-\frac{2m-1}{4m}}$ such that $t^\alpha > C t $ (where $C$ is a positive constant defined in the proof of Proposition 3.4 of \cite{audibert2007fast}), then for any $k,j=1,\ldots,K$ and all $\bx\in\Omega$, either $f_k^*(\bx)-f_j^*(\bx)=0$ or $|f_k^*(\bx)-f_j^*(\bx)|>\tau_{\min}$. Here, $\tau_{\min}$ is a positive constant that depends only on the parameters in Assumptions \ref{assum_densitybound}, \ref{assump:kernel} and \ref{assum_margin}. Since $T$ is sufficiently large, if $\alpha>1$, such a $t$ exists because $t^\alpha < C t $ holds for, say, $t=T^{-\frac{2m-1}{4m}}$.

Since Assumption \ref{assum_optimalregionprob} holds, we have that  $\min_{1 \leq k \leq K}\PP(X\in\mathcal{R}_{k})\geq p^*$, which contradicts $|f_k^*(\bx)-f_j^*(\bx)|>\tau_{\min}$ for all $\bx\in\Omega$. Thus, we have $f_k^*(\bx)-f_j^*(\bx)=0$ for any $k,j=1,\ldots,K$, and the regret is trivial, i.e., $R_T=0$.\hfill\Halmos

\subsection{Proof of Lemma~\ref{lemma:regularity}}

First, we have that $\cX_0^{(1)} = \cX_0^{(2)} = \{\bx \in \Omega: f^*_1(\bx)=f^*_2(\bx)\}$, thus $\cQ_{\bx_0}=\{1,2\}=\cK$ for $\bx_0\in \cX_0^{(1)}= \cX_0^{(2)}$.  We will show that it satisfies the two conditions in Assumption \ref{assum:regularity} as follows.

\noindent\textit{Assumption \ref{assum:regularity} (i).} If $|f_1^*(\bx)-f_2^*(\bx)|\leq T^{-\frac{2m-1}{4m}}$ for all $\bx\in\Omega$, then obviously the first scenario in Assumption \ref{assum:regularity} (i) holds. It remains to consider the scenario that $\|\nabla f^*_1(\bx)- \nabla f^*_2(\bx)\|_2>2\tilde{c}$ for all $\bx\in \cX_0^{(1)}=\cX_0^{(2)}$. We consider two cases.

\textit{Case 1, $d=1$.} By Assumption \ref{assump:kernel}, $f_1^{*\prime}(x)-  f^{*\prime}_2(x)$ is continuous on a compact set $\Omega$, then by the Heine–Cantor theorem, it is uniformly continuous. Then, there exists a constant $\mathfrak{r}>0$ such that for any $x_0\in\cX_0^{(k)}$ and $|x-x_0|\leq\mathfrak{r}$,
\begin{align*}
    \left|f_1^{*\prime}(x)-  f^{*\prime}_2(x) - (f_1^{*\prime}(x_0)-  f^{*\prime}_2(x_0))\right|\leq \tilde c,
\end{align*}
which, together with $\left|f_1^{*\prime}(x_0)-  f^{*\prime}_2(x_0)\right|>2\tilde{c}$, implies that
\begin{align*}
    \left|f_1^{*\prime}(x)-  f^{*\prime}_2(x)\right| \geq  \left|f_1^{*\prime}(x_0)-  f^{*\prime}_2(x_0)\right| - \tilde c > \tilde c.
\end{align*}
For any $x_1\in B(\cX_0^{(k)})\setminus\cX_0^{(k)}$ and $0\leq \mathfrak{t} \leq 1$, we have $|x_0+\mathfrak{t}(x_1-x_0)-x_0|\leq \mathfrak{r}$, thus $$|f_1^{*\prime}(x_0+\mathfrak{t}(x_1-x_0))-  f^{*\prime}_2(x_0+\mathfrak{t}(x_1-x_0))|>\tilde{c},$$ which implies Assumption \ref{assum:regularity} (i) holds.

\textit{Case 2, $d>1$.}  It suffices to consider $\cX_0^{(2)}$ because $\cX_0^{(1)}=\cX_0^{(2)}$ in this two-arm scenario. Since $\bx_0$ is the projection of $\bx_1$ on $\cX_0^{(2)}$, $\bx_0$ is the solution of the optimization problem
\begin{align*}
    \argmin_{\bx\in \cX_0^{(2)}} &\|\bx_1-\bx\|_2,\\
    \text { s.t. } & f^*_1(\bx)-f^*_2(\bx)=0.
\end{align*}
By Assumption \ref{assump:kernel}, $f^*_1(\bx)-f^*_2(\bx)$ has continuous first-order derivative. Therefore, the Lagrange multiplier theorem implies that there exists a unique Lagrange multiplier $\lambda^*$ such that $$\frac{\bx_1-\bx_0}{\|\bx_1-\bx_0\|_2}=\lambda^*\nabla \big(f^*_1(\bx_0)-f^*_2(\bx_0)\big).$$ Since we choose $\bx_1\in B(\cX_0^{(k)})\setminus\cX_0^{(k)}$, it can be seen that $\bx_1\neq \bx_0$, and thus $\lambda^*\neq 0$.

Therefore, $\bx_1-\bx_0$ is parallel to $\nabla \big(f^*_1(\bx_0)-f^*_2(\bx_0)\big)$, which gives that
\begin{align*}
    \big|\nabla_{\bv}\big(f^*_1(\bx_0)-  f^*_2(\bx_0)\big)\big|&=\bigg|\frac{\nabla \big(f^*_1(\bx_0)-f^*_2(\bx_0)\big)^\intercal(\bx_1-\bx_0)}{\|\bx_1-\bx_0\|_2}\bigg|\nonumber\\
    &=\|\nabla \big(f^*_1(\bx_0)-f^*_2(\bx_0)\big)\|_2>2\tilde{c}.
\end{align*}

 Similar to Case 1, Assumption \ref{assump:kernel} implies that $\nabla_{\bv}\big(f^*_1(\bx)-  f^*_2(\bx)\big)$ is continuous for all $\bx=\bx_0+\mathfrak{t}(\bx_1-\bx_0)\in \Omega$, which, together with the Heine–Cantor theorem, implies that $\nabla_{\bv}\big(f^*_1(\bx)-  f^*_2(\bx)\big)$ is uniformly continuous. Therefore, similar to Case 1 again, there exists $\mathfrak{r}>0$ such that if $\|\bx-\bx_0\|_2=\|\mathfrak{t}(\bx_1-\bx_0)\|_2\leq\mathfrak{r}$, then $|\nabla_{\bv}\big(f^*_1(\bx)-  f^*_2(\bx)\big)|>\tilde{c}$. By the fact that $\bx=\bx_0+\mathfrak{t}(\bx_1-\bx_0)$ for $0\leq\mathfrak{t}\leq 1$ and certainly $\|\bx-\bx_0\|_2=\|\mathfrak{t}(\bx_1-\bx_0)\|_2\leq\mathfrak{r}$, we have
 \begin{align*}
     &\big|\nabla_{\bv}\big(f^*_1(\bx_0+\mathfrak{t}(\bx_1-\bx_0))-  f^*_2(\bx_0+\mathfrak{t}(\bx_1-\bx_0))\big)\big|>\tilde{c}
 \end{align*}
 for all $0\leq\mathfrak{t}\leq 1$. Thus, Assumption \ref{assum:regularity} (i) holds for the case $d>1$. Together with Case 1, we show that Assumption \ref{assum:regularity} (i) holds.

\noindent\textit{Assumption \ref{assum:regularity} (ii).} Since $\cQ_{\bx_0}=\{1,2\}=\cK$, there is no arm in $\cK\setminus\cQ_{\bx_0}$, and Assumption \ref{assum:regularity} (ii) holds trivially. This finishes the proof. \hfill\Halmos

\subsection{Proof of Proposition~\ref{prop:sfd}}

We first present the following lemma needed for this proof. Lemma \ref{lem_hoeffding} is the Hoeffding’s inequality for Bernoulli distribution.

\begin{lemma}[Hoeffding’s inequality for Bernoulli distribution]\label{lem_hoeffding}
    Let $X_1,...,X_n$ be independent Bernoulli random variables. Define $S_n=X_1+\cdots+X_n$, then for all $t>0$,
\begin{align*}
    \PP\left(\EE\left[S_n\right]-S_n \geq t\right) \leq \exp \left(-2 t^2 / n\right).
\end{align*}
\end{lemma}

\noindent\textit{Proof of Proposition~\ref{prop:sfd}.} For $t\in\cT_q$, Lemma~\ref{prop:armioptimal} implies that, under event $\bigcap_{1 \leq h \leq q-1} \mathcal{A}_h$, for any arm $k=1,\ldots,K$, if $\bx_t\in \mathcal{R}_k$, we have $k\in\cK_{q-1}(\bx_t)$. Since event $\bigcap_{1 \leq h \leq q-1} \mathcal{A}_h$ occurs,
\begin{align*}
    \PP(\pi_t=k|\bx_t\in \mathcal{R}_k)\geq 1/K.
\end{align*}
Furthermore, Assumption \ref{assum_optimalregionprob} implies that
\begin{align}
    \PP(\bx_t\in\mathcal{R}_k)\geq p^*,
\end{align}
which follows that
\begin{align}\label{eq_app_lemsd}
    \PP(\pi_t=k)\geq \PP(\pi_t=k,\bx_t\in \mathcal{R}_k)\geq p^*/K.
\end{align}
Note that the decision series $\{\pi_t\}_{t\in\cT_q}$ are i.i.d. conditional on $\{\cD_{h}\}_{h=1}^{q-1}$. By Hoeffding's inequality of Bernoulli distribution, we have
\begin{align*}
 & \PP\left(\tau_{k,q}< \frac{p^*}{2K}\tau_q\bigg|\{\cD_{h}\}_{h=1}^{q-1}\right)\nonumber\\
 = & \PP\left(\sum_{t\in\cT_q}\II(\pi_t=k)< \frac{p^*}{K}\tau_q-\frac{p^*}{2K}\tau_q\bigg|\{\cD_{h}\}_{h=1}^{q-1}\right)\nonumber\\
 \leq & \PP\left(\sum_{t\in\cT_q}\II(\pi_t=k)< \sum_{t\in\cT_q}\PP(\pi_t=k)-\frac{p^*}{2K}\tau_q\bigg|\{\cD_{h}\}_{h=1}^{q-1}\right)\nonumber\\
 = & \PP\left(\sum_{t\in\cT_q}\PP(\pi_t=k)-\sum_{t\in\cT_q}\II(\pi_t=k)>\frac{p^*}{2K}\tau_q\bigg|\{\cD_{h}\}_{h=1}^{q-1}\right)\nonumber\\
    \leq & e^{-2(\frac{p^*}{2K}\tau_q)^2/\tau_q}\nonumber\\
   = & e^{-\frac{\tau_qp*^2}{2K^2}},
\end{align*}
where the first inequality is by \eqref{eq_app_lemsd}.
Therefore, we can conclude that $ \PP(\tau_{k,q}\geq \frac{p^*}{2K}\tau_q\big|\{\cD_{h}\}_{h=1}^{q-1})\geq 1-e^{-\frac{\tau_qp*^2}{2K^2}}$. \hfill\Halmos

\subsection{Proof of Proposition~\ref{prop:dst&dist}}

When $q=1$, since we randomly pull arms with equal probability, it follows that $$\mu_1^{(k)}(\bx) = f_{\bx_t|\pi_t=k}(\bx)=\frac{\PP(\pi_t=k|\bx_t = \bx)p(\bx)}{\PP(\pi_t=k)}=\frac{1/Kp(\bx)}{1/K}=p(\bx).$$ Therefore, the conclusion trivially holds because of Assumption \ref{assum_densitybound}.

When $q>1$, the decisions in the epoch $q$ are only dependent on the history samples, thus it is obvious that $\cD_{k,q}$ are i.i.d. conditional on $\{\cD_{h}\}_{h=1}^{q-1}$. Then, we have
\begin{align}\label{eq_pflmbon1}
    \mu_k^{(q)}(\bx) & =f_{\bx_t|\pi_t=k,\{\cD_{h}\}_{h=1}^{q-1},\bigcap_{1 \leq h \leq q-1} \mathcal{A}_h}(\bx) \nonumber\\
    &=\frac{f_{\bx_t|\{\cD_{h}\}_{h=1}^{q-1},\bigcap_{1 \leq h \leq q-1} \mathcal{A}_h}(\bx)\PP(\pi_t=k|\bx_t=\bx,\{\cD_{h}\}_{h=1}^{q-1},\bigcap_{1 \leq h \leq q-1} \mathcal{A}_h)}{\PP(\pi_t=k|\{\cD_{h}\}_{h=1}^{q-1},\bigcap_{1 \leq h \leq q-1} \mathcal{A}_h)}\nonumber\\
    &=p(\bx) \frac{\PP(\pi_t=k|\bx_t=\bx,\{\cD_{h}\}_{h=1}^{q-1},\bigcap_{1 \leq h \leq q-1} \mathcal{A}_h)}{\PP(\pi_t=k|\{\cD_{h}\}_{h=1}^{q-1},\bigcap_{1 \leq h \leq q-1} \mathcal{A}_h)}.
\end{align}
When $\bx \in \cS_{k,q}$, i.e., $k$ is potentially the best arm at point $\bx_t = \bx$, our algorithm ensures that
\begin{align*}
   1/K \leq \PP\left(\pi_t=k|\bx_t = \bx,\{\cD_{h}\}_{h=1}^{q-1},\bigcap_{1 \leq h \leq q-1} \mathcal{A}_h\right)\leq 1.
\end{align*}
When $\bx \notin \cS_{k,q}$,
\begin{align*}
 \PP\left(\pi_t=k|\bx_t = \bx,\{\cD_{h}\}_{h=1}^{q-1},\bigcap_{1 \leq h \leq q-1} \mathcal{A}_h\right)=0.
\end{align*}
Moreover,
\begin{align*}
  1\geq &\PP\left(\pi_t=k|\{\cD_{h}\}_{h=1}^{q-1},\bigcap_{1 \leq h \leq q-1} \mathcal{A}_h\right)\nonumber\\\geq &\PP\left(\pi_t=k|\bx_t\in\cS_{k,q},\{\cD_{h}\}_{h=1}^{q-1},\bigcap_{1 \leq h \leq q-1} \mathcal{A}_h\right)\PP(\bx_t\in\cS_{k,q}|\{\cD_{h}\}_{h=1}^{q-1},\bigcap_{1 \leq h \leq q-1} \mathcal{A}_h)\nonumber\\
  &\geq \frac{1}{K}p^*
\end{align*}
which implies when $\bx \in \cS_{k,q}$, $\frac{\PP(\pi_t=k|\bx_t=\bx,\{\cD_{h}\}_{h=1}^{q-1},\bigcap_{1 \leq h \leq q-1} \mathcal{A}_h)}{\PP(\pi_t=k|\{\cD_{h}\}_{h=1}^{q-1},\bigcap_{1 \leq h \leq q-1} \mathcal{A}_h)}=0$, and when $\bx \in \cS_{k,q}$,
\begin{align}\label{eq_pflmbon2}
    \frac{1}{K}\leq \frac{\PP(\pi_t=k|\bx_t=\bx,\{\cD_{h}\}_{h=1}^{q-1},\bigcap_{1 \leq h \leq q-1} \mathcal{A}_h)}{\PP(\pi_t=k|\{\cD_{h}\}_{h=1}^{q-1},\bigcap_{1 \leq h \leq q-1} \mathcal{A}_h)} \leq \frac{K}{p^*}.
\end{align}
Combining \eqref{eq_pflmbon1} and \eqref{eq_pflmbon2}, together with $p_{\min }\leq p(\bx)\leq p_{\max }$ as in Assumption \ref{assum_densitybound}, we have that $\frac{1}{K} p_{\min }\leq \mu_k^{(q)}\leq {\frac{K}{p^*} p_{\max }}$ when $\bx \in \cS_{k,q}$, and $\mu_k^{(q)}=0$ when $\bx \notin \cS_{k,q}$.\hfill\Halmos

\subsection{Proof of Proposition~\ref{prop:cnstlength}}

\begin{lemma}\label{lem_prev7}
    Under Assumption \ref{assum:regularity},  there exists a positive constant $c^{\prime\prime}$ such that for all $k=1,\ldots,K$ and $\bx\in\Omega\setminus (B(\cX_0^{(k)})\cup \mathcal{R}_k)$, $\max_{1\leq j\leq K}f_j^*(\bx)-f_k^*(\bx)> c^{\prime\prime}$.
\end{lemma}

\noindent\textit{Proof of Lemma \ref{lem_prev7}}.
Let $h(\bx) = \max_{1\leq j\leq K}f_j^*(\bx)-f_k^*(\bx)$, where we suppress the dependency on $k$ for the ease of notation. By the continuity of max function, we have $h(\bx)$ is continuous.
We prove Lemma \ref{lem_prev7} by contradiction.

Consider an arm $k=1,\ldots,K$. Suppose Lemma \ref{lem_prev7} does not hold, that is, for any $\epsilon>0$, there exists $\bs\in\Omega\setminus (B(\cX_0^{(k)})\cup \mathcal{R}_k)$, s.t.  $0<h(\bs)\leq \epsilon$. Let $\epsilon_n=1/n$, $n=1,2,...$. Then, for every $\epsilon_n$, there exists $\bx_n\in\Omega\setminus (B(\cX_0^{(k)})\cup \mathcal{R}_k)$ such that $0<h(\bx_n)\leq \epsilon_n$. By the Bolzano–Weierstrass theorem, since $X_n=\{\bx_1,\bx_2,...\}$ are on a bounded set, there exists a convergent subsequence $X_{n}^{\prime}=\{\bx_{n},\bx_{n'},...\}$. Let $$\lim_{l\rightarrow\infty}\bx_{n_l}=\bx_0.$$
Hence, by the continuity of $h(\bx)$, we have $$0\leq h(\bx_0) = \lim_{l\rightarrow\infty} h(\bx_{n_l})\leq \lim_{l\rightarrow\infty} \epsilon_{n_l} = 0,$$ which implies $\bx_0\in \cX_0^{(k)}\cup \mathcal{R}_k$.
Since $\lim_{l\rightarrow\infty}\bx_{n_l}=\bx_0$, we have $\|\bx_{n_g}-\bx_0\|_2\leq \mathfrak{r}$ for some large enough $n_g$, which contradicts the fact that $\bx_{n_g}\in\Omega\setminus (B(\cX_0^{(k)})\cup \mathcal{R}_k)$. Thus, there exists a constant $c^{\prime\prime}>0$ s.t. for all $\bx\in\Omega\setminus (B(\cX_0^{(k)})\cup \mathcal{R}_k)$,  $h(\bx)> c^{\prime\prime}$, which concludes the proof. \hfill\Halmos

 \begin{lemma}\label{lem:RksubsetSkq}
    Under event $\bigcap_{1 \leq j \leq {q-1}}\mathcal{A}_{j}$, it holds that $\tilde{\mathcal{R}}_k\subseteq S_{k,q}$ as long as $T$ is large enough.
\end{lemma}

\noindent\textit{Proof of Lemma~\ref{lem:RksubsetSkq}.} Recall that $\tilde{\mathcal{R}}_k=\{\bx\in\Omega|\max_{1\leq i\leq K}f^*_i(\bx)-f^*_k(\bx)\leq T^{-\frac{2m-1}{4m}}\}$.  Then, under event $\bigcap_{1 \leq j \leq {q-1}}\mathcal{A}_{j}$, for any $\bx\in\tilde{\mathcal{R}}_k$, denote $i=\arg\max_{j\in\cK_{q-2}(\bx)}\hat{f}_{j,{q-1}}(\bx)$. We prove that $k\in \cK_{q-1}(\bx)$, which is equivalent to $\tilde{\mathcal{R}}_k\subseteq S_{k,q}$, by induction. First, obviously $k\in \cK_{0}(\bx)$. For $h\leq q-2$, assume $k\in \cK_{h}(\bx)$, which implies that $\bx\in S_{k,h+1}$. Then, denote $b_h= \arg\max_{j\in\cK_{h}(\bx)}\hat{f}_{j,{h+1}}(\bx)$, we have that
\begin{align*}
   \hat{f}_{b_h,h+1}(\bx)-\hat{f}_{k,h+1}(\bx)&= (\hat{f}_{b_h,h+1}(\bx_t)-f^*_{b_h}(\bx))+f^*_{b_h}(\bx)-f^*_k(\bx)-(\hat{f}_{k,h+1}(\bx)-f^*_k(\bx)) \nonumber\\
   & \leq |\hat{f}_{b_h,h+1}(\bx)-f^*_{b_h}(\bx)|+f^*_{b_h}(\bx)-f^*_k(\bx)+|\hat{f}_{k,h+1}(\bx)-f^*_k(\bx)| \nonumber\\
   &\leq 1/8\epsilon_{h+1}+T^{-\frac{2m-1}{4m}}+1/8\epsilon_{h+1}\nonumber\\
    &\leq \epsilon_{h+1},
\end{align*}
where the second inequality is because the definition of $\tilde{\mathcal{R}}_k$ and event $\bigcap_{1 \leq h \leq q-2} \mathcal{A}_h$, and the last inequality is because $T^{-\frac{2m-1}{4m}}\leq 1/4\epsilon_{h+1}$ for all $h\leq Q-1$ as long as $T$ is sufficiently large. Therefore, with the induction assumption we have $k\in \cK_{h+1}(\bx)$, from which we can conclude that $k\in \cK_{q-1}(\bx_t)$.\hfill\Halmos

\noindent\textit{Proof of Proposition~\ref{prop:cnstlength}.} For any arm $k=1,\ldots,K$, let the projection of $\cS_{k,q}$ on dimension $j$ for $j=1,...,d$ be decomposed into intervals $\cup_{l=1}^{p_j^q}[a_{j,l}^{q},b_{j,l}^{q}]$
with $a_{j,1}^{q}\leq b_{j,1}^{q}<a_{j,2}^{q}\leq b_{j,2}^{q}<\ldots<a_{j,p_j^q}^{q}\leq b_{j,p_j^q}^{q}$, where $p_j^q$ might be infinity (which in fact is impossible after we proved Proposition~\ref{prop:cnstlength}). Under event $\bigcap_{1 \leq j \leq {q-1}}\mathcal{A}_{j}$, we know that $\cS_{k,1},...,\cS_{k,q-1}$ is $\mathscr{C}$-regular. To prove that $\cS_{k,q}$ is $\mathscr{C}$-regular, we need to show the minimum length of each interval
\begin{align}\label{eq_app_lemcn1}
    \min_{1\leq l\leq p_j^q,1\leq j\leq d}(b_{j,l}^{q} - a_{j,l}^{q})\geq \mathscr{C}
\end{align}
and the minimum distance between the intervals
\begin{align}\label{eq_app_lemcn2}
    \min_{1\leq l\leq p_j^q-1,1\leq j\leq d} (a_{j,l+1}^{q} - b_{j,l}^{q})\geq \mathscr{C}
\end{align}
for a positive constant $\mathscr{C}$, when $T$ is sufficiently large. Then, we choose a sufficiently large $C_5$ and assume $T>C_5$.

\subsubsection{Proof of \eqref{eq_app_lemcn1}.}

Divide $\cS_{k,q}$ into disjoint regions, denoted as $\cS_{k,q}=S_{q,1}^{(k)}\cup...\cup S_{q,n_{q,k}}^{(k)}$. It suffices to prove that there exists some constant $\mathscr{C}$ s.t. $\min_{1\leq l\leq n_{q,k}} {\rm Vol}(S_{q,l}^{(k)})\geq \mathscr{C}$. This is because if there exists $l$ such that the projection of $S_{q,l}^{(k)}$ on $j$-th dimension converges to zero, we must have ${\rm Vol}(S_{q,l}^{(k)})$ converges to zero as well.
By Assumption \ref{assum_optprob}, we have that the volume of each disjoint component of $\tilde{\mathcal{R}}_k$ is lower bounded by $\tilde{p}$. Since $\tilde{\mathcal{R}}_k\subseteq \cS_{k,q}$, we have $\min_{j:S_{q,j}^{(k)}\cap \tilde{\mathcal{R}}_k\neq \emptyset } {\rm Vol}(S_{q,j}^{(k)})\geq \tilde{p}$. Therefore, it only remains to discuss $S_{q,v}^{(k)}$ where $v\in\{l:S_{q,l}^{(k)}\cap \tilde{\mathcal{R}}_k= \emptyset\}$.

If $S_{q,v}^{(k)}\cap \tilde{\mathcal{R}}_k= \emptyset$, we first show that for any $\bx\in S_{q,v}^{(k)}$, $\bx\in B(\cX_0^{(k)})$ in Assumption \ref{assum:regularity}, when $T$ is large enough. We prove this argument by contradiction. Suppose there exists an $\bx\in S_{q,v}^{(k)}$ and $\bx\notin B(\cX_0^{(k)})$. Then, by Lemma \ref{lem_prev7}, $\max_{1\leq j\leq K}f_j^*(\bx)-f_k^*(\bx)> c^{\prime\prime}$. Denote $i=\arg\max_{1\leq j\leq K}f_j^*(\bx)$. Since $\bx\in S_{q,v}^{(k)}$, we have
\begin{align*}
\hat{f}_{i,{q-1}}(\bx)-\hat{f}_{k,q-1}(\bx)\leq\epsilon_{q-1},
\end{align*}
which leads to
\begin{align}\label{eq_applemik1}
    f_i^*(\bx)-f_k^*(\bx)&=f_i^*(\bx)-\hat{f}_{i,{q-1}}(\bx)+\hat{f}_{i,q-1}(\bx)-\hat{f}_{k,q-1}(\bx)+\hat{f}_{k,q-1}(\bx)-f_k^*(\bx)\nonumber\\
    & \leq \frac{1}{8}\epsilon_{q-1}+\epsilon_{q-1}+\frac{1}{8}\epsilon_{q-1}\nonumber\\
    & = \frac{5}{4}\epsilon_{q-1},
\end{align}
where the inequality is because under event $\bigcap_{1 \leq j \leq {q-1}}\mathcal{A}_{j}$, for $\bx\in S_{q,v}^{(k)}$, $$|\hat{f}_{k,q-1}(\bx)-f_k^*\left(\bx\right)|\leq \frac{1}{8}\epsilon_{q-1}.$$

By \eqref{eq_applemik1}, together with $\max_{1\leq j\leq K}f_j^*(\bx)-f_k^*(\bx)> c^{\prime\prime}$, we obtain that
\begin{align*}
    \frac{5}{4}\epsilon_{q-1} > c^{\prime\prime},
\end{align*}
which is impossible because $\epsilon_{q-1}$ can be sufficiently small when $T$ is large enough. Therefore, we must have that for any $\bx\in S_{q,v}^{(k)}$, $\bx\in B(\cX_0^{(k)})$, as long as $T$ is large enough.

We then provide the following claims, which are about all arms in $\cQ_{\bx_0}$ for $\bx_0\in \cX_0^{(k)}$. Proofs of all claims are relegated to Section \ref{subsec:proofofclaim}. Let $\bx_1$ be any point in $B(\cX_0^{(k)})\setminus\cX_0^{(k)}$, $\bx_0$ be the projection of $\bx_1$ on $\cX_0^{(k)}$, and $\bv_1 = \bx_1 - \bx_0$.

\begin{claim}\label{lemcnsl_claim1} For any $i,j\in \cQ_{\bx_0}$, if $$\left|\nabla_{\bv_1/\|\bv_1\|_2}\big(f^*_i(\bx_0+\mathfrak{t}\bv_1)-  f^*_j(\bx_0+\mathfrak{t}\bv_1)\big)\right|>\tilde{c}$$ for all $0\leq\mathfrak{t}\leq 1$, then $\nabla_{\bv_1/\|\bv_1\|_2} \big(f^*_i(\bx_0+\mathfrak{t}\bv_1)-  f^*_j(\bx_0+\mathfrak{t}\bv_1)\big)$ does not change sign for all $0\leq\mathfrak{t}\leq 1$.
\end{claim}

\begin{claim}\label{lemcnsl_claim3}
    For all $i\in \cQ_{\bx_0}$ and a given $\bx_1\in \big(B(\cX_0^{(i)})\setminus\cX_0^{(i)}\big)\cap S_{q,v}^{(i)}$, choose $\bx_0$ as the projection of $\bx_1$ on $\cX_0^{(i)}$, then when $T$ is large enough, $\bx_0+\mathfrak{t}(\bx_1-\bx_0) \in S_{q,v}^{(i)}$ for all $0\leq \mathfrak{t} \leq 1$.
\end{claim}

Recall that we focus on such $S_{q,v}^{(k)}$ when $S_{q,v}^{(k)}\cap \tilde{\mathcal{R}}_k= \emptyset$, and we have shown that for any $\bx\in S_{q,v}^{(k)}$, it holds that $\bx\in B(\cX_0^{(k)})$, which leads to that $\bx\in \big(B(\cX_0^{(k)})\setminus\cX_0^{(k)}\big)\cap S_{q,v}^{(k)}$. With Claim \ref{lemcnsl_claim3}, we must have $S_{q,v}^{(k)}\cap \tilde{\mathcal{R}}_k \neq \emptyset$, since the projection of $\bx$ on $\cX_0^{(k)}$ belongs to $S_{q,v}^{(k)}\cap \tilde{\mathcal{R}}_k$, which is contradictory. Thus, $S_{q,v}^{(k)}\cap \tilde{\mathcal{R}}_k= \emptyset$ is impossible. Therefore, only the first case is valid and it gives $\min_{1\leq l\leq n_{q,k}} {\rm Vol}(S_{q,l}^{(k)})=\min_{j:S_{q,j}^{(k)}\cap \tilde{\mathcal{R}}_k\neq \emptyset } {\rm Vol}(S_{q,j}^{(k)})\geq \tilde{p}$, which shows that \eqref{eq_app_lemcn1} holds.

\subsubsection{Proof of \eqref{eq_app_lemcn2}.}

By Corollary~\ref{prop:sample-support}, we know that $\cS_{k,q}\subseteq \cS_{k,q-1}$. Then, any point that does not belong to $\cS_{k,q-1}$ would not belong to $\cS_{k,q}$, which implies that any interval $[a_{j,l}^{q-1}, b_{j,l}^{q-1}],l=1,...,p_j^{q-1}$ would not expand on epoch $q$. It follows that for any $j$ and $l$, the previous interval $[a_{j,l}^{q-1}, b_{j,l}^{q-1}]$ can either remain the same interval, shrink to a smaller interval or split into more than one intervals on epoch $q$.

Next we show that the intervals with the former two cases would not interfere with \eqref{eq_app_lemcn2}. Suppose an interval $[a_{j,l_1}^{q-1}, b_{j,l_1}^{q-1}]$ satisfies either one of the former two cases and changes to $[a_{j,l_2}^{q}, b_{j,l_2}^{q}]$ on epoch $q$ (since this interval remains the same interval or shrinks to a smaller interval, it only maps to one interval on epoch $q$), then $a_{j,l_1}^{q-1}\leq a_{j,l_2}^{q}$ and $b_{j,l_1}^{q-1}\geq b_{j,l_2}^{q}$. Also, we have $b_{j,l_2-1}^{q}\leq b_{j,l_1-1}^{q-1}$ and $a_{j,l_2+1}^{q}\geq a_{j,l_1+1}^{q-1}$ because any region ruled out from $S_{q-1}^{(k)}$ with its projection on dimension $j$ being $(b_{j,l_1-1}^{q-1},a_{j,l_1}^{q-1})$ or $(b_{j,l_1}^{q},a_{j,l_1+1}^{q})$ is also ruled out from $\cS_{k,q}$. Thus, $$a_{j,l_2}^{q}-b_{j,l_2-1}^{q}\geq a_{j,l_1}^{q-1}-b_{j,l_1-1}^{q-1}\geq
\mathscr{C}$$ and $$a_{j,l_2+1}^{q}-b_{j,l_2}^{q}\geq a_{j,l_1+1}^{q-1}-b_{j,l_1}^{q-1}\geq \mathscr{C}.$$

Hence, we only need to consider the intervals that split into more than one intervals and the gap generated by the new intervals. Specifically, suppose there exists an interval on epoch $q-1$ such that it splits into more than one intervals $\cup_{l=1}^n[a_l,b_l]$ with $n>1$. The generated gap $(b_{l},a_{l+1})$ implies that any region included in $S_{q-1}^{(k)}$ with its projection on dimension $j$ being $(b_{l},a_{l+1})$ is ruled out from $\cS_{k,q}$. Without loss of generality, we assume $j=1$ and the denote the gap as $(b_1,a_2)$.
Let $$S_1 := \cS_{k,q}\cap ([a_1,b_1]\times \prod_{j=2}^d [0,1]), \mbox{ and }S_2 := \cS_{k,q}\cap ([a_2,b_2]\times \prod_{j=2}^d [0,1]).$$ Then, we have that the projections of $S_1$ and $S_2$ on the first dimension are disjoint, and $S_1\cap S_2 = \emptyset$.

Take any two points $\bu_1 = (b_1,...,u_{1,d})\in S_1$ and $\bu_2 = (a_2,...,u_{2,d})\in S_2$.
Similar to the proof in the previous part, we have $\bu_1\in B(\cX_0^{(k)})$ and $\bu_2\in B(\cX_0^{(k)})$ and $\bx_0,\tilde{\bx}_0\in \cX_0^{(k)}$ is chosen as the projection of $\bu_1$ and $\bu_2$ on $\cX_0^{(k)}$, respectively. We provide the following claim.

\begin{claim}\label{lemcnsl_claim4}
    For sufficiently large $T$, $\|\bu_1-\bx_0\|_2\leq \frac{\mathfrak{r}_0}{3}$ and $\|\bu_2-\tilde\bx_0\|_2\leq \frac{\mathfrak{r}_0}{3}$.
\end{claim}

With Claim \ref{lemcnsl_claim4}, we can finish the proof. Note that Assumption \ref{assum_optprob} gives us that $\bx_0\in \tilde{\mathcal{R}}_{k,l_1}$ and $\tilde\bx_0\in \tilde{\mathcal{R}}_{k,l_2}$ for some $l_1$ and $l_2$. As shown in the proof of Claim \ref{lemcnsl_claim4}, we have for all $0\leq \mathfrak{t}\leq 1$, $\mathfrak{t}\bu_1 + (1-\mathfrak{t})\bx_0\in \cS_{k,q}$. Similarly, $\mathfrak{t}\bu_2 + (1-\mathfrak{t})\tilde\bx_0\in \cS_{k,q}$ for all $0\leq \mathfrak{t}\leq 1$. Then, we must have $l_1\neq l_2$, because otherwise $\bu_1$ and $\bu_2$ can be connected, which contradicts the assumption that $S_1\cap S_2 = \emptyset$. Since the projections of $\tilde{\mathcal{R}}_{k,l_1}$ and $\tilde{\mathcal{R}}_{k,l_2}$ has distance at least $\mathfrak{r}_0$, and the distance from $\bu_1$ ($\bu_2$) to $\tilde{\mathcal{R}}_{k,l_1}$ (respectively, $\tilde{\mathcal{R}}_{k,l_2}$) is at most $\frac{\mathfrak{r}_0}{3}$, we have the distance of the projections of $\bu_1$ and $\bu_2$ is at least $\mathfrak{r}_0 - \frac{2\mathfrak{r}_0}{3} = \frac{\mathfrak{r}_0}{3}$, which shows that $a_2-b_1\geq \frac{\mathfrak{r}_0}{3}$, and finishes the proof. \hfill\Halmos

\subsubsection{Proof of Claims \ref{lemcnsl_claim1}--\ref{lemcnsl_claim4}.}\label{subsec:proofofclaim}

\noindent\textit{Proof of Claim \ref{lemcnsl_claim1}.} If this claim is not true, then there exist $\mathfrak{t}_1,\mathfrak{t}_2\in [0,1]$ such that $$\nabla_{\bv_1/\|\bv_1\|_2} \big(f^*_i(\bx_0+\mathfrak{t}_1\bv_1)-  f^*_j(\bx_0+\mathfrak{t}_1\bv_1)\big) > \tilde c$$ and $$\nabla_{\bv_1/\|\bv_1\|_2} \big(f^*_i(\bx_0+\mathfrak{t}_2\bv_1) -  f^*_j(\bx_0+\mathfrak{t}_2\bv_1)\big)< -\tilde c.$$ By the continuity of $\nabla_{\bv_1/\|\bv_1\|_2} \big(f^*_i(\bx_0+\mathfrak{t}\bv_1)-  f^*_j(\bx_0+\mathfrak{t}\bv_1)\big)$, which is because of Assumption \ref{assump:kernel}, we must have that there exists $\mathfrak{t}_3$ between $\mathfrak{t}_1$ and $\mathfrak{t}_2$ such that $$\nabla_{\bv_1/\|\bv_1\|_2} \big(f^*_i(\bx_0+\mathfrak{t}_3\bv_1)-  f^*_j(\bx_0+\mathfrak{t}_3\bv_1)\big)=0,$$ which contradicts the assumption of Claim \ref{lemcnsl_claim1}.

\noindent\textit{Proof of Claim \ref{lemcnsl_claim3}.} We prove Claim \ref{lemcnsl_claim3} by mathematical induction. For the first epoch, we have $\cS_{i,1}=S_{1,1}^{(i)}=\Omega$ for all $i\in \cK$, thus the conclusion holds. Suppose that Claim \ref{lemcnsl_claim3} holds for epoch $g$, i.e., for all $i\in \cQ_{\bx_0}$, if $\bx_1\in \big(B(\cX_0^{(i)})\setminus\cX_0^{(i)}\big)\cap S_{g,v}^{(i)}$ for some $1\leq v\leq n_{g,i}$ and $\bx_0^\prime$ is the projection of $\bx_1$ on $\cX_0^{(i)}$, then $\bx_0^\prime+\mathfrak{t}(\bx_1-\bx_0^\prime) \in S_{g,v}^{(i)}$ for all $0\leq \mathfrak{t} \leq 1$.

Then, we show the claim holds for epoch $g+1$. Suppose $\bx_2\in \big(B(\cX_0^{(i)})\setminus\cX_0^{(i)}\big)\cap S_{g+1,v}^{(i)}$, and the projection of $\bx_2$ on $\cX_0^{(i)}$ is $\bx_0$. Denote $\bv_2=\bx_2-\bx_0$. Under event $\mathcal{A}_{g}$, it holds that for all $u\in \cQ_{\bx_0}$, %
$$\|\hat{f}_{u,g}-f_u^*\|_{L_2(S_{g}^{(u)})}\rightarrow 0, \|\hat{f}_{u,g}-f_u^*\|_{H^m(S_{g}^{(u)})}=O(1)$$
as $T\rightarrow\infty$. By the Gagliardo–Nirenberg interpolation inequality and letting $m_1 = \frac{3/2+m}{2} >3/2$, we have
\begin{align*}
    \|\hat{f}_{u,g}-f_u^*\|_{H^{m_1}(S_{g}^{(u)})}\lesssim \|\hat{f}_{u,g}-f_u^*\|_{L_2(S_{g}^{(u)})}^{\frac{m-m_1}{m}} \|\hat{f}_{u,g}-f_u^*\|_{H^m(S_{g}^{(u)})}^{\frac{m_1}{m}},
\end{align*}
and thus $\|\hat{f}_{u,g}-f_u^*\|_{H^{m_1}(S_{g}^{(u)})}\rightarrow 0$. By the embedding between Sobolev space and Hölder space $\mathcal{H}^{m_1}(\Omega) \subset \mathcal{C}^{\beta=m_1-\frac{1}{2}}(\Omega)$, we further have $\beta>1$ and thus when $T$ is large enough,
\begin{align*}
    \sup_{\bx \in S_{g}^{(u)}, u\in \cK, \|\bv\|_2 = 1}\left|\nabla_{\bv} f^*_u(\bx)-\nabla_{\bv}\hat{f}_{u,g}(\bx)\right| < \frac{\tilde c}{8}
\end{align*}
for the constant $\tilde c>0$ in Assumption \ref{assum:regularity}. Hence, for sufficiently large $T$,
\begin{align}\label{eq_lm6_gdsmall}
    \sup_{\bx \in S_{g}^{(u)}\cap S_{g}^{(j)}, u,j\in \cK, \|\bv\|_2 = 1}\left|\nabla_{\bv}\big( f^*_u(\bx)- f^*_j(\bx)\big)-\nabla_{\bv}\big(\hat{f}_{u,g}(\bx)-\hat{f}_{j,g}(\bx)\big)\right| < \frac{\tilde c}{4}.
\end{align}

For any $\mathfrak{t}_1 \in (0,1)$, let $i_{\mathfrak{t}_1} = \argmax_{i\in \cK_{g-1}(\bx_0+\mathfrak{t}_1\bv_2)}\hat{f}_{i,g}(\bx_0+\mathfrak{t}_1\bv_2)$. Then, $$\bx_0+\mathfrak{t}_1\bv_2\in S_{g+1}^{(i_{\mathfrak{t}_1})}.$$ If $i_{\mathfrak{t}_1} \notin \cQ_{\bx_0}$, by Assumption \ref{assum:regularity}, we have $$\max_{1\leq j\leq K}f_j^*(\bx_0+\mathfrak{t}_1\bv_2)-f_{i_{\mathfrak{t}_1}}^*(\bx_0+\mathfrak{t}_1\bv_2)> c^\prime,$$ which implies that for any $i \in \cQ_{\bx_0+\mathfrak{t}_1\bv_2}$,
\begin{align}\label{eq_applemcns_ccc}
    0 \leq & \hat{f}_{i_{\mathfrak{t}_1},g}(\bx_0+\mathfrak{t}_1\bv_2) - \hat{f}_{i,g}(\bx_0+\mathfrak{t}_1\bv_2) \nonumber\\
    \leq & \hat{f}_{i_{\mathfrak{t}_1},g}(\bx_0+\mathfrak{t}_1\bv_2) - f^*_{i_{\mathfrak{t}_1}}(\bx_0+\mathfrak{t}_1\bv_2)+ f^*_{i_{\mathfrak{t}_1}}(\bx_0+\mathfrak{t}_1\bv_2) - f^*_i(\bx_0+\mathfrak{t}_1\bv_2) + f^*_i(\bx_0+\mathfrak{t}_1\bv_2) - \hat{f}_{i,g}(\bx_0+\mathfrak{t}_1\bv_2)\nonumber\\
    \leq & \frac{1}{8}\epsilon_g - c^\prime + \frac{1}{8}\epsilon_g,
\end{align}
where the third inequality is because $\bx_0+\mathfrak{t}_1\bv_2\in S_g^{(i)}\cap S_g^{(i_{\mathfrak{t}_1})}$ and event $\mathcal{A}_g$. Since $\frac{1}{4}\epsilon_g- c^\prime<0$ as long as $T$ is large enough, it can be seen that \eqref{eq_applemcns_ccc} cannot hold when $T$ is large enough. Then, we have that $i_{\mathfrak{t}_1}\in \cQ_{\bx_0}$.

Consider $f_{i_{\mathfrak{t}_1}}^*(\bx_2)$ and $f_i^*(\bx_2)$. Since $\bx_2\in \big(B(\cX_0^{(i)})\setminus\cX_0^{(i)}\big)\cap S_{g+1,v}^{(i)}$ and $S_{g+1}^{(i)}\subseteq S_g^{(i)}$ by Proposition~\ref{prop:HMsupport}, our induction assumption implies that  $\bx_0+\mathfrak{t}\bv_2\in S_g^{(i)}$ for all $0\leq \mathfrak{t}\leq 1$. Therefore, for $i_{\mathfrak{t}_1},i\in \cQ_{\bx_0}$, we have three cases: (i) $|f_{i_{\mathfrak{t}_1}}^*(\bx_2) - f_i^*(\bx_2)|\leq T^{-\frac{2m-1}{4m}}$; (ii) $f_{i_{\mathfrak{t}_1}}^*(\bx_2) > f_i^*(\bx_2) + T^{-\frac{2m-1}{4m}}$; and (iii) $f_{i_{\mathfrak{t}_1}}^*(\bx_2) < f_i^*(\bx_2) - T^{-\frac{2m-1}{4m}}$. Let us consider these three cases one by one.

 Case (i). $|f_{i_{\mathfrak{t}_1}}^*(\bx_2) - f_i^*(\bx_2)|\leq T^{-\frac{2m-1}{4m}}$. Then, Assumption \ref{assum:regularity} implies that $|f_{i_{\mathfrak{t}_1}}^*(\bx_0 + \mathfrak{t}_1\bv_2) - f_i^*(\bx_0+\mathfrak{t}_1\bv_2)|\leq T^{-\frac{2m-1}{4m}}\leq \frac{1}{4}\epsilon_g$. By event $A_g$, we have
\begin{align*}
    & \hat{f}_{i_{\mathfrak{t}_1},g}(\bx_0 + \mathfrak{t}_1\bv_2)-\hat{f}_{i,g}(\bx_0+ \mathfrak{t}_1\bv_2)\nonumber\\
    \leq & \hat{f}_{i_{\mathfrak{t}_1},g}(\bx_0 + \mathfrak{t}_1\bv_2) - f_{i_{\mathfrak{t}_1}}^*(\bx_0 + \mathfrak{t}_1\bv_2) + f_{i_{\mathfrak{t}_1}}^*(\bx_0 + \mathfrak{t}_1\bv_2) - f_i^*(\bx_0+\mathfrak{t}_1\bv_2) + f_i^*(\bx_0+\mathfrak{t}_1\bv_2) - \hat{f}_{i,g}(\bx_0 + \mathfrak{t}_1\bv_2)\nonumber\\
    \leq & \frac{1}{8}\epsilon_g +\frac{1}{4}\epsilon_g+ \frac{1}{8}\epsilon_g = \frac{1}{2}\epsilon_g < \epsilon_g,
\end{align*}
which implies $\bx_0+\mathfrak{t}_1\bv_2\in S_{g+1}^{(i)}$.

Case (ii). $f_{i_{\mathfrak{t}_1}}^*(\bx_2) > f_i^*(\bx_2) +T^{-\frac{2m-1}{4m}}$.
In this case, Claim \ref{lemcnsl_claim1} implies that $$\nabla_{\bv_2/\|\bv_2\|_2}\big(f_{i_{\mathfrak{t}_1}}^*(\bx_0+\mathfrak{t}\bv_2)-f_i^*(\bx_0+\mathfrak{t}\bv_2)\big) > \tilde c$$ for all $0\leq\mathfrak{t}\leq 1$. Then, $f_{i_{\mathfrak{t}_1}}^*(\bx_0+\mathfrak{t}\bv_2)-f_i^*(\bx_0+\mathfrak{t}\bv_2)$ is strictly increasing. Next we prove $\bx_2\in S_g^{(i_{\mathfrak{t}_1})}$, by mathematical induction.

First, we know that $\bx_2\in S_1^{(i_{\mathfrak{t}_1})}=\Omega$. Then, for ${\tilde{g}}\leq g-1$, assume $\bx_2\in S_{\tilde{g}}^{(i_{\mathfrak{t}_1})}$. If $\bx_2\notin S_{\tilde{g}+1}^{(i_{\mathfrak{t}_1})}$, let $l_{\tilde{g}} = \argmax_{i\in \cK_{{\tilde{g}}-1}(\bx_2)}\hat{f}_{i,\tilde{g}}(\bx_2)$, which implies $\bx_2\in S_{{\tilde{g}}+1}^{(l_{\tilde{g}})}\subset S_{{\tilde{g}}}^{(l_{\tilde{g}})}$, and we have
\begin{align*}
    \hat f_{l_{\tilde{g}},\tilde{g}}(\bx_2) - \hat{f}_{i_{\mathfrak{t}_1},\tilde{g}}(\bx_2) > \epsilon_{{\tilde{g}}},
\end{align*}
which leads to
\begin{align*}
    & f_{l_{\tilde{g}}}^*(\bx_2) - f_i^*(\bx_2)\nonumber\\
    \geq & f_{l_{\tilde{g}}}^*(\bx_2) - f_{i_{\mathfrak{t}_1}}^*(\bx_2)\nonumber\\
    = & f_{l_{\tilde{g}}}^*(\bx_2) - \hat f_{l_{\tilde{g}},{\tilde{g}}}(\bx_2) + \hat f_{l_{\tilde{g}},{\tilde{g}}}(\bx_2) - \hat{f}_{i_{\mathfrak{t}_1},{\tilde{g}}}(\bx_2) + \hat{f}_{i_{\mathfrak{t}_1},{\tilde{g}}}(\bx_2) - f_{i_{\mathfrak{t}_1}}^*(\bx_2)\nonumber\\
    > & -\frac{1}{8}\epsilon_{\tilde{g}} + \epsilon_{\tilde{g}} - \frac{1}{8}\epsilon_{\tilde{g}}\nonumber\\
    = & \frac{3}{4}\epsilon_{\tilde{g}},
\end{align*}
where the second inequality holds under event $\cA_{\tilde g-1}$ and $\bx_2\in S_{{\tilde{g}}}^{(l_{\tilde{g}})}\cap S_{\tilde{g}}^{(i_{\mathfrak{t}_1})}$.
Hence, in epoch ${\tilde{g}}+1$, let $l_{{\tilde{g}}+1} = \argmax_{i\in \cK_{{\tilde{g}}}(\bx_2)}\hat{f}_{i,{{\tilde{g}}+1}}(\bx_2)$. Then, under event $\cA_{{\tilde{g}}+1}$ since ${\tilde{g}}+1\leq g$ and $\bx_2\in S_{{\tilde{g}}+1}^{(l_{\tilde{g}+1})}\cap S_{\tilde{g}+1}^{(i)}$, we have
\begin{align*}
     & \hat f_{l_{{\tilde{g}}+1},{{\tilde{g}}+1}}(\bx_2) - \hat{f}_{i,{{\tilde{g}}+1}}(\bx_2)\nonumber\\
     \geq & \hat f_{l_{{\tilde{g}}},{{\tilde{g}}+1}}(\bx_2) - f_{l_{{\tilde{g}}}}^*(\bx_2) + f_{l_{\tilde{g}}}^*(\bx_2) - f_i^*(\bx_2) + f_i^*(\bx_2) - \hat{f}_{i,{{\tilde{g}}+1}}(\bx_2)\nonumber\\
     \geq & -\frac{1}{8}\epsilon_{{\tilde{g}}+1} + \frac{3}{4}\epsilon_{{\tilde{g}}} - \frac{1}{8}\epsilon_{{\tilde{g}}+1}\nonumber\\
     = & -\frac{1}{4}\epsilon_{{\tilde{g}}+1} + \frac{3}{2}\epsilon_{{\tilde{g}}+1}\nonumber\\
     > & \epsilon_{{\tilde{g}}+1},
\end{align*}
which implies that $i$ has been excluded at epoch ${\tilde{g}}+2\leq g+1$, and contradicts our assumption $\bx_2\in \big(B(\cX_0^{(i)})\setminus\cX_0^{(i)}\big)\cap S_{g+1,v}^{(i)}$. Therefore, we must have $\bx_2 \in S_g^{(i_{\mathfrak{t}_1})}$, and by our induction assumption, $\bx_0+\mathfrak{t}\bv_2\in S_g^{(i_{\mathfrak{t}_1})}\cap S^{(i)}_g$ for all $0\leq \mathfrak{t}\leq 1$.
Then, by \eqref{eq_lm6_gdsmall}, we have
\begin{align}\label{eq_lm6_gdlb1}
    &\nabla_{\bv_2/\|\bv_2\|_2} \big(\hat{f}_{{i_{\mathfrak{t}_1}},g}(\bx_0+\mathfrak{t}\bv_2)-\hat{f}_{i,g}(\bx_0+\mathfrak{t}\bv_2)\big)\nonumber \\
    > &\nabla_{\bv_2/\|\bv_2\|_2} \big( f^*_{i_{\mathfrak{t}_1}}(\bx_0+\mathfrak{t}\bv_2)- f^*_i(\bx_0+\mathfrak{t}\bv_2)\big) - \frac{\tilde c}{4} > \frac{3\tilde c}{4},
\end{align}
for all $0\leq\mathfrak{t}\leq 1$. Thus, we have $$\frac{{\rm d}}{{\rm d}\mathfrak{t}}  \big(\hat{f}_{i_{\mathfrak{t}_1},g}(\bx_0 + \mathfrak{t}\bv_2)-\hat{f}_{i,g}(\bx_0+ \mathfrak{t}\bv_2)\big)> \frac{3\tilde c}{4}\|\bv_2\|_2>0,$$ which implies that $ \hat{f}_{i_{\mathfrak{t}_1},g}(\bx_0 + \mathfrak{t}\bv_2)-\hat{f}_{i,g}(\bx_0+ \mathfrak{t}\bv_2)$ is strictly increasing with respect to $\mathfrak{t}$. Then, we have
\begin{align}\label{eq_lm6_fhfh2}
    & \hat{f}_{i_{\mathfrak{t}_1},g}(\bx_0 + \mathfrak{t}_1\bv_2)-\hat{f}_{i,g}(\bx_0+ \mathfrak{t}_1\bv_2)\nonumber\\
    \leq &\hat{f}_{i_{\mathfrak{t}_1},g}(\bx_2)-\hat{f}_{i,g}(\bx_2)< \epsilon_g,
\end{align}
where the second inequality is because $\bx_2 \in S_{g+1}^{(i)}$. By \eqref{eq_lm6_fhfh2}, $\bx_0+\mathfrak{t}_1\bv_2\in S_{g+1}^{(i)}$.  %

Case (iii). $f_{i_{\mathfrak{t}_1}}^*(\bx_2) < f_i^*(\bx_2) -T^{-\frac{2m-1}{4m}}$. In this case, Claim \ref{lemcnsl_claim1} implies that $$\nabla_{\bv_2/\|\bv_2\|_2}\big(f^*_{i_{\mathfrak{t}_1}}(\bx_0+\mathfrak{t}(\bx_{\mathfrak{t}_1}-\bx_0))-  f^*_i(\bx_0+\mathfrak{t}(\bx_{\mathfrak{t}_1}-\bx_0))\big)\leq -\tilde c$$ and $f_i^*(\bx_0+\mathfrak{t}(\bx_{\mathfrak{t}_1}-\bx_0)) - f_{i_{\mathfrak{t}_1}}^*(\bx_0+\mathfrak{t}(\bx_{\mathfrak{t}_1}-\bx_0))$ is strictly increasing. By the definition of $i_{\mathfrak{t}_1}$, we have that $\bx_{\mathfrak{t}_1}\coloneqq\bx_0+\mathfrak{t}_1\bv_2\in S_g^{(i_{\mathfrak{t}_1})}$. By the induction assumption, we know that for any $0\leq \mathfrak{t}\leq 1$, $\bx_0+\mathfrak{t}(\bx_{\mathfrak{t}_1}-\bx_0)\in S_g^{(i_{\mathfrak{t}_1})}\cap S_g^{(i)}$. Then,
\begin{align*}
    &\nabla_{\bv_2/\|\bv_2\|_2} \big(\hat{f}_{i_{\mathfrak{t}_1},g}(\bx_0+\mathfrak{t}(\bx_{\mathfrak{t}_1}-\bx_0))-\hat{f}_{i,g}(\bx_0+\mathfrak{t}(\bx_{\mathfrak{t}_1}-\bx_0))\big) \nonumber\\
    < &\nabla_{\bv_2/\|\bv_2\|_2} \big( f^*_{i_{\mathfrak{t}_1}}(\bx_0+\mathfrak{t}(\bx_{\mathfrak{t}_1}-\bx_0))- f^*_i(\bx_0+\mathfrak{t}(\bx_{\mathfrak{t}_1}-\bx_0))\big) + \frac{\tilde c}{4} < -\frac{3\tilde c}{4},
\end{align*}
for all $0\leq\mathfrak{t}\leq 1$. Thus, $$\frac{{\rm d}}{{\rm d}\mathfrak{t}}  \big(\hat{f}_{i_{\mathfrak{t}_1},g}(\bx_0 + \mathfrak{t}\bv_2)-\hat{f}_{i,g}(\bx_0+ \mathfrak{t}\bv_2)\big)< -\frac{3\tilde c}{4}\|\bv_2\|_2<0,$$ which implies that $ \hat{f}_{i_{\mathfrak{t}_1},g}(\bx_0 + \mathfrak{t}_1\bv_2)-\hat{f}_{i,g}(\bx_0+ \mathfrak{t}_1\bv_2)$ is strictly increasing with respect to $\mathfrak{t}_1$. Then, we have
\begin{align}\label{eq_lm6_fhfh3}
    & \hat{f}_{i_{\mathfrak{t}_1},g}(\bx_{\mathfrak{t}_1})-\hat{f}_{i,g}(\bx_{\mathfrak{t}_1})\nonumber\\
    \leq & \hat{f}_{i_{\mathfrak{t}_1},g}(\bx_0)-\hat{f}_{i,g}(\bx_0)\nonumber\\
    = & \hat{f}_{i_{\mathfrak{t}_1},g}(\bx_0) - f_{i_{\mathfrak{t}_1}}^*(\bx_0) + f_{i_{\mathfrak{t}_1}}^*(\bx_0) - f_i^*(\bx_0) + f_i^*(\bx_0) - \hat{f}_{i,g}(\bx_0)\nonumber\\
    \leq & \frac{1}{8}\epsilon_g + \frac{1}{8}\epsilon_g < \epsilon_g,
\end{align}
where the second equality is because $i_{\mathfrak{t}_1},i\in \cQ_{\bx_0}$, which gives us $f_{i_{\mathfrak{t}_1}}^*(\bx_0) = f_i^*(\bx_0)$. It can be seen that \eqref{eq_lm6_fhfh3} implies $\bx_0+\mathfrak{t}_1\bv_2\in S_{g+1}^{(i)}$. Combining all these three cases, we have proved Claim \ref{lemcnsl_claim3}. \hfill\Halmos

\noindent\textit{Proof of Claim \ref{lemcnsl_claim4}.} We will only show $\|\bu_1-\bx_0\|_2\leq \frac{\mathfrak{r}_0}{3}$ because $\|\bu_2-\tilde\bx_0\|_2\leq \frac{\mathfrak{r}_0}{3}$ can be obtained similarly. If $k \in \cQ_{\bu_1}$, then $|\cQ_{\bu_1}|>1$. This is because if $\cQ_{\bu_1}=\{k\}$, then since $k$ is the only optimal arm on $\bu_1$, we have
\begin{align} \label{eq_lm6_cm4_eq1}
    f^*_k(\bu_1)-\max_{{i}\in \cK, i\neq k} f^*_i(\bu_1)>0.
\end{align}

Denote the first element of vector $a$ as $a_{[1]}$. For any point $\tilde{\bu}_1$ s.t. $a_2>\tilde{\bu}_{1[1]}>b_1$, $k$ is not the optimal arm on $\tilde{\bu}_1$, otherwise it should be in $\cS_{k,q}$. Thus,
\begin{align*}
     f^*_k(\tilde{\bu}_1)-\max_{{i}\in \cK, i\neq k} f^*_{i}(\tilde{\bu}_1)<0,
\end{align*}
 which contradicts \eqref{eq_lm6_cm4_eq1} because of the continuity of $f^*_k(\bx)-\max_{i\in \cK, i\neq k} f^*_i(\bx)$ and $\bu_{1[1]}=b_1$. Therefore, we must have that if $k \in \cQ_{\bu_1}$, then $|\cQ_{\bu_1}|>1$ and $\bu_1\in\cX_0^{(k)}$, which further implies
$\bx_0= \bu_1$ and the claim clearly holds.

It remains to consider $k \notin \cQ_{\bu_1}$. Take any $i_u\in \cQ_{\bu_1}$, which implies  $f_{i_u}^*(\bu_1) > f_k^*(\bu_1)$.  We further show that $f_{i_u}^*(\bu_1) > f_k^*(\bu_1)+T^{-\frac{2m-1}{4m}}$. Denote $o:=\arg\max_{i\in\mathcal{K}_{q-2}(\bu_1)}\hat{f}_{i,q-1}(
\bu_1)$. First, since $i_u\in \cQ_{\bu_1}$, we have
\begin{align}\label{eq:fostat-fiustatless0}
     f_o^*(\bu_1)-f_{i_u}^*(\bu_1)\leq 0.
\end{align}
Since for all $i\in\cK_{q-2}({\bu}_1)$, Assumption \ref{assump:kernel} and event $\mathcal{A}_{q-1}$ guarantee the existence of first-order derivative of $\hat{f}_{i,q-1}(\bx)$ on ${\bu}_1$, we have $\hat{f}_{i,q-1}(\bx)$ is continuous on ${\bu}_1$, thus $\max_{i\in\cK_{q-2}({\bu}_1)}\hat{f}_{i,q-1}(\bx)$ is continuous on ${\bu}_1$. Similarly, since ${\bu}_1 \in\cS_{k,q}$, we have $\hat{f}_{k,q-1}(\bx)$ is continuous on $\bu_1$. Still, by the fact that $\tilde{\bu}_1\notin\cS_{k,q}$, the continuity of $\max_{i\in\cK_{q-2}(\bx)}\hat{f}_{i,q-1}(\bx)-\hat{f}_{k,q-1}(\bx)$ on $\bu_1$ implies that
\begin{align*}
    \max_{i\in\cK_{q-2}(\bu_1)}\hat{f}_{i,q-1}(\bu_1)-\hat{f}_{k,q-1}(\bu_1)=\hat{f}_{o,q-1}(\bu_1)-\hat{f}_{k,q-1}(\bu_1)=\epsilon_{q-1}.
\end{align*}
Thus, by event $\mathcal{A}_{q-1}$, we have
\begin{align}\label{eq:fostar-fkstar}
    f_o^*(\bu_1)-f_{k}^*(\bu_1)&=f_o^*(\bu_1)-\hat{f}_{o,q-1}(\bu_1)+\hat{f}_{o,q-1}(\bu_1)-\hat{f}_{k,q-1}(\bu_1)+\hat{f}_{k,q-1}(\bu_1)-f_{k}^*(\bu_1)\nonumber\\
    &\geq -\frac{1}{8}\epsilon_{q-1}+\epsilon_{q-1}-\frac{1}{8}\epsilon_{q-1}= \frac{3}{4}\epsilon_{q-1}.
\end{align}
Combining \eqref{eq:fostat-fiustatless0} and \eqref{eq:fostar-fkstar}, we have
\begin{align}\label{eq:fiu-fk>T-2m-14m}
    f_{i_u}^*(\bu_1)-f_k^*(\bu_1)\geq \frac{3}{4}\epsilon_{q-1}>T^{-\frac{2m-1}{4m}},
\end{align}
as long as $T$ is sufficiently large.

Next, we show Claim \ref{lemcnsl_claim4} by contradiction. Suppose $\|\bu_1-\bx_0\|_2 > \frac{\mathfrak{r}_0}{3}$. Clearly this cannot hold if $\mathfrak{r}\leq \frac{\mathfrak{r}_0}{3}$ because $\bu_1\in \big(B(\cX_0^{(k)})\setminus\cX_0^{(k)}\big)\cap S_{k,q}$. Hence in the following, we assume $\mathfrak{r} > \frac{\mathfrak{r}_0}{3}$. By Assumption \ref{assum:regularity} and Claim \ref{lemcnsl_claim1}, we can only have $$\frac{{\rm d}}{{\rm d}\mathfrak{t}}\big(f_{i_u}^*(\bx_0+\mathfrak{t}(\bu_1-\bx_0))-f_k^*(\bx_0+\mathfrak{t}(\bu_1-\bx_0))\big) > \tilde c\|\bu_1-\bx_0\|_2,$$ because otherwise $|f_{i_u}^*(\bu_1) - f_k^*(\bu_1)|\leq T^{-\frac{2m-1}{4m}}$ or strictly decreasing, which contradicts \eqref{eq:fiu-fk>T-2m-14m}. By Claim \ref{lemcnsl_claim3}, since $\bu_1\in \big(B(\cX_0^{(k)})\setminus\cX_0^{(k)}\big)\cap S_{k,q}$, and $\bx_0$ is the projection of $\bu_1$ on $\cX_0^{(k)}$, it holds that $\bx_0+\mathfrak{t}(\bu_1-\bx_0)\in S_{k,q}\subseteq  S_{k,q-1}$. Moreover, recall that $i_u\in \cQ_{\bu_1}$. Since (ii) in Assumption \ref{assum:regularity} implies that $\cQ_{\bu_1}\subset \cQ_{\bx_0}$, we have $i_u\in \cQ_{\bx_0}$. Thus by Assumption \ref{assum:regularity} and Claim \ref{lemcnsl_claim1}, it holds that $\bx_0+\mathfrak{t}(\bu_1-\bx_0)\in R_{i_u}$, which implies that $\bx_0+\mathfrak{t}(\bu_1-\bx_0)\in S_{i_u,q-1}$. Therefore, $\bx_0+\mathfrak{t}(\bu_1-\bx_0)\in S_{i_u,q-1}\cap  S_{k,q-1}$ for all $0\leq \mathfrak{t}\leq 1$. Then, by \eqref{eq_lm6_gdlb1}, we have that
\begin{align*}
    \frac{{\rm d}}{{\rm d}\mathfrak{t}} \big(\hat{f}_{i_u,{q-1}}(\bx_0+\mathfrak{t}(\bu_1-\bx_0))-\hat{f}_{k,{q-1}}(\bx_0+\mathfrak{t}(\bu_1-\bx_0))\big) > \frac{3\tilde c}{4}\|\bu_1-\bx_0\|_2>\frac{\mathfrak{r}_0\tilde c}{4},
\end{align*}
for all $0\leq\mathfrak{t}\leq 1$. Then, we apply first order Taylor expansion to function $$p(\mathfrak{t})=\hat{f}_{i_u,{q-1}}(\bx_0+\mathfrak{t}(\bu_1-\bx_0))-\hat{f}_{k,{q-1}}(\bx_0+\mathfrak{t}(\bu_1-\bx_0))$$ at $\bx_0$, which gives us %
\begin{align}\label{eq_lm6_gdc1lb2}
    & \hat{f}_{i_u,{q-1}}(\bu_1) - \hat{f}_{k,{q-1}}(\bu_1)\nonumber\\
    = & \hat{f}_{i_u,{q-1}}(\bx_0) - \hat{f}_{k,{q-1}}(\bx_0) +
    \frac{{\rm d}}{{\rm d}\mathfrak{t}} \big(\hat{f}_{i_u,{q-1}}(\bx_0+\xi(\bu_1-\bx_0))-\hat{f}_{k,{q-1}}(\bx_0+\xi(\bu_1-\bx_0))\big) \nonumber\\
    \geq & \hat{f}_{i_u,{q-1}}(\bx_0) - \hat{f}_{k,{q-1}}(\bx_0) + \frac{\mathfrak{r}_0\tilde c}{4},
\end{align}
where $\xi$ lies between $0$ and $1$.
 Then, under event $\bigcap_{1 \leq j \leq {q-1}}\mathcal{A}_{j}$, it can be seen that
\begin{align}\label{eq_lm6_gdc1lb3}
   \hat{f}_{i_u,q-1}(\bx_0) - \hat{f}_{k,q-1}(\bx_0) & =\hat{f}_{i_u,q-1}(\bx_0) - f_{i_u}^*(\bx_0) + f_{i_u}^*(\bx_0) - f_k^*(\bx_0) + f_k^*(\bx_0)- \hat{f}_{k,q-1}(\bx_0)\nonumber\\
   & \geq -\frac{1}{8}\epsilon_{q-1} - \frac{1}{8}\epsilon_{q-1} = -\frac{1}{4}\epsilon_{q-1}.
\end{align}
Combining \eqref{eq_lm6_gdc1lb2} and \eqref{eq_lm6_gdc1lb3}, $\epsilon_{q-1}<\frac{\mathfrak{r}_0\tilde c}{8}$ as long as $T$ is sufficiently large, then
we obtain
\begin{align*}
    \hat{f}_{i_u,q-1}(\bu_1) - \hat{f}_{k,q-1}(\bu_1) \geq \frac{\mathfrak{r}_0\tilde c}{4} - \frac{1}{4}\epsilon_{q-1} \geq \frac{\mathfrak{r}_0\tilde c}{8}>\epsilon_{q-1},
\end{align*}
which implies $\bu_1\notin \cS_{k,q}$ and leads to a contradiction. This shows Claim \ref{lemcnsl_claim4} holds. \hfill\Halmos

\subsection{Proof of Proposition~\ref{prop:Aqhold}}\label{pf:Aqhold}

We begin by checking the conditions in Theorem \ref{thm:superrorbound}. First, under event
$\bigcap_{1 \leq h \leq {q-1}} \mathcal{A}_h$, Proposition~\ref{prop:cnstlength} implies that for any $q=1,...,Q$ and $k=1,\ldots,K$, the sample support  $\cS_{k,q}$ is $\mathscr{C}$-regular. In addition, Proposition~\ref{prop:dst&dist} implies that under event
$\bigcap_{1 \leq h \leq {q-1}} \mathcal{A}_h$, $\cD_{k,q}$ are i.i.d. conditionally on $\{\cD_{h}\}_{h=1}^{q-1}$, and $\frac{1}{K} p_{\min }\leq \mu_k^{(q)}(\bx)\leq {\frac{K}{p^*} p_{\max }}$ on $\cS_{k,q}$. It remains to check
\begin{align}\label{eq_pfthmaq_wncond}
    w_{n}^{-\frac{2m-1}{4m^2}} \gamma_{n}+w_{n}^{-\frac{1}{2m}} \sqrt{\log (d / \delta) / n}=o(1)
\end{align}
for $n=\tau_{k,q}$ and $\delta=1/T$. Note that by Proposition~\ref{prop:sfd}, we have that
\begin{align*}
    \PP\left(\tau_{k,q}\geq \frac{p^*}{2K}\tau_q\bigg|\{\cD_{h}\}_{h=1}^{q-1},\bigcap_{1 \leq h \leq q-1} \mathcal{A}_h\right)\geq 1-e^{-\frac{\tau_qp*^2}{2K^2}}.
\end{align*}
Then, by the union bound and $T>e^{1\vee (1/C_{9})}$, where $C_{9}=\frac{p^{*2}}{2K^2}C_2$, it follows that
\begin{align}\label{eq:tauq+1conditiononAj}
    \PP\left(\min_k\tau_{k,q}\geq \frac{p^*}{2K}\tau_q\bigg|\{\cD_{h}\}_{h=1}^{q-1},\bigcap_{1 \leq h \leq q-1} \mathcal{A}_h\right)\geq 1-Ke^{-\frac{\tau_qp*^2}{2K^2}}\geq 1-\frac{K}{T}.
\end{align}
Define the event $\cD_q = \{\min_k\tau_{k,q}\geq \frac{p^*}{2K}\tau_q\}$. Under event $\cD_q$, we have that $$\min_k\tau_{k,q}\geq \frac{p^*}{2K}\tau_q\geq \frac{p^*}{2K}\tau_1>C_9 (\log(d)+\log(T))\log(T),$$ which implies $\log(T) /\tau_{k,q}=o(1)$ and $\log(d) /\tau_{k,q}=o(1)$. According to the definition of $w_{n}$ and $\gamma_{n}$, we obtain
\begin{align*}
    & w_{n}^{-\frac{2m-1}{4m^2}} \gamma_{n}+w_{n}^{-\frac{1}{2m}} \sqrt{\log (d / \delta) / n}\\
    \lesssim &(\tau_{k,q})^{\frac{2m-1}{4m(2m+1)}} (\tau_{k,q})^{-\frac{m}{2m+1}}+\left(\frac{\log(d)+\log(T)}{\tau_{k,q}}\right)^{-\frac{1}{4m}}\sqrt{(\log(d) +\log(T)) / \tau_{k,q}}\nonumber\\
    = & (\tau_{k,q})^{\frac{-2m(2m-1)-1}{4m(2m+1)}}+\left(\frac{\log(d)+\log(T)}{\tau_{k,q}}\right)^{\frac{2m-1}{4m}}\nonumber\\
     = & o(1),
\end{align*}
where the last line is because $m>3/2$, which verifies \eqref{eq_pfthmaq_wncond}. Therefore, the conditions of Theorem \ref{thm:superrorbound} hold for any $q=1,...,Q$ and $k=1,\ldots,K$.

Next, we show \eqref{eq_thmaq_1} in Proposition~\ref{prop:Aqhold}.
Define the event $\mathcal{B}_{q+1}$ as
\begin{align*}
\mathcal{B}_{q+1}=\mathcal{B}_{q+1}^{(1)}\cap\mathcal{B}_{q+1}^{(2)}\cap\mathcal{B}_{q+1}^{(3)},
\end{align*}
where
\begin{align*}
    \mathcal{B}_{q+1}^{(1)}&=\bigcap_{1\leq k\leq K}\left\{
            \|\hat{f}_{k,{q+1}} - f^*_k\|_{L_2(\cS_{k,q+1})}
        \lesssim s_0^{\frac{1}{2}}\left(\tau_{k,q+1}^{-\frac{m}{2m+1}}+\sqrt{(\log(d) +\log(T)) / \tau_{k,q+1}}\right)\right\},\nonumber\\
    \mathcal{B}_{q+1}^{(2)}&=\bigcap_{1\leq k\leq K}\left\{\left\|\hat{f}_{k,{q+1}} - f^*_k\right\|_{\mathcal{N}_\Phi(\cS_{k,q+1})} \lesssim s_0\right\},\nonumber\\
    \mathcal{B}_{q+1}^{(3)}&=\bigcap_{1\leq k\leq K}\left\{ \|\hat{f}_{k,{q+1}} - f^*_k\|_{L_\infty(\cS_{k,q+1})} \lesssim s_0^{\frac{2m+1}{4m}}\left(\tau_{k,q+1}^{-\frac{m}{2m+1}}+\sqrt{(\log(d) +\log(T)) / \tau_{k,q+1}}\right)^{1-\frac{1}{2m}}\right\}.
\end{align*}
Therefore, by Theorem \ref{thm:superrorbound}, together with the union bound, we have that
\begin{align}\label{eq:probeventbq}
    \PP\left(\mathcal{B}_{q+1}\bigg|\bigcap_{1 \leq h \leq q} \mathcal{A}_h,\{\cD_{h}\}_{h=1}^{q},\min_i\tau_{k,q+1}\geq \frac{p^*}{2K}\tau_{q+1}\right)\geq 1-\frac{2K}{T}.
\end{align}
By \eqref{eq:tauq+1conditiononAj} and \eqref{eq:probeventbq}, together with Proposition~\ref{prop:cnstlength}, we can obtain that
\begin{align*}
    & \PP\left(\mathcal{A}_{q+1}\bigg|\bigcap_{1 \leq h \leq q} \mathcal{A}_h,\{\cD_{h}\}_{h=1}^{q}\right)\nonumber\\
    \geq & \PP\left(\mathcal{B}_{q+1}\cap\left(\min_k\tau_{k,q+1}\geq \frac{p^*}{2K}\tau_{q+1}\right)\bigg|\bigcap_{1 \leq h \leq q} \mathcal{A}_h,\{\cD_{h}\}_{h=1}^{q}\right)\nonumber\\
    = & \PP\left(\mathcal{B}_{q+1}\bigg|\bigcap_{1 \leq h \leq q} \mathcal{A}_h,\{\cD_{h}\}_{h=1}^{q},\min_k\tau_{k,q+1}\geq \frac{p^*}{2K}\tau_{q+1}\right)\PP\left(\min_k\tau_{k,q+1}\geq \frac{p^*}{2K}\tau_{q+1}\bigg|\bigcap_{1 \leq h \leq q} \mathcal{A}_h,\{\cD_{h}\}_{h=1}^{q}\right)\nonumber\\
    \geq & \left(1-\frac{2K}{T}\right)\left(1-\frac{K}{T}\right)\nonumber\\
    \geq & 1-\frac{3K}{T}.
\end{align*}
Furthermore, when marginalizing over the history samples $\{\cD_{h}\}_{h=1}^{q}$, we have
\begin{align}\label{eq:probAq+1conditiononAj}
\PP\left(\mathcal{A}_{q+1}\bigg|\bigcap_{1 \leq h \leq q} \mathcal{A}_h\right)\geq  1-\frac{3K}{T},
\end{align}
which shows the first inequality in \eqref{eq_thmaq_1} of Proposition~\ref{prop:Aqhold}.

Finally, we show the second inequality in \eqref{eq_thmaq_1} of Proposition~\ref{prop:Aqhold}. Since epoch 1 is the pure exploration region and the covariate support for all arms are the whole space $\Omega$, we have
\begin{align*}
\PP\left(\mathcal{A}_{1}\bigg|\min_k\tau_{k,1}\geq \frac{p^*}{2K}\tau_{1}\right)\geq 1-\frac{2K}{T},
\end{align*}
and
\begin{align*}
    \PP\left(\min_k\tau_{k,1}\geq \frac{p^*}{2K}\tau_{1}\right)\geq 1-\frac{K}{T},
\end{align*}
which give us that
\begin{align}\label{eq:probA1}
    \PP(\mathcal{A}_{1})\geq 1-\frac{3K}{T}.
\end{align}
Combining \eqref{eq:probAq+1conditiononAj} and \eqref{eq:probA1}, by induction, we can conclude that
\begin{align*}
   \PP\left(\bigcap_{1 \leq h \leq q} \mathcal{A}_h\right)=& \PP\left(\mathcal{A}_{q}\bigg|\bigcap_{1 \leq h \leq q-1} \mathcal{A}_h\right)\PP\left(\bigcap_{1 \leq h \leq q-1} \mathcal{A}_h\right)\nonumber\\
   \geq &\left(1-\frac{3K}{T}\right)\PP\left(\bigcap_{1 \leq h \leq q-1} \mathcal{A}_h\right)\nonumber\\
   \geq & \ldots \geq \left(1-\frac{3K}{T}\right)^{q-1} \PP\left(\mathcal{A}_{1}\right)\nonumber\\
   \geq &\left(1-\frac{3K}{T}\right)^{q}\nonumber\\
   \geq & 1-q\frac{3K}{T},
\end{align*}
where the last inequality is by Bernoulli's inequality. This completes the proof of the second inequality in \eqref{eq_thmaq_1} of Proposition~\ref{prop:Aqhold}.\hfill\Halmos

\subsection{Proof of Lemma~\ref{prop:armioptimal}}

We prove the first statement by induction. Denote $a=\arg\max_{1\leq j \leq K}f^*_j(\bx_t)$. First, obviously $a\in \cK_{0}(\bx_t)$. For $h-1\leq q-3$, assume $a\in \cK_{h-1}(\bx_t)$, which implies that $\bx_t\in S_{a,h}$. Then, denote $b_h= \arg\max_{j\in\cK_{h-1}(\bx_t)}\hat{f}_{j,{h}}(\bx_t)$, we have that
\begin{align*}
   \hat{f}_{b_h,h}(\bx_t)-\hat{f}_{a,h}(\bx_t)&= (\hat{f}_{b_h,h}(\bx_t)-f^*_{b_h}(\bx_t))+f^*_{b_h}(\bx_t)-f^*_a(\bx_t)-(\hat{f}_{a,h}(\bx_t)-f^*_a(\bx_t)) \nonumber\\
   & \leq |\hat{f}_{b_h,h}(\bx_t)-f^*_{b_h}(\bx_t)|+f^*_{b_h}(\bx_t)-f^*_a(\bx_t)+|\hat{f}_{a,h}(\bx_t)-f^*_a(\bx_t)| \nonumber\\
   &\leq 1/8\epsilon_h+1/8\epsilon_h\nonumber\\
    &\leq \epsilon_h,
\end{align*}
where the second inequality is because $f^*_{b_h}(\bx_t)\leq f^*_a(\bx_t)$ and event $\bigcap_{1 \leq h \leq q-2} \mathcal{A}_h$ .
Therefore, the induction assumption implies that $a\in \cK_{h}(\bx_t)$, from which we can conclude that $a:=\arg\max_{1\leq j \leq K}f^*_j(\bx_t)\in \cK_{q-2}(\bx_t)$.

If for any $j\in \cK_{q-2}(\bx_t)\setminus \{k\}$, $\hat{f}_{k,q-1}(\bx_t)-\hat{f}_{j, q-1}(\bx_t)>\epsilon_{q-1}$, then  under event $\bigcap_{1 \leq h \leq q-1} \mathcal{A}_h$, it follows that
\begin{align*}
    f^*_k(\bx_t)-f^*_j(\bx_t)&= (f^*_k(\bx_t)-\hat{f}_{k,q-1}(\bx_t))+\hat{f}_{k,q-1}(\bx_t)-\hat{f}_{j, q-1}(\bx_t)+(\hat{f}_{j, q-1}(\bx_t)-f^*_j(\bx_t))\nonumber\\
    & \geq -|f^*_k(\bx_t)-\hat{f}_{k,q-1}(\bx_t)|+\hat{f}_{k,q-1}(\bx_t)-\hat{f}_{j, q-1}(\bx_t)-|\hat{f}_{j, q-1}(\bx_t)-f^*_j(\bx_t)|\nonumber\\
    &\geq -1/8\epsilon_{q-1}+\epsilon_{q-1}-1/8\epsilon_{q-1}\nonumber\\
    &\geq 0,
\end{align*}
where the second inequality is because $\hat{f}_{k,q-1}(\bx_t)-\hat{f}_{j, q-1}(\bx_t)>\epsilon_{q-1}$ and $\min_{k\in\cK_{q-2}(\bx_t)}-|\hat{f}_{k,q-1}(\bx_t)-f^*_{k}(\bx_t)| \geq -1/8\epsilon_{q-1}$ under event $\bigcap_{1 \leq h \leq q-1} \mathcal{A}_h$.
Therefore, arm $k$ has higher expected reward than the remaining arms in $\cK_{q-2}(\bx_t)$. Since the optimal arm $a\in\cK_{q-2}(\bx_t)$, it is only possible that arm $k$ is the optimal arm $a$. \hfill\Halmos

\subsection{Proof of Theorem \ref{thm:regretupperbound}}

According to the definition of expected cumulative regret, we have
\begin{align}\label{eq:Rt2}
    R_T&=\sum_{q=1}^{Q}\sum_{t\in \cT_q}\EE(\max_{k}f_k^*\left(\bx_t\right)-f_{\pi_t}^*\left(\bx_t\right))\nonumber\\
    &=\underbrace{\sum_{t\in \cT_1}\EE(\max_{k}f_k^*\left(\bx_t\right)-f_{\pi_t}^*\left(\bx_t\right))}_{I_1}+\underbrace{\sum_{q=2}^{Q}\sum_{t\in \cT_q}\EE\left((\max_{k}f_k^*\left(\bx_t\right)-f_{\pi_t}^*\left(\bx_t\right))\II(\bx_t\in\cH_{\pi_t, q})\right)}_{I_2}\nonumber\\
    &+\underbrace{\sum_{q=2}^{Q}\sum_{t\in \cT_q}\EE\left((\max_{k}f_k^*\left(\bx_t\right)-f_{\pi_t}^*\left(\bx_t\right))\II(\bx_t\in\cM_{\pi_t,q})\right)}_{I_3}.
\end{align}

Since $f_k^*\in \cF$ and $\cF_j\subset \cN_{\Phi}([0,1])$ for $k=1,\ldots,K$ and $j=1,...,d$, where $\cF$ and $\cF_j$ are as in \eqref{eq_functionclass}, the reproducing property of RKHSs implies that for any $\bx\in \Omega$,
\begin{align*}
    |f_k^*(\bx)| = & \left|\sum_{j=1}^{d} f^*_{k, (j)}\left(x_{j}\right)\right|
    = \left|\sum_{j=1}^{d} \langle f^*_{k, (j)}, \Phi(x_{j} - \cdot)\rangle_{\cN_{\Phi}([0,1])} \right|\nonumber\\
    \leq & \sum_{j=1}^{d} \|f^*_{k, (j)}\|_{\cN_{\Phi}([0,1])} \|\Phi(x_{j} - x_{j} )\|_{\cN_{\Phi}([0,1])}\nonumber\\
    \leq & s_0 \Phi(0) \max_{1\leq k\leq K, 1\leq j\leq d} \|f^*_{k, (j)}\|_{\cN_{\Phi}([0,1])},
\end{align*}
where the first equality is by \eqref{eq_addreward}, the first inequality is by the Cauchy--Schwarz inequality, and the last inequality is because the sparsity of $f_k^*$ is upper bounded by $s_0$. Therefore, $\|f_k^*\|_{L_\infty(\Omega)}$ is bounded by a $C_{12}s_0$, where $C_{12}$ is a positive constant not depending on $k$ and $s_0$.

By the definition of $\tau_1$, the first term $I_1$ in \eqref{eq:Rt2} can be bounded by
\begin{align}\label{eq_regret_I1}
    I_1 \leq 2\tau_1\max _{j}\|f_j^*\|_{L_\infty(\Omega)}\leq 2C_1C_{12}s_0 (32s_0^{\frac{2m+1}{4m}})^{\frac{4m+2}{2m-1}}(\log(d)+\log(T))\log(T).
\end{align}

It remains to bound $I_2$ and $I_3$. For the exploitation region $\cH_{\pi_t, q}$, by Proposition~\ref{prop:armioptimal}, if $\bigcap_{1 \leq h \leq q-1} \mathcal{A}_h$ occurs, then $\pi_t$ is indeed the optimal arm, which implies the regret $\max_{k}f_k^*\left(\bx_t\right)-f_{\pi_t}^*\left(\bx_t\right)$ is zero. Therefore, $I_2$ can be written as
\begin{align}\label{eq_I2_reform1}
    I_2 = & \sum_{q=2}^{Q}\sum_{t\in \cT_q}\EE\left((\max_{k}f_k^*\left(\bx_t\right)-f_{\pi_t}^*\left(\bx_t\right))\II(\bx_t\in\cH_{\pi_t, q})\II\left(\bigcap_{1 \leq h \leq q-1} \mathcal{A}_h\right)\right)\nonumber\\
    & +\sum_{q=2}^{Q}\sum_{t\in \cT_q}\EE\left((\max_{k}f_k^*\left(\bx_t\right)-f_{\pi_t}^*\left(\bx_t\right))\II(\bx_t\in\cH_{\pi_t, q})\II\left(\bigcup_{1 \leq h \leq q-1} \mathcal{A}_h^\complement\right)\right)\nonumber\\
    =&\sum_{q=2}^{Q}\sum_{t\in \cT_q}\EE\left((\max_{k}f_k^*\left(\bx_t\right)-f_{\pi_t}^*\left(\bx_t\right))\II(\bx_t\in\cH_{\pi_t, q})\II\left(\bigcup_{1 \leq h \leq q-1} \mathcal{A}_h^\complement\right)\right)\nonumber\\
    \leq & 2C_{12}s_0\sum_{q=2}^{Q}\sum_{t\in \cT_q}\PP\left(\bigcup_{1 \leq h \leq q-1} \mathcal{A}_h^\complement\right).
\end{align}
Proposition~\ref{prop:Aqhold} provides that
\begin{align}\label{eq:regretAjc}
    & \sum_{q=2}^{Q}\sum_{t\in \cT_q}\PP\left(\bigcup_{1 \leq h \leq q-1} \mathcal{A}_h^\complement\right) \leq \sum_{q=2}^{Q}\tau_q\left((q-1)\frac{3K}{T}\right)\nonumber\\
    \leq & \sum_{q=2}^{Q}\tau_q\left(Q\frac{3K}{T}\right) \leq  \left(Q\frac{3K}{T}\right) T= 3KQ \leq C_{13}\log(T),
\end{align}
where the third inequality is because $$Q\leq \left\lceil{\frac{2m-1}{(4m+2)\log(2)}}\log \left(\frac{T}{C_2(\log(d)+\log(T))\log(T)}\right)\right\rceil.$$
By \eqref{eq_I2_reform1} and \eqref{eq:regretAjc}, the term $I_2$ can be bounded by
\begin{align}\label{eq_I2_bound}
    I_2 \leq C_{14}s_0 \log(T) .
\end{align}

Next, we consider bounding $I_3$, which depends on the exploration region $\cM_{\pi_t,q}$. Similarly, $I_3$ can be written by
\begin{align*}
    I_3 =& \underbrace{\sum_{q=2}^{Q}\sum_{t\in \cT_q}\EE\left((\max_{k}f_k^*\left(\bx_t\right)-f_{\pi_t}^*\left(\bx_t\right))\II(\bx_t\in\cM_{\pi_t,q})\II\left(\bigcap_{1 \leq h \leq q-1} \mathcal{A}_h\right)\right)}_{I_{31}}\nonumber\\
    +&\sum_{q=2}^{Q}\sum_{t\in \cT_q}\EE\left((\max_{k}f_k^*\left(\bx_t\right)-f_{\pi_t}^*\left(\bx_t\right))\II(\bx_t\in\cM_{\pi_t,q})\II\left(\bigcup_{1 \leq h \leq q-1} \mathcal{A}_h^\complement\right)\right)\nonumber\\
    \leq & I_{31} + 2C_{12}s_0\sum_{q=2}^{Q}\sum_{t\in \cT_q}\PP\left(\bigcup_{1 \leq h \leq q-1} \mathcal{A}_h^\complement\right)\nonumber\\
    \leq & I_{31} + C_{14}s_0 \log(T).
\end{align*}

In the exploration region $\cM_{\pi_t,q}$, it can be shown that the regret is small as follows. Under event $\bigcap_{1 \leq h \leq q-1} \mathcal{A}_h$, our selection rule in SPARKLE ensures that for $\bx_t\in\cM_{\pi_t,q}$,
\begin{align*}
    &\max_{j}f_j^*\left(\bx_t\right)-f_{\pi_t}^*\left(\bx_t\right)\nonumber\\
    =&f_a^*\left(\bx_t\right)-\hat{f}_{a, q-1}(\bx_t)+\hat{f}_{a, q-1}(\bx_t)-\hat{f}_{\pi_t,q-1}(\bx_t)+\hat{f}_{\pi_t,q-1}(\bx_t)-f_{\pi_t}^*\left(\bx_t\right)\nonumber\\
    \leq& \frac{1}{8}\epsilon_{q-1}+\epsilon_{q-1}+\frac{1}{8}\epsilon_{q-1}\nonumber\\
    =&\frac{5}{4}\epsilon_{q-1},
\end{align*}
where we let $a:=\arg\max_{j}f_j^*\left(\bx_t\right)$. The inequality is given by event $\bigcap_{1 \leq h \leq q-1} \mathcal{A}_h$, and the fact that
$\bx_t\in S_{\pi_t,q}\cap S_{a,q}$ implied by $\bx_t\in \cM_{\pi_t,q}$ and $a:=\arg\max_{j}f_j^*\left(\bx_t\right)$ (Lemma \ref{prop:armioptimal}).

Since $$\frac{5}{4}\epsilon_{q-1}\geq \frac{5}{4}\epsilon_{Q-1}=\frac{5}{2}\epsilon_{Q}\geq T^{-\frac{2m-1}{4m}}\geq \delta_0$$ for sufficiently large $T$, we have that $I_{31}$ can be bounded by
\begin{align}\label{eq_I31}
    I_{31} \leq & \sum_{q=2}^{Q}\sum_{t\in \cT_q}\EE\left((\max_{k}f_k^*\left(\bx_t\right)-f_{\pi_t}^*\left(\bx_t\right))\II(\bx_t\in\cM_{\pi_t,q})\II\left(\bigcap_{1 \leq h \leq q-1} \mathcal{A}_h\right)\right)\nonumber\\
    \leq & \sum_{q=2}^{Q}\tau_q \frac{5\epsilon_{q-1}}{4}\PP\left(\exists k,j \mbox{ such that } 0<\left|f_k^*\left(\bx_t\right) - f_j^*\left(\bx_t\right)\right|\leq \frac{5\epsilon_{q-1}}{4}\right) \nonumber\\
    \leq & \sum_{q=2}^{Q} \frac{5\epsilon_{q-1}}{4}\tau_q\sum_{k,j\in \cK}\PP\left(0< \left|f_k^*\left(X\right) - f_j^*\left(X\right)\right|\leq \frac{5\epsilon_{q-1}}{4}\right) \nonumber\\
    \leq & C_0K^2\sum_{q=2}^{Q}\tau_q\left(\frac{5}{4}\epsilon_{q-1}\right)^{1+\alpha}\nonumber\\
    = & O\left(s_0^{\frac{2m+1}{4m}(\frac{4m+2}{2m-1})}(\log(d)+\log(T))^{\frac{(2m-1)(1+\alpha)}{4m+2}} (\log T)^{\frac{(2m-1)(1+\alpha)}{4m+2}}T^{1-\frac{(2m-1)(1+\alpha)}{4m+2}}\right),
\end{align}
where the third inequality is by the union bound, and the fourth inequality is because of the margin condition in Assumption \ref{assum_margin}.

By plugging \eqref{eq_regret_I1}, \eqref{eq_I2_bound} and \eqref{eq_I31} into \eqref{eq:Rt2}, we obtain the final regret bound $R_T=\tilde O\left(s_0^{\frac{2m+1}{2m}(\frac{2m+1}{2m-1})}T^{1-\frac{(2m-1)(1+\alpha)}{4m+2}}\log(d)\right)$. This finishes the proof. \hfill\Halmos

\subsection{Proof of Theorem \ref{thm_lb}}

First, we adopt the same basis function $u(x)$ applied in \cite{hu2022smooth}: consider an infinitely differentiable function $u_1$ defined as
\begin{align*}
u_1(x)= \begin{cases}\exp \left\{-\frac{1}{\left(\frac{1}{2}-x\right)\left(x-\frac{1}{4}\right)}\right\}, & \text { if } x \in\left(\frac{1}{4}, \frac{1}{2}\right), \\ 0, & \text { otherwise, }\end{cases}
\end{align*}
and take $u: \mathbb{R}_{+} \rightarrow \mathbb{R}_{+}$ to be
\begin{align*}
  u(x)=\left(\int_{\frac{1}{4}}^{\frac{1}{2}} u_1(t) d t\right)^{-1} \int_x^{\infty} u_1(t) d t.
\end{align*}
One can verify that $u(x)$ is a non-increasing infinitely differentiable function satisfying $u(x)=1$ for $x\in [0,1/4]$, $u(x)\in(0,1)$ for $x\in(1/4,1/2)$, and $u(x)=0$ for $x\in[1/2,\infty]$. Fix any $0<\alpha\leq 1/m$ and denote $\mathcal{X}_{\tilde{\delta}}=\{x:0<u(|x|) \leq \tilde{\delta}\}$, then by cumbersome calculations we can find that there exists a positive constant $C_\alpha$ only dependent on $\alpha$ such that for any $\tilde{\delta}\geq \tilde{\delta}_0:=(\frac{1}{20C_\alpha})^{\frac{1}{\alpha}}$,
\begin{align}\label{eq:lebesguelessthan10xx}
    \text{Leb}[\mathcal{X}_{\tilde{\delta}}]\leq C_\alpha\tilde{\delta}^\alpha.
\end{align}

Define data points $x_k=\frac{2k-1}{2q},k=1,...,q$, and $g_k(x)=q^{-m}u(|q(x-x_k)|)\mathbb{I}(x\in[\frac{k-1}{q},\frac{k}{q}])$. Let $q=\lceil T^{\frac{1}{2m+1}}\rceil$, $N=\lfloor q^{1-\alpha m}\rfloor$, $\sigma = (\sigma_1,...,\sigma_N) \in \{-1,1\}^N$, $\eta_{-1}^\sigma(x) = \frac{1}{2}$, and
\begin{align*}
    \eta_1^\sigma(x) = \frac{1}{2} + \sum_{k=1}^{N} \sigma_k g_k(x).
\end{align*}
 Clearly, under the condition that $\alpha m \leq 1$, the integer $N\geq 1$. Assume $x_t$'s are i.i.d. uniformly distributed on $[0,1]$ and the rewards $Y^{(1)}$ and $Y^{(-1)}$ given $x_t$ for arms $\pm 1$ are corrupted by Gaussian noise with mean zero and variance $\sigma_\epsilon^2$ satisfying $\frac{8}{9}<\sigma_\epsilon^2<1$.

We check that the constructed instance satisfies the assumptions of our problem setting one by one.
\begin{itemize}

\item  Nonvanishing Density (Assumption \ref{assum_densitybound}).

Since we let $x_t$ be i.i.d. uniformly distributed on $\Omega=[0,1]^d$, this assumption is satisfied.

\item  Sub-Gaussian Noise (Assumption \ref{assum_subG}).

    This assumption is obviously satisfied because we assume a Gaussian noise, that is, the conditional distributions of $Y^{(1)}$ and $Y^{(-1)}$ given $x$ are Gaussian distribution with mean $\eta_{1}^\sigma$ and $\eta_{-1}^\sigma$.

\item Smooth Kernels (Assumption \ref{assump:kernel}).

Following similar steps in the proof of Theorem 3 in \cite{hu2022smooth}, we have $\eta_1^\sigma(x)$ is $(m, C, \Omega)$-H\"older for some positive constant $C$, which implies that $\eta_1^\sigma(x)$ lies in $\cB(H^m(\Omega), L)$ for some positive constant $L$.

It is obvious that $\eta_{-1}^\sigma\equiv 1/2$ also lies in $\cB(H^m(\Omega), L)$ for this $L$.

\item Compatibility Condition (Assumption \ref{assum_compatibility}).

The verification for this condition is trivial, because $s_0=1$ gives that $f_{(j)}=f$.

\item Nearly Optimal Region (Assumption \ref{assum_optprob}).

Since $|\eta_1^\sigma(x)-\eta_{-1}^\sigma(x)|\leq T^{-\frac{m}{2m+1}}\leq T^{-\frac{2m-1}{4m}}$ as long as $T$ is sufficiently large, we have $\tilde{\mathcal{R}}_i=[0,1]$ for $i=\pm 1$. Thus, Assumption \ref{assum_optprob} is satisfied.

\item Regularized support (Assumption \ref{assum:regularity}).

Also, since $|\eta_1^\sigma(x)-\eta_{-1}^\sigma(x)|\leq T^{-\frac{2m-1}{4m}}$ as long as $T$ is sufficiently large, by Lemma \ref{lemma:regularity}, we have Assumption \ref{assum:regularity} is satisfied.

\item Sufficient Optimal Region (Assumption \ref{assum_optimalregionprob})
Since $\mathcal{R}_i\supset \{x\in [0,1]|\sum_{k=1}^N \sigma_k g_k(x)=0\}$ and ${\rm Vol}\left(\{x\in [0,1]|\sum_{k=1}^N \sigma_k g_k(x)=0\}\right)\geq 1-Nq^{-1}\geq 1-T^{-\frac{\alpha m}{2m+1}}$ for $i=\pm 1$, we can define $p^*$ as any positive number smaller than $1-T^{-\frac{\alpha m}{2m+1}}$.

 \item Margin Condition (Assumption \ref{assum_margin}).

Let $\delta_0=(\frac{1}{20C_\alpha})^{\frac{1}{\alpha}}T^{-\frac{m}{2m+1}}$ and $\delta_0\leq T^{-\frac{2m-1}{4m}}$ as long as $T$ is sufficiently large, then for any $\delta \geq \delta_0$, we have
\begin{align*}
    \PP(0<|\eta_1^\sigma(X)-\eta_{-1}^\sigma(X)| \leq \delta) = & \PP\left(0<\left|\sum_{k=1}^{N} \sigma_k g_k(X)\right|\leq \delta\right)\nonumber\\
    = &  \PP\left(0<q^{-m}\sum_{k=1}^{N}u(|q(X-x_k)|)\mathbb{I}(x\in B_k)\leq \delta\right)\nonumber\\
    =&\PP\left(0<\sum_{k=1}^{N}u(|q(X-x_k)|)\mathbb{I}(x\in B_k)\leq \delta q^{m}\right)\nonumber\\
    =&N\PP\left(0<u(|q(X-1/(2q)|)\leq \delta q^{m}\right)
\end{align*}
where $B_k=[\frac{k-1}{q},\frac{k}{q}]$. By \eqref{eq:lebesguelessthan10xx}, since $\delta q^{m}\geq \delta_0 q^m\geq(\frac{1}{20C_\alpha})^{\frac{1}{\alpha}}=\tilde{\delta}_0$, we have
$ \text{Leb}[\mathcal{X}_{\delta q^{m}}]\leq C_\alpha \delta^\alpha q^{m\alpha}$. It follows that $$\PP\left(0<u(|q(X-1/(2q))|)\leq \delta q^{m}\right)=q^{-1}\text{Leb}[\mathcal{X}_{\delta q^{m}}] \leq C_\alpha \delta^\alpha q^{m\alpha-1},$$ which gives
\begin{align*}
    \PP(0<|\eta_1^\sigma(X)-\eta_{-1}^\sigma(X)| \leq \delta)= & N\PP\left(0<u(|q(X-1/(2q))|)\leq \delta q^{m}\right)\\
    \leq & C_\alpha N q^{m\alpha-1}\delta^\alpha\leq C_\alpha  \delta^\alpha.
\end{align*}
Thus, $\eta_1^\sigma$ and $\eta_{-1}^\sigma$ satisfy Assumption \ref{assum_margin}.

\end{itemize}

Next, we prove the lower bound of cumulative regret based on the problem instance we constructed. By the proof of Lemma 3.1 in \cite{rigollet2010nonparametric}, for any $\delta\geq \delta_0$, we have
\begin{align}\label{eq_RTpilb1}
    \sup_{f_1, f_2\in \cF}R_T(\pi,f_1,f_2)\geq \delta(\sup_{f_1, f_2\in \cF} S_T(\pi,f_1,f_2)-T\delta^\alpha),
\end{align}
where
\begin{align*}
    S_T(\pi,f_1,f_2) = \sum_{t=1}^T P_X\left(\pi_t(X) \neq \pi^*(X), f_1(X) \neq f_2(X)\right).
\end{align*}
Thus, it suffices to derive a lower bound of $\sup_{f_1, f_2\in \cF}\EE S_T(\pi,f_1,f_2)$, which can be further lower bounded by $\sup_{\eta_1^\sigma \in \cF} \EE S_T(\pi, \eta_1^\sigma,1/2)$.

For any policy $\pi$ and any $t=1, \ldots, T$, denote the joint distribution of
$
\left(x_1, Y_1\left(\pi_1\left(x_1\right)\right)\right), \ldots,\\\left(x_t, Y_t\left(\pi_t\left(x_t\right)\right)\right)
$
as $\mathbb{P}_{\pi, \sigma}^t$, and denote $\mathbb{E}_{\pi, \sigma}^t$ as the corresponding expectation,
where $\left(x_t, Y_t(1), Y_t(-1)\right)$ are generated i.i.d from $\mathbb{P}_\sigma$. Note that
\begin{align}\label{eq_suplb}
    \sup_{\eta_1^\sigma \in \cF} \EE S_T(\pi, \eta_1^\sigma,1/2) = & \sup_{\sigma \in \{-1,1\}^N} \sum_{t=1}^T \EE_{\pi, \sigma}^{t-1} P_X(\pi_t(x_t)\neq {\rm sign}(\eta_1^\sigma(x_t)-1/2))\nonumber\\
    = & \sup_{\sigma \in \{-1,1\}^N} \sum_{k=1}^N\sum_{t=1}^T\EE_{\pi, \sigma}^{t-1} P_X(\pi_t(x_t)\neq \sigma_k, x_t\in B_k )\nonumber\\
    \geq & \frac{1}{2^N}\sum_{k=1}^N\sum_{t=1}^T\sum_{\sigma \in \{-1,1\}^N}\EE_{\pi, \sigma}^{t-1} P_X(\pi_t(x_t)\neq \sigma_k, x_t\in B_k).
\end{align}
Let
\begin{align}\label{eq_Q_kt}
    Q_k^t = \sum_{\sigma_{[-k]}\in \Sigma_{N-1}}\sum_{l\in \{-1,1\}}\EE_{\pi, {\sigma_{l,-k}}}^{t-1} P_X(\pi_t(x_t)\neq l, x_t\in B_k),
\end{align}
where $\sigma_{[-k]}=\left(\sigma_1, \ldots, \sigma_{k-1}, \sigma_{k+1}, \ldots, \sigma_N\right)$, $\sigma_{l,-k}=\left(\sigma_1, \ldots, \sigma_{k-1}, l, \sigma_{k+1}, \ldots, \sigma_N\right)$ for $l=\{-1,1\}$, and $\Sigma_{N-1}=\{-1,1\}^{N-1}$.

Theorem 2.2 ($iii$) of \cite{tsybakov2009introduction} implies
\begin{align}\label{eq_lbqkt}
\sum_{l \in\{-1,1\}} \EE_{\pi, {\sigma_{l,-k}}}^{t-1} P_X(\pi_t(x_t)\neq l, x_t\in B_k) & \geq  q^{-1} \sum_{l \in\{-1,1\}} \EE_{\pi, {\sigma_{l,-k}}}^{t-1} P_X^k(\pi_t(x_t)\neq l) \nonumber\\
& \geq \frac{1}{4} q^{-1} \exp \left(-\mathcal{K}\left(\mathbb{P}_{\pi,{\sigma_{-1,-k}}}^{t-1} \times P_X^k, \mathbb{P}_{\pi, {\sigma_{1,-k}}}^{t-1} \times P_X^k\right)\right) \nonumber\\
& =\frac{1}{4} q^{-1} \exp \left(-\mathcal{K}\left(\mathbb{P}_{\pi,{\sigma_{-1,-k}}}^{t-1}, \mathbb{P}_{\pi, {\sigma_{1,-k}}}^{t-1}\right)\right),
\end{align}
where $P_X^k(\cdot)$ denotes the conditional distribution $P_X\left(\cdot \mid X \in B_k\right)$, and $\mathsf{KL}(P, Q)$ denote the Kullback--Leibler (KL) divergence between distributions $P$ and $Q$.

For any $t = 2,...,T$, let $\mathcal{F}_t$ denote the $\sigma$-algebra generated by $x_t,\left(x_s, Y_s^{\left(\pi_s\left(x_s\right)\right)}\right), s=1, \ldots, t-1$.
Let $\mathbb{P}_{\pi, \sigma}^{\cdot \mid \mathcal{F}_t}$ be the conditional distribution given $\mathcal{F}_t$, and
$\EE_{x_t}$ be the expectation with respect to the marginal distribution of $x_t$. The chain rule for KL divergence gives us for any $t = 1,...,T$,
\begin{align*}
    & \mathsf{KL}\left(\mathbb{P}_{\pi, {\sigma_{-1,-k}}}^t, \mathbb{P}_{\pi, {\sigma_{1,-k}}}^t\right) \\
    ={}& \mathsf{KL}\left(\mathbb{P}_{\pi, {\sigma_{-1,-k}}}^{t-1}, \mathbb{P}_{\pi, {\sigma_{1,-k}}}^{t-1}\right)+\mathbb{E}_{\pi, {\sigma_{-1,-k}}}^{t-1} \EE_{x_t}\left(\mathsf{KL}\left(\mathbb{P}_{\pi, {\sigma_{-1,-k}}}^{(x_t,Y_t^{\left(\pi_t\left(x_t\right)\right)})\mid \mathcal{F}_t}, \mathbb{P}_{\pi, {\sigma_{1,-k}}}^{(x_t,Y_t^{\left(\pi_t\left(x_t\right)\right)})\mid \mathcal{F}_t}\right)\right) \nonumber\\
    ={}& \mathsf{KL}\left(\mathbb{P}_{\pi, {\sigma_{-1,-k}}}^{t-1}, \mathbb{P}_{\pi, {\sigma_{1,-k}}}^{t-1}\right)+\mathbb{E}_{\pi, {\sigma_{-1,-k}}}^{t-1} \EE_{x_t}\left(\mathsf{KL}\left(\mathbb{P}_{\pi, {\sigma_{-1,-k}}}^{Y_t^{\left(\pi_t\left(x_t\right)\right)} \mid \mathcal{F}_t}, \mathbb{P}_{\pi, {\sigma_{1,-k}}}^{Y_t^{\left(\pi_t\left(x_t\right)\right)} \mid \mathcal{F}_t}\right)\right).
\end{align*}
By a standard formula for KL divergence between two Gaussian distributions, we have
\begin{align*}
    \mathsf{KL}\left(\mathbb{P}_{\pi, {\sigma_{1,-k}}}^{Y_t^{\left(\pi_t\left(x_t\right)\right)} \mid \mathcal{F}_t}, \mathbb{P}_{\pi, {\sigma_{-1,-k}}}^{Y_t^{\left(\pi_t\left(x_t\right)\right)} \mid \mathcal{F}_t}\right) & \leq \frac{1}{\sigma_\epsilon^2}\left(\eta_1^{\sigma_{1,-k}}\left(x_t\right)-\eta_1^{\sigma_{-1,-k}}\left(x_t\right)\right)^2 \mathbb{I}\left(\pi_t\left(x_t\right)=1\right)\nonumber\\
    &\leq \frac{4}{\sigma_\epsilon^2}q^{-2m} \mathbb{I}\left(\pi_t\left(x_t\right)=1, x_t \in B_k\right).
\end{align*}

By induction, we can conclude that for any $t = 1,...,T$,
\begin{align}\label{eq_ubKL}
    \mathsf{KL}\left(\mathbb{P}_{\pi,{\sigma_{-1,-k}}}^{t-1}, \mathbb{P}_{\pi, {\sigma_{1,-k}}}^{t-1}\right) \leq \frac{4}{\sigma_\epsilon^2}q^{-2m}  \mathrm{~N}_{k, \pi},
\end{align}
where
\begin{align*}
    \mathrm{N}_{k, \pi}=\mathbb{E}_{\pi, {\sigma_{-1,-k}}}^{T-1} \EE_X\left[\sum_{t=1}^T \mathbb{I}\left(\pi_t(X)=1, X \in B_k\right)\right].
\end{align*}
Combining \eqref{eq_Q_kt}, \eqref{eq_lbqkt}, and \eqref{eq_ubKL} yields
\begin{align}\label{eq_lbqktx1}
    Q_k^t \geq \frac{1}{4}2^{N-1}q^{-1} \exp \left(-\frac{4}{\sigma_\epsilon^2}q^{-2m}  \mathrm{~N}_{k, \pi}\right).
\end{align}
In addition, note that the definition of $Q^t_k$ implies
\begin{align}\label{eq_lbqktx2}
    \sum_{t=1}^T Q_k^t \geq 2^{N-1} \mathrm{~N}_{k, \pi}.
\end{align}
By \eqref{eq_suplb}, \eqref{eq_lbqktx1} and \eqref{eq_lbqktx2}, we obtain that
\begin{align}\label{eq:supEST}
    \sup_{\eta_1^\sigma \in \cF} \EE S_T(\pi, \eta_1^\sigma,1/2) & \geq \frac{2^{N-1}}{2^N} \sum_{k=1}^N \max \left\{\frac{T}{4q} \exp \left(-\frac{4}{\sigma_\epsilon^2}q^{-2m}  \mathrm{~N}_{k, \pi}\right), \mathrm{N}_{k, \pi}\right\} \nonumber\\
    & \geq \frac{N}{4} \inf _{z \geq 0}\left\{\frac{T}{4q} \exp \left(-\frac{4}{\sigma_\epsilon^2}q^{-2m}  z\right)+z\right\}\nonumber\\
    & \geq \frac{\sigma_\epsilon^2}{16}Nq^{2m}\geq \max\{\frac{\sigma_\epsilon^2}{16} q^{2m+1-\alpha m}-\frac{\sigma_\epsilon^2}{16}q^{2m},\frac{\sigma_\epsilon^2}{16}q^{2m}\},
\end{align}
provided
\begin{align*}
     \sigma_\epsilon^{-2}Tq^{-(2m+1)} > 1,
\end{align*}
which can be ensured since $\sigma_\epsilon^2<1$ and
\begin{align*}
    Tq^{-(2m+1)} \geq  T T^{-1}=1.
\end{align*}

Note that $\max\{\frac{\sigma_\epsilon^2}{16} q^{2m+1-\alpha m}-\frac{\sigma_\epsilon^2}{16}q^{2m},\frac{\sigma_\epsilon^2}{16}q^{2m}\}\geq \frac{9\sigma_\epsilon^2}{160} q^{2m+1-\alpha m}$ when $\alpha m<1$ with sufficiently large $T$, and $\max\{\frac{\sigma_\epsilon^2}{16} q^{2m+1-\alpha m}-\frac{\sigma_\epsilon^2}{16}q^{2m},\frac{\sigma_\epsilon^2}{16}q^{2m}\}= \frac{\sigma_\epsilon^2}{16} q^{2m}$ when $\alpha m=1$.
Therefore, by taking $\delta = \delta_0=(\frac{1}{20C_\alpha})^{\frac{1}{\alpha}}T^{-\frac{m}{2m+1}}$ in \eqref{eq_RTpilb1}, we have
\begin{align*}
    \sup_{f_1, f_2\in \cF}R_T(\pi,f_1,f_2)\geq & \delta\left(\max\{\frac{\sigma_\epsilon^2}{16} q^{2m+1-\alpha m}-\frac{\sigma_\epsilon^2}{16}q^{2m},\frac{\sigma_\epsilon^2}{16}q^{2m}\}-C_\alpha T\delta^\alpha\right)\nonumber\\
    \geq & (\frac{1}{20C_\alpha})^{\frac{1}{\alpha}}T^{-\frac{m}{2m+1}}\left(\frac{9\sigma_\epsilon^2}{160}T^{\frac{2m+1-\alpha m}{2m+1}}-\frac{1}{20}T^{1-\frac{m\alpha}{2m+1}}\right)\nonumber\\
    \geq & CT^{1-\frac{m}{2m+1}(1+\alpha)},
\end{align*}
where $C$ is a positive constant dependent on $\alpha$, and the last inequality is because $\sigma_\epsilon^2>\frac{8}{9}$.

Thus, we have
\begin{align*}
    R_T(\pi) =  {\Omega}( T^{1-\frac{m}{2m+1}(1+\alpha)}),
\end{align*}
which finishes the proof. \hfill\Halmos

\subsection{Proof of Theorem \ref{thm_lb_withd}}

The proof of this theorem is similar to that of Theorem \ref{thm_lb}; however, we will highlight the key differences. In the absence of specific remarks, the proof follows the same structure as that of Theorem \ref{thm_lb}.

Let the covariate support be $\Omega=[0,1]^d$, define grid $$G_q=\left\{\left(\frac{2 j_1+1}{2 q}, \ldots, \frac{2 j_d+1}{2 q}\right): j_i \in\{0, \ldots, q-1\}, i=1, \ldots, d\right\},$$ and number the data point in $G_q$ as $\bx_1,...,\bx_{q^d}$.
Let $g_k(\bx)=q^{-m}u(\|q(\bx-\bx_k)\|)\mathbb{I}(\bx\in B_k)$, where $B_k=\left\{\bx^{\prime} \in \Omega: g_q\left(\bx^{\prime}\right)=g_q(\bx_k)\right\} $ and  $g_q(\bx)=\arg \min _{\bx^{\prime} \in G_q}\left\|\bx-\bx^{\prime}\right\|$. Let $q=\lceil T^{\frac{1}{2m+d}}\rceil$, $N=\lfloor q^{d-\alpha m}\rfloor$, $\sigma = (\sigma_1,...,\sigma_N) \in \{-1,1\}^N$, $\eta_{-1}^\sigma(x) = \frac{1}{2}$, and
\begin{align*}
    \eta_1^\sigma(\bx) = \frac{1}{2} + \sum_{k=1}^{N} \sigma_k g_k(\bx).
\end{align*}
Clearly, under the condition that $\alpha m \leq d$, the integer $N\geq 1$. Assume $x_t$'s are i.i.d. uniformly distributed on $\Omega$ and the rewards $Y^{(1)}$ and $Y^{(-1)}$ given $x_t$ for arms $\pm 1$ are corrupted by Gaussian noise with mean zero and variance $\sigma_\epsilon^2$, with $\frac{8}{9}<\sigma_\epsilon^2<1$. Then, we only focus on checking the following assumptions, as the rest ones are trivial to verify.
\begin{itemize}

\item Optimal Region (Assumption \ref{new_assum_optprob}).

 Since $|\eta_1^\sigma(\bx)-\eta_{-1}^\sigma(\bx)|\leq T^{-\frac{2m-d}{4m}}$ as long as $T$ is sufficiently large, we have $\tilde{\mathcal{R}}_i=\Omega$ for $i=\pm 1$. Thus, Assumption \ref{new_assum_optprob} is satisfied.

\item Regularized Support (Assumption \ref{new_assum:regularity}).

Also, since $|\eta_1^\sigma(\bx)-\eta_{-1}^\sigma(\bx)|\leq T^{-\frac{2m-d}{4m}}$ as long as $T$ is sufficiently large, similar to the proof of Lemma \ref{lemma:regularity}, we have Assumption \ref{new_assum:regularity} is satisfied.

\item Sufficient Optimal Region (Assumption \ref{assum_optimalregionprob}).

Since $\mathcal{R}_i\supset \{\bx\in \Omega|\sum_{k=1}^N \sigma_k g_k(\bx)=0\}$ and ${\rm Vol}\left(\{\bx\in \Omega|\sum_{k=1}^N \sigma_k g_k(\bx)=0\}\right)\geq 1-Nq^{-d}\geq 1-T^{-\frac{\alpha m}{2m+d}}$ for $i=\pm 1$, we can define $p^*$ as any positive number smaller than $1-T^{-\frac{\alpha m}{2m+d}}$.

 \item Margin Condition (Assumption \ref{new_assum_margin}).

Let $\delta_0=(\frac{1}{20C_\alpha})^{\frac{1}{\alpha}}T^{-\frac{m}{2m+d}}$ and $\delta_0\leq T^{-\frac{2m-d}{4m}}$ as long as $T$ is sufficiently large, then for any $\delta \geq \delta_0$, we have
\begin{align*}
    \PP(0<|\eta_1^\sigma(X)-\eta_{-1}^\sigma(X)| \leq \delta) = & \PP\left(0<\left|\sum_{k=1}^{N} \sigma_k g_k(\bx)\right|\leq \delta\right)\nonumber\\
    = &  \PP\left(0<q^{-m}\sum_{k=1}^{N}u(\|q(\bx-\bx_k)\|)\mathbb{I}(\bx\in B_k)\leq \delta\right)\nonumber\\
    =&\PP\left(0<\sum_{k=1}^{N}u(\|q(\bx-\bx_k)\|)\mathbb{I}(\bx\in B_k)\leq \delta q^{m}\right)\nonumber\\
    =&N\PP\left(0<u(\|q(\bx-\bx_1)\|)\leq \delta q^{m}\right).
\end{align*}
 By \eqref{eq:lebesguelessthan10xx}, since $\delta q^{m}\geq \delta_0 q^m\geq(\frac{1}{20C_\alpha})^{\frac{1}{\alpha}}=\tilde{\delta}_0$, we have
$ \text{Leb}[\mathcal{X}_{\delta q^{m}}]\leq C_\alpha \delta^\alpha q^{m\alpha}$. It follows that $$\PP\left(0<u(\|q(\bx-\bx_1)\|)\leq \delta q^{m}\right)=q^{-d}\text{Leb}[\mathcal{X}_{\delta q^{m}}] \leq C_\alpha \delta^\alpha q^{m\alpha-d},$$ which gives
\begin{align*}
    \PP(0<|\eta_1^\sigma(X)-\eta_{-1}^\sigma(X)| \leq \delta)= & N\PP\left(0<u(\|q(\bx-\bx_k)\|)
    \leq \delta q^{m}\right)\\
    \leq & C_\alpha N q^{m\alpha-d}\delta^\alpha\leq C_\alpha  \delta^\alpha.
\end{align*}
 Thus, $\eta_1^\sigma$ and $\eta_{-1}^\sigma$ satisfy Assumption \ref{assum_margin}.

\end{itemize}

Then, by following the similar steps before arriving \eqref{eq:supEST}, we can obtain that
\begin{align*}
    \sup_{\eta_1^\sigma \in \cF} \EE S_T(\pi, \eta_1^\sigma,1/2) & \geq \frac{2^{N-1}}{2^N} \sum_{k=1}^N \max \left\{\frac{T}{4q^d} \exp \left(-\frac{4}{\sigma_\epsilon^2}q^{-2m}  \mathrm{~N}_{k, \pi}\right), \mathrm{N}_{k, \pi}\right\} \nonumber\\
    & \geq \frac{N}{4} \inf _{z \geq 0}\left\{\frac{T}{4q^d} \exp \left(-\frac{4}{\sigma_\epsilon^2}q^{-2m}  z\right)+z\right\}\nonumber\\
    & \geq \frac{\sigma_\epsilon^2}{16}Nq^{2m}\geq \max\{\frac{\sigma_\epsilon^2}{16} q^{2m+d-\alpha m}-\frac{\sigma_\epsilon^2}{16}q^{2m},\frac{\sigma_\epsilon^2}{16}q^{2m}\},
\end{align*}
provided
\begin{align*}
     \sigma_\epsilon^{-2}Tq^{-(2m+d)} > 1,
\end{align*}
which can be ensured since $\sigma_\epsilon^2<1$ and
\begin{align*}
    Tq^{-(2m+d)} \geq  T T^{-1}=1.
\end{align*}
Therefore, by taking $\delta = \delta_0=(\frac{1}{20C_\alpha})^{\frac{1}{\alpha}}T^{-\frac{m}{2m+d}}$ in \eqref{eq_RTpilb1}, we have
\begin{align*}
    \sup_{f_1, f_2\in \cF}R_T(\pi,f_1,f_2)\geq & \delta\left(\max\{\frac{\sigma_\epsilon^2}{16} q^{2m+d-\alpha m}-\frac{\sigma_\epsilon^2}{16}q^{2m},\frac{\sigma_\epsilon^2}{16}q^{2m}\}-C_\alpha T\delta^\alpha\right)\nonumber\\
    \geq & (\frac{1}{20C_\alpha})^{\frac{1}{\alpha}}T^{-\frac{m}{2m+d}}\left(\frac{9\sigma_\epsilon^2}{160}T^{\frac{2m+d-\alpha m}{2m+d}}-\frac{1}{20}T^{1-\frac{m\alpha}{2m+d}}\right)\nonumber\\
    \geq & CT^{1-\frac{m}{2m+d}(1+\alpha)},
\end{align*}
where $C$ is a positive constant dependent on $\alpha$, and the last inequality is because $\sigma_\epsilon^2>\frac{8}{9}$.

Thus, we have
\begin{align*}
    R_T(\pi) =  {\Omega}( T^{1-\frac{m}{2m+d}(1+\alpha)}),
\end{align*}
and finish the proof. \hfill\Halmos

\subsection{Proof of Theorem \ref{thm:lbwithoutregularity}}

We still consider the hard instances considered in Theorem \ref{thm_lb} while with distinct set of problem parameters. Let $\Delta\in(\frac{m}{\alpha m+1},\frac{1}{\alpha})$. Moreover, assume $\Delta<\frac{2m-1}{4}$ such that $\frac{\Delta}{2\Delta+1}<\frac{2m-1}{4m+2}$. Such $\Delta$ exists since $\alpha m > \frac{2m + 1}{2m - 1}$, which implies $\frac{2m-1}{4}>\frac{m}{\alpha m+1}$.

Define $\bar{p}=T^{\frac{1}{2 \Delta+1}}$, $\bar{q}=\lceil \bar{p}^{\frac{\Delta}{m}}\rceil$, $\bar{N}=\lfloor \bar{p}^{1-\alpha \Delta}\rfloor$, $B_k=[\frac{k-1}{\bar{q}},\frac{k}{\bar{q}}]$, $\mathcal{X}_0=[0,1]-\bigcup_{i=1}^{\bar{N}} B_k$. Assume $x_t$'s are i.i.d. uniformly distributed on $\Omega=\bigcup_{i=1}^{\bar{N}} \tilde{B}_k \cup \mathcal{X}_0$, where $\tilde{B}_k=[\frac{2k-1}{2\bar{q}}-\frac{1}{2\bar{p}},\frac{2k-1}{2\bar{q}}+\frac{1}{2\bar{p}}]$, and $\frac{1}{\bar{p}}$ is much smaller than $\frac{1}{\bar{q}}$ since $\Delta<\frac{1}{\alpha} < \frac{2m-1}{2m+1}m < m$. It holds that $1\leq \bar{N}\leq \bar{q}/2$ when $T$ is sufficiently large since $\alpha\Delta < 1$ and $\Delta>\frac{m}{\alpha m+1}$.

Define data points $x_k=\frac{2k-1}{2\bar{q}},k=1,...,\bar{q}$, and $g_k(x)=\bar{q}^{-m}u(|\bar{q}(x-x_k)|)\mathbb{I}(x\in[\frac{k-1}{\bar{q}},\frac{k}{\bar{q}}])$. Let
\begin{align*}
    \eta_1^\sigma(x) = \frac{1}{2} + \sum_{k=1}^{\bar{N}} \sigma_k g_k(x).
\end{align*}
All other settings are kept the same as in proof of Theorem \ref{thm_lb}. We check that functions in the problem class satisfy Assumption \ref{assum_densitybound}--\ref{assum_compatibility}, and \ref{assum_optimalregionprob}--\ref{assum_margin}, while violating Assumption \ref{assum_optprob}--\ref{assum:regularity}.
\begin{itemize}
\item  Nonvanishing Density (Assumption \ref{assum_densitybound}).

Since we let $x_t$ be i.i.d. uniformly distributed on $\Omega$, and $\Omega$ is compact, this assumption is satisfied.

\item  Sub-Gaussian Noise (Assumption \ref{assum_subG}).

    This assumption is obviously satisfied because we assume a Gaussian noise, that is, the conditional distributions of $Y^{(1)}$ and $Y^{(-1)}$ given $x$ are Gaussian distribution with mean $\eta_{1}^\sigma$ and $\eta_{-1}^\sigma$.

\item Smooth Kernels (Assumption \ref{assump:kernel}).

We have proved that $\eta_1^\sigma(x)$ lies in $\cB(H^m([0,1]), L)$ in proof of Theorem \ref{thm_lb}, thus $\eta_1^\sigma(x)$ also lies in $\cB(H^m(\Omega), L)$ since $\Omega\subseteq[0,1]$.

\item Compatibility Condition (Assumption \ref{assum_compatibility}).

The verification for this condition is trivial, because $s_0=1$ gives that $f_{(j)}=f$.

\item Sufficient Optimal Region (Assumption \ref{assum_optimalregionprob}).

Since $\mathcal{R}_i\supset \{x\in [0,1]|\sum_{k=1}^N \sigma_k g_k(x)=0\}$ and ${\rm Vol}\left(\{x\in \Omega|\sum_{k=1}^N \sigma_k g_k(x)=0\}\right)\geq 1-\bar{N}\bar{q}^{-1}\geq 1/2$ for $i=\pm 1$, we can define $p^*=1/2$.

 \item Margin Condition (Assumption \ref{assum_margin}).

Let $\delta_0=0$, we have
\begin{align*}
    \PP(0<|\eta_1^\sigma(X)-\eta_{-1}^\sigma(X)| \leq \delta) = & \PP\left(0<\left|\sum_{k=1}^{\bar{N}} \sigma_k g_k(X)\right|\leq \delta\right) =  \frac{\bar{N}/\bar{p}}{\bar{N}/\bar{p}+1-\bar{N}/\bar{q}}\mathbb{I}(\delta\geq \bar{q}^{-m}).
\end{align*}
Since $\frac{\bar{N}/\bar{p}}{\bar{N}/\bar{p}+1-\bar{N}/\bar{q}}\leq \frac{2\bar{N}}{\bar{q}}\leq 2(\bar{q}^{-\alpha m})$ by the fact that $1-\frac{\bar{N}}{\bar{q}}\geq 1/2$,  $\eta_1^\sigma$ and $\eta_{-1}^\sigma$ satisfy Assumption \ref{assum_margin}.
\end{itemize}

Next we show that the instances violates both Assumptions \ref{assum_optprob} and \ref{assum:regularity}. Recall that $\Omega = \bigcup_{i=1}^{\bar{N}} \tilde{B}_k \cup \mathcal{X}_0$. By the definition of the nearly optimal region $\tilde{\mathcal{R}}_k=\{\bx\in\Omega|\max_{1\leq i\leq K}f^*_i(\bx)-f^*_k(\bx)\leq T^{-\frac{2m-1}{4m}}\}$, and since the difference between two arms on each $\tilde{B}_k$ is $\bar{q}^{-m}=T^{-\frac{\Delta}{2\Delta+1}}>T^{-\frac{2m-1}{4m+2}}>T^{-\frac{2m-1}{4m}}$, it follows that for any $\tilde{B}_k$, it is a connected component entirely contained within either $\tilde{\mathcal{R}}_1$ or $\tilde{\mathcal{R}}_{-1}$, disjoint from other components. Since $\PP(X\in\tilde{B}_k)=\frac{1/\bar{p}}{\bar{N}/\bar{p}+1-\bar{N}/\bar{q}}\lesssim \frac{1}{\bar{p}}$, which converges to zero as $T$ becomes large, the problem class violates Assumption \ref{assum_optprob}.

For Assumption \ref{assum:regularity}, we have $\cX_0^{(1)} =\cX_0^{(-1)}= \mathcal{X}_0$. As long as $T$ is large enough, for any constant $\mathfrak{r}>0$, there exists a $l$ such that $\tilde{B}_l\subseteq B(\cX_0^{(1)})$. Then, for $x_1\in \tilde{B}_l$, clearly both $|f_i^*(\bx_1)-f_j^*(\bx_1)|> T^{-\frac{2m-1}{4m}}$ and $\left|\nabla_{\bv}\big(f^*_i(\bx_1)-  f^*_j(\bx_1)\big)\right|=0$ hold. Therefore, Assumption \ref{assum:regularity} is also violated.

We proceed by proving a lower bound for the designed problem class. It holds that $\PP(X\in\tilde{B}_k)=\frac{1/\bar{p}}{\bar{N}/\bar{p}+1-\bar{N}/\bar{q}}> \frac{1}{\bar{p}}$ since $\bar{N}/\bar{p}+1-\bar{N}/\bar{q}<1$.
By following the similar steps before arriving \eqref{eq:supEST}, we can obtain that
\begin{align*}
    \sup_{\eta_1^\sigma \in \cF} \EE S_T(\pi, \eta_1^\sigma,1/2) & \geq \frac{2^{\bar{N}-1}}{2^{\bar{N}}} \sum_{k=1}^{\bar{N}} \max \left\{\frac{T}{4\bar{p}} \exp \left(-\frac{4}{\sigma_\epsilon^2}\bar{p}^{-2\Delta}  \mathrm{~N}_{k, \pi}\right), \mathrm{N}_{k, \pi}\right\} \nonumber\\
    & \geq \frac{\bar{N}}{4} \inf _{z \geq 0}\left\{\frac{T}{4\bar{p}} \exp \left(-\frac{4}{\sigma_\epsilon^2}\bar{p}^{-2\Delta}  z\right)+z\right\}\nonumber\\
    & \geq \frac{\sigma_\epsilon^2}{16}\bar{N}\bar{p}^{2\Delta}\geq \max\{\frac{\sigma_\epsilon^2}{16} \bar{p}^{2\Delta+1-\alpha \Delta}-\frac{\sigma_\epsilon^2}{16}\bar{p}^{2\Delta},\frac{\sigma_\epsilon^2}{16}\bar{p}^{2\Delta}\},
\end{align*}
provided
\begin{align*}
     \sigma_\epsilon^{-2}T\bar{p}^{-(2\Delta+1)} > 1,
\end{align*}
which can be ensured since $\sigma_\epsilon^2<1$ and
\begin{align*}
    T\bar{p}^{-(2\Delta+1)} \geq  T T^{-1}=1.
\end{align*}
Therefore, by taking $\delta = (\frac{1}{40})^{\frac{1}{\alpha}}T^{-\frac{\Delta}{2\Delta+1}}$ in \eqref{eq_RTpilb1}, we have
\begin{align*}
    \sup_{f_1, f_2\in \cF}R_T(\pi,f_1,f_2)\geq & \delta\left(\max\{\frac{\sigma_\epsilon^2}{16} \bar{p}^{2\Delta+1-\alpha \Delta}-\frac{\sigma_\epsilon^2}{16}\bar{p}^{2\Delta},\frac{\sigma_\epsilon^2}{16}\bar{p}^{2\Delta}\}-2 T\delta^\alpha\right)\nonumber\\
    \geq & (\frac{1}{40})^{\frac{1}{\alpha}}T^{-\frac{\Delta}{2\Delta+1}}\left(\frac{9\sigma_\epsilon^2}{160}T^{\frac{2\Delta+1-\alpha \Delta}{2\Delta+1}}-\frac{1}{20}T^{1-\frac{\Delta\alpha}{2\Delta+1}}\right)\nonumber\\
    \geq & CT^{1-\frac{\Delta}{2\Delta+1}(1+\alpha)},
\end{align*}
where $C$ is a positive constant dependent on $\alpha$, and the last inequality is because $\sigma_\epsilon^2>\frac{8}{9}$.

Thus, we can conclude that as $T\rightarrow \infty$,
\begin{align}
    &\frac{R_T}{\tilde O\left(s_0^{\frac{2m+1}{2m}(\frac{2m+1}{2m-1})}T^{1-\frac{(2m-1)(1+\alpha)}{4m+2}}\right)}\nonumber\\
    =&\Omega \left(\frac{T^{1-\frac{\Delta}{2\Delta+1}(1+\alpha)}}{\tilde O(T^{1-\frac{(2m-1)(1+\alpha)}{4m+2}})}\right)\rightarrow\infty,
\end{align}
which finishes the proof.

\end{document}